\documentclass[10pt,twocolumn]{IEEEtran}

\usepackage[justification=centering]{caption}
\usepackage{graphicx,amsmath,amssymb,amstext,enumerate}
\usepackage{longtable}
\usepackage{amsfonts,hhline,rotating,cite,pstcol,subfigure}
\usepackage{epsfig}
\usepackage{epstopdf}
\DeclareMathOperator\erf{erf}
\usepackage[table,xcdraw]{xcolor}
\usepackage{url}
\usepackage{multirow}
\usepackage{amsthm}
\usepackage{float}
\usepackage{algorithmic,algorithm,cite,enumerate}
\usepackage{color} 
\usepackage{multirow} 
\graphicspath{{./figures/} }

 \allowdisplaybreaks

\begin{document}
\title{Big Data Meet Cyber-Physical Systems: A Panoramic Survey}
\author{{Rachad~Atat, Lingjia~Liu,~{\em Senior Member, IEEE}, Jinsong Wu,~{\em Senior Member, IEEE},
Guangyu Li, Chunxuan Ye, and Yang Yi,~{\em Senior Member, IEEE}}  
\IEEEcompsocitemizethanks{\IEEEcompsocthanksitem \indent
A part of this work was presented in the Ph.D. thesis of R. Atat under the guidance of Dr. L. Liu~\cite{RAtat_thesis_2017}. This work is supported in part by the U.S. National Sciecne Foundation (NSF) under grants NSF/ECCS-1802710, NSF/ECCS-1811497, and NSF/CNS-1811720, Chile CONICYT under grant Fondecyt Regular 181809, and China Hunan Provincial Nature Science Foundation under grant 2018JJ2535.
Corresponding authors are L. Liu (ljliu@ieee.org) and J. Wu (wujs@ieee.org).
R. Atat is with the Electrical and Computer Engineering Department, Texas A\&M University at Qatar, Doha, Qatar; L. Liu and Y. Yi are with the Bradley Department of Electrical and Computer Engineering, Virginia Tech, Blacksburg, VA, USA; J. Wu is with the Department of Electrical Engineering, Universidad de Chile, Santiago, Chile; G. Li is with School of Computer Science and Engineering, Nanjing University of Science \& Technology, 210094 Nanjing, P. R. China; and Chunxuan Ye is with InterDigital Communications, LLC., San Diego, CA 92121, USA.\protect\\
}
}

%
%


\maketitle


\begin{abstract}
The world is witnessing an unprecedented growth of cyber-physical systems (CPS), which are foreseen to revolutionize our world {via} creating new services and applications in a variety of sectors such as environmental monitoring, mobile-health systems, intelligent transportation systems and so on. The {information and communication technology }(ICT) sector is experiencing a significant growth in { data} traffic, driven by the widespread usage of smartphones, tablets and video streaming, along with the significant growth of sensors deployments that are anticipated in the near future. {It} is expected to outstandingly increase the growth rate of raw sensed data. In this paper, we present the CPS taxonomy {via} providing a broad overview of data collection, storage, access, processing and analysis. Compared with other survey papers, this is the first
panoramic survey on big data for CPS, where our objective is to provide a panoramic summary of different CPS aspects. Furthermore, CPS {require} cybersecurity to protect {them} against malicious attacks and unauthorized intrusion, which {become} a challenge with the enormous amount of data that is continuously being generated in the network. {Thus, we also} provide an
overview of the different security solutions proposed for CPS big data storage, access and analytics. We also discuss big data meeting green challenges in the contexts of CPS.
\end{abstract}


\IEEEpeerreviewmaketitle

\section{Introduction}

\IEEEPARstart{T}{he} growing number of ``things'', such as embedded devices, sensors, radio-frequency identification (RFID), and actuators have revolutionized the world through their integrated communication and tight interactions to create pervasive and global cyber-physical systems (CPS). Indeed, it is expected that over 50 billion sensors will be connected to the Internet, with an average of 6.58 devices per person by 2020~\cite{cisco}. This has allowed the rapid development of myriad of applications in health-care, public safety, environmental management, vehicular networks, industrial automation, and so on. A CPS mainly consists of physical components and a cyber twin interconnected together, where a cyber twin is a simulation model representative of the physical things such as a computer program~\cite{lee2015cyber}. Internet of Things (IoT), on the other hand, allows different CPS to be connected together for information transfer. This means that IoT acts as a connection bridge to network different cyber-physical things. The global expansion of interconnected CPS is facilitated by standardization efforts. For instance, the standardization activities of IoT are being led by the industry (AllSeen Alliance, Open Interconnect Consortium, Industrial Interconnect Consortium) and IEEE P2413 project on standards specifications of IoT {architectural} framework~\cite{muhonen2015standardization}.  CPS have resulted in a tsunami of new information, also known as big data, which can help boost revenues of many businesses, by identifying customer needs and providing them with superior services. However, this enormous amount of data is so large in size and complex in real-time that it exceeds the processing capacities of conventional systems. That is why, cloud computing techniques along with machine learning tools, data mining, artificial intelligence, and fog computing can help the sensed data to be easily stored, processed, and analyzed to uncover hidden patterns, unknown correlations and other useful information~\cite{6674155}. That is why big data are referred to as ``\textit{the 21$^{\text{st}}$ century new oil}''. The characteristics of big data was well summarized in the Introduction section of \cite{7473821}. The relevances of big data era and CPS actually are also highly relevant to global sustainability development goals recently discussed in \cite{wu2018information}.

Before any processing or analysis, data {need} to be acquired. The technological advancements in sensors have led to smarter, more efficient and low-cost sensors; the fact that facilitated their wide deployment. Two main sources to sense the data from: i) context-aware computing {and communications}\cite{6829939}, ii) social computing. With context-aware computing and communications\cite{6829939}, data are sensed from physical sensors, virtual sensors which retrieve data using web services technology, logical sensors which combines both virtual and physical sensors such as gathering weather information, global sensors which collect data from middleware infrastructure, and remote sensors for earth sciences applications~\cite{6512846,7565634}. As for social computing, participatory sensing and mobile crowd-sensing have led to shaping the structure of social networks, in which users collect and share sensed data using their own smartphones rather than relying on sensors~\cite{7470600,7406686,7402272}.

Cloud computing facilitates big data storage, processing and management in CPS, by breaking them down into workflows, which are then distributed over multiple dedicated servers. This allows CPS to provide pervasive sensing services beyond the capacities of individual things, in addition to lower latency and power consumption and larger scalability.

Once the data have been collected, making sense of them becomes one the most important aspects of CPS. However, it is important first to eliminate redundant information and reduce data complexity so useful information extraction can be efficiently performed. In this survey, we will discuss about different tools to assist with the data mining process, mainly, feature selection, dimensionality reduction, knowledge discovery in databases, information visualization, computer vision, classification/clustering techniques, and real-time analysis.

After the data { are} transformed into manageable { sizes}, data mining tools (HDFS~\cite{MChenn}, MapReduce~\cite{Derbeko20161}, R~\cite{Fan:2013:MBD:2481244.2481246}, S), real-time big data analytic tools (Storm~\cite{Fan:2013:MBD:2481244.2481246,6842585}, Splunk~\cite{zadrozny2013big}), and cloud-based big data analytic tools (GFS~\cite{6974788}, BigTable~\cite{2016:SPA:2974459.2974485}, MapReduce) can be used to extract useful information and make sense of data, which would revolutionize the field of smart cities, environmental monitoring and others.

The ubiquitous cyber-physical world is susceptible to security threats to a large degree. These security vulnerabilities are made easier with the inability to effectively handle the large amount of data that is constantly flowing through the network; that, in addition to the lack of qualified security experts. Sensitive data stored in the cloud can be accessed or altered by unauthorized users. Cyber-security attacks on the computations, such as false data injection, can affect the integrity and accuracy of extracted results. For all these reasons, research efforts have been shifting towards proposing robust security solutions for big data CPS. In this survey, we provide an overview of these proposed solutions.

In recent years, there { have been growing interests} in green information and communications technologies. Addressing green issues for CPS allows for a more sustainable and energy efficient systems. Green solutions are proposed for many aspects of CPS, mainly for i) data collection/storage, such as minimizing the number of relay transmissions, removing redundant transmission links, and the use of data compression techniques; ii) CPS computing such as dynamic voltage and frequency scaling and traffic engineering techniques; iii) CPS processing such as designing energy-efficient orchestrators, checkpointing aided parallel execution (CAPE), reducing the amount of exchanged data between clouds, the use of cloudlets which are closer to users than distant clouds, among other solutions.

We summarize the contributions of our paper as follows:
\begin{itemize}
\item First, to the best of our knowledge, this is the first panoramic survey on big data for cyber-physical systems. Unlike other previous relevant literature surveys~\cite{Atzori}\cite{Acharjya}\cite{6512846,Bellavista,7065282}\cite{7087016}\cite{BRao}\cite{6674155}, this paper provides a broader viewpoint on CPS from different aspects, mainly data collection and storage, processing, analytics, cybersecurity, and green solutions, rather than focusing on specific aspects of CPS.
\item Second, we provide a broader summary of security solutions proposed for big data CPS in data collection, storage, and access, as well as in data processing and analytics; unlike previous survey papers such as~\cite{Derbeko20161,SYI,Wang:2013:SCS:2459506.2459606} that only {focused} on security of a particular component or element of big data.
\item Third, we provide a summary in big data, storage, and communications for CPS.
\item Fourth, we provide a broader view of big data meeting green challenges for CPS, unlike~\cite{beloglazov2011a,7399696,7430133,7473821,7473815}.
\end{itemize}

\begin{table}[htbp!]
\centering
\caption{List of acronyms and their descriptions.}
\label{acronym}
\begin{tabular}
{|l|l|}
\hline
Acronym & Description                            \\
\hline
BLE     & Bluetooth low energy                   \\
CAPE    & Checkpointing aided parallel execution \\
CPS     & Cyber-physical systems                 \\
D2D     & Device-to-device                       \\
DVFS    & Dynamic voltage and frequency scaling  \\
GFS     & Google file system                     \\
GIS     & Geographical information systems       \\
HDFS    & Hadoop distributed file system         \\
IoT     & Internet of Things                     \\
KDD     & Knowledge discovery in databases       \\
M2M     & Machine-to-machine                     \\
MBAN    & Medical body area network              \\
MCS     & mobile crowd-sensing                   \\
MEMS    & Micro-electromechanical systems        \\
MIMO    & Multiple input multiple output         \\
MTC     & Machine-type communications            \\
NoSQL   & Not only structured query language     \\
RFID    & Radio-frequency identification         \\
RS      & Remote sensing                         \\
RSS     & received signal strength               \\
SDN     & Software defined network               \\
VM      & Virtual machine                        \\
QoC     & Quality of context \\
\hline
\end{tabular}
\end{table}

{The sections of this paper are organized as follows.} Section~\ref{related} compares this paper with other related surveys in literature. Section~\ref{data_sources} describes {the different sources, generations, and collections of big data for CPS,} mainly context-aware computing, communications, and social computing, while Section~\ref{CPS_app} presents different CPS-related big data applications. Then, in Section~\ref{storage}, we discuss about big data caching and routing for better CPS data management. In Section~\ref{processing}, we present the different big data processing tools from cloud data processing and multi-cloud processing to big data clustering techniques, NoSQL and fog computing that help facilitate the analysis of large volume of data. In Section~\ref{analysis}, an overview of big data analytics techniques and tools is provided. Section~\ref{security} provides a summary of the different security solutions proposed for CPS, mainly in storage, access and analysis. Section~\ref{green} addresses the relevances of big data and green challenges for CPS.
Finally, Section\ref{challenges} identifies CPS big data challenges and open issues. Conclusion remarks are presented in Section~\ref{conclusion}. A list of acronyms used in this paper is provided in Table~\ref{acronym}.

\section{Relevant Works}
\label{related}
Several survey papers have focused on specific aspects of big data, IoT, and CPS. However, to the best of our knowledge, none of them have {investigated the interconnections} of these concepts together in an extensive survey like this one.

In~\cite{Atzori}, the authors provided an overview on IoT from definitions to technologies to applications and standardizations, while presenting the different solutions proposed in literature for deterring security threats and preserving data integrity in RFID systems. In~\cite{6512846}, the authors {provided} a brief overview on IoT, with emphasis on sensor networks and their relationships to IoT. Then, context-aware computing definitions, categories and characteristics from an IoT perspective were thoroughly presented. In~\cite{Bellavista}, context data { were} surveyed for mobile ubiquitous environments, with the main focus being on providing efficient context data distribution for real network systems {via} highlighting context distribution architecture's layers, network deployments and taxonomy. In~\cite{7065282}, different mobile crowd sensing incentives and types were {reviewed} to motivate normal users to participate and contribute to different sensing applications. In~\cite{7087016}, caching big data techniques to optimize computational time and reduce storage overhead as well as improve system performance, scalability and efficiency were presented. In~\cite{6674155}, different cloud computing schedulers using Hadoop-MapReduce with their pros and cons were presented for purpose of improving scheduling in cloud environment. In~\cite{Acharjya}, big data analytics from challenges to solutions to open research issues were explored for efficient information extraction and decisions making.

In~\cite{Derbeko20161}, different security and privacy protocols for MapReduce were surveyed for integrity, correctness and confidentiality of cloud data computations. As for fog computing, which extends cloud computing to the edge of the networks, \cite{SYI} {reviewed} the different security and privacy challenges, solutions and open issues. For specific CPS applications, \cite{Wang:2013:SCS:2459506.2459606} provided a thorough survey on cybersecurity issues for smart grids, with emphasis on security requirements, network vulnerabilities, and secure communication protocols and architectures. Green challenges facing big data cloud computing and processing were surveyed in~\cite{beloglazov2011a,7399696,7473821,7473815}, with different green IoT applications in~\cite{7430133}.

CPS has also been surveyed in literature. For instance, in~\cite{6096958}, CPS features, energy control, transmission, resource allocation and software designs were presented. The work \cite{Chaari2016260} discussed the integration of cloud computing with CPS by categorizing them in three different areas: remote brain, big data manipulation, and virtualization. The paper \cite{gunes2014survey} {surveyed the general concepts, challenges and applications for CPS.}

Most of the previous surveys {have not discussed big data in the contexts of CPS,} which is the main focus of this paper. Furthermore, the green and security challenges and solutions are explored specifically for big data in CPS. Therefore, this survey paper provides a big umbrella addressing promising research {frontiers and insights in many challenges} and open issues facing big data meeting CPS. Table~\ref{survey_related} provides a summary comparison of the previously mentioned related surveys in the field.

\begin{table}[htbp!]
\centering
\caption{A summary of different related surveys.}
\label{survey_related}
\begin{tabular}{|
>{}l |
>{}l |}
\hline
Survey                                 & Scope                                                                                                                   \\ \hline
\cite{Atzori}                        & IoT definitions, technologies and applications                                                  \\ \hline
\cite{6512846}                       & IoT context-aware definitions, categories and characteristics                                                           \\ \hline
\cite{Bellavista}                    & Context data distribution for mobile ubiquitous environments                                                        \\ \hline
\cite{6096958}                       & \begin{tabular}[c]{@{}l@{}}CPS features, energy, transmission, resource allocation \\ and software designs\end{tabular} \\ \hline
\cite{Chaari2016260}                 & Cloud computing in CPS                                                                                                  \\ \hline
\cite{gunes2014survey}               & CPS general concepts, challenges and applications                                                                       \\ \hline
\cite{7065282}                       & Mobile crowd-sensing incentives                                                                                         \\ \hline
\cite{7087016}                       & Caching big data                                                                                                        \\ \hline
\cite{6674155}                       & Cloud computing schedulers                                                                                              \\ \hline
\cite{Acharjya}                      & Big data analytics challenges and solutions                                                                             \\ \hline
\cite{Derbeko20161}                  & Security and privacy protocols for MapReduce                                                                            \\ \hline
\cite{SYI}                           & Fog computing security challenges, solutions and open issues                                                            \\ \hline
\cite{Wang:2013:SCS:2459506.2459606} & Cybersecurity for smart grids                                                                                           \\ \hline
\cite{beloglazov2011a}               & Energy efficient cloud data centers and computing                                                                       \\ \hline
\cite{7399696}                       & Green big data processing for smart grids                                                                               \\
                               \hline
\cite{7430133}                       & Green IoT applications  \\

\hline
\cite{7473821,7473815}                       & Big data meet green challenges                                                                                                                                                           \\ \hline
\end{tabular}
\end{table}

\section{Sources, Generations, and Collections of Big Data for CPS}
\label{data_sources}
\begin{figure*}[htbp!]
  \begin{center}
    \includegraphics[scale = 0.36]{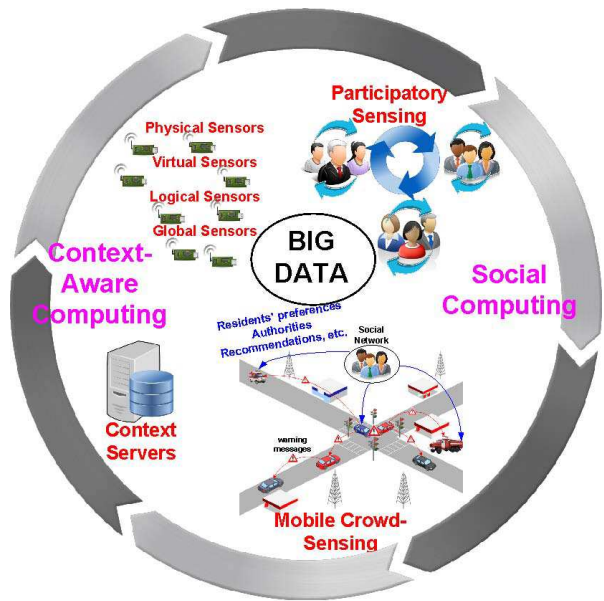}
    \vspace{-0.2cm}
    \caption{\label{types} An illustration of the different CPS big data sources and types.}
  \end{center}
\end{figure*}
Several factors have driven the expansive {uses} of sensors, mainly {for} micro-electromechanical systems (MEMS) such as accessible design boards (Raspberry pi, Onion, Arduino)~\cite{nugroho2016development}, and the new efficient hardware architectures and components, which made sensors more robust to hardware wearing from harsh environments. Moreover, many of these sensors incorporate accelerometers that are 1000x more powerful in terms of sensitivity than those used in a Nintendo Wii~\cite{Akdeniz:2000:ILS}. In this section, we discuss the different sensed big data sources and types by grouping them into social computing and context-aware computing. An illustration is provided in Fig~\ref{types}.
\subsection{Context Aware Computing and Communications}
The general definition of context-aware communications and networking (CACN) was provided in \cite{6829939}. Context-aware computing could be considered as CACN operating in the higher networking layers. The data that sensors collect for its specific application purposes {are considered raw data, which have been} directly collected from the environment without further processing. With raw data alone, it becomes challenging to analyze and interpret {them, and} let alone the big data generated by the large scale deployment of sensors. For data {which provide relevant information that is meaningful} and easily interpretable, sensors need to engage in context-aware computing; that is, sensors need to store processed meaningful information, also known as ``context information'', that is easily understandable~\cite{10.1109/MOBIQ.2006.340411,6829939}. An example to highlight the difference between raw sensor data and context information would be the blood sugar readings collected by bio-medical sensors on {the bodies of patients,} which are considered raw data. When these readings are processed and represented as a patient's average glucose level in the blood, they are referred to as context information. The quality of context (QoC) metric is used to assess the quality, validity, precision and {up-to-date context} information~\cite{Bellavista}. Context-aware computing from acquisition, processing, storing and reasoning, can be performed by the applications themselves, or by using libraries and toolkits, or even by using a middleware platform~\cite{6512846}.

As for sources of context information, they can be retrieved directly from physical sensors, from virtual sensors (they collect data from different sources by using web services technology and represent it as sensor data), from logical sensors ( a combination of physical and virtual sensors), from a middleware infrastructure (global sensor networks), from context servers (databases, web services), or even manually provided such as retrieving users' preferences. The context information can be further classified into primary context and secondary context, which provides information on how the data {were} obtained. For example, reading RFID tags directly from different production parts in industrial plants is considered {the} primary context, while obtaining the same information from the plant's database is referred to as {the} secondary context~\cite{6512846,Atzori}.

\subsection{Remote sensing}
Collecting sensor data of objects from a distance is referred to as remote sensing (RS)~\cite{doi:10.1080/01431168008948242}. RS is an integral part of earth sciences. For instance, space-borne and airborne sensors collect multi-spatial and multi-temporal RS big data from the global atmosphere for purposes of earth observation and climate monitoring~\cite{Ma:2015:RSB:2794090.2794274}. Other remote sensing applications include Google Earth that provides pictures of the earth's surface, weather reporting, traffic monitoring, hydrology and oceanography~\cite{7486259}.

In~\cite{7109130}, the authors proposed a big data analytical architecture for real-time RS data processing using earth observatory system. The real-time processing includes filtration, load balancing and parallel processing of the useful RS data. The RS datasets are normally geographically distributed across several data centers, leading to difficulties in loading, scheduling and transmission of data. Moreover, the high dimensionality of the RS data makes their storage and data access rather complicated~\cite{Ma:2015:RSB:2794090.2794274}. That is why{,} in~\cite{6897952}, Wang \textit{et al.} proposed a wavelet transform  to represent RS big data by decomposing the datasets into multiscale detail coefficients, which are estimated using expectation-maximization likelihood. In~\cite{7128315}, the authors evaluated the quality of RS data using statistical inference via using the prior knowledge of the dataset to get an unbiased estimator for the quality.
\subsection{Social Computing}
With the explosive increase in smartphones usage, mobile data {have experienced} an unprecedented growth, carrying enormous amount of information on user applications, network performance data, service characteristics, geographic information, subscriber's profile, and so on~\cite{7474340}. This has led to shaping the notion of ``mobile big data'', which, unlike traditional big data in computer networks,  {have their own unique characteristics.} One of these characteristics is the ability to partition mobile data in time and space domains, such as in minutes, hours, days, location, and so on. Furthermore, due to the features of smartphones' usage, the same traffic, on one hand, can be highly likely requested by a group of subscribers in certain time and location; and on the other hand, subscribers in close proximity may exhibit similar behavior and mobility patterns, all of which can help optimize network performance~\cite{7562344}.

Social computing allows the integration of these social behaviors and contexts into web technologies to assist with predicting social dynamics, which can render the operation, planning and maintenance of social wireless networks easier than ever~\cite{7302079,parameswaran2007social}. For instance, due to high social correlations and relationships among subscribers, a user social network can be formed, in which the habit, interests, mobility, and sharing patterns can be used to construct social community structures and analyze communication behaviors. One such an example of user social application is the popular \textit{Pokemon Go} game, where users in close proximity share real-time maps to hunt for Pokemon characters~\cite{7562344}. Another example where social computing can be beneficial is in emergency situations, such as the spread of infectious diseases, where taking the appropriate policies by analyzing human interactions and predicting the emergency's evolution can help protect the public health~\cite{7302079}.  This article \cite{J_FGCS_HNing_0_2016} introduced Cybermatics as a broader vision of the IoT (called hyper IoT) to address science and technology issues in the heterogeneous cyber–physical–social–thinking (CPST) hyperspace. Next, we list two different social computing tools for data collection, mainly, participatory sensing and crowd-sensing.

\subsubsection{Participatory sensing}
Participatory sensing or community sensing allows users to collect and share information either within social groups (social sensing) or with everyone (public sensing) using their own smart devices~\cite{7470600,7406686}. This means that sensors can be substituted by users for purpose of data collection, which can significantly reduce the monetary costs of deploying physical sensors. However, with participatory sensing comes several challenges such as the quality and trustworthiness of collected data, the willingness of participants to engage in the sensing tasks and protecting participants' personal information.

In~\cite{7470600}, the authors proposed a participant coordination architecture that selects the most efficient  participants without exposing participants' personal information to the application server. To protect participants' privacy, in~\cite{7097702}, Chang \textit{et al.} proposed a secure scheme called PURE which allows participants to reach the global model {to} estimate via peer reviewing the local regression models. This enables participants to only report intermediate results back to the server without the need of sharing local private data with the server. In~\cite{7565025}, Messaoud \textit{et al.} proposed a mobile sensing scheme that reduces the sensing time required by participants, and increases the fairness of sensing tasks assignment to ensure participants' commitment to sensing while maintaining same data quality as in non-fair schemes. In an attempt to maximize the overall data quality, in~\cite{7563897}, Wang \textit{et al.} proposed a multi-task allocation framework (MTPS) which pays participants a compensation from a shared budget for each sensing task, with additional compensation if a participant is assigned more than one task. This greedy framework allows the allocation of multiple tasks to participants. In~\cite{7044580}, participatory sensing for environmental data collection was used, where the urban resolution metric was used to measure the quality of urban sensing. In~\cite{7337442}, participatory sensing {was} applied to vehicular networks, where location, speed, and fuel consumption of vehicles can be communicated {with} the server through phones aboard via a WiFi interface to reduce data transfer delay time.

\subsubsection{Mobile crowd-sensing}
Mobile crowd-sensing (MCS) can be considered as an extension to participatory sensing. In addition to collecting data from mobile devices (mobile sensing), MCS uses social sensing {via} integrating and fusing the contributed data from mobile devices with that of the mobile social network services in order to provide solutions to more complex queries~\cite{7402272}. In vehicular networks, with participatory sensing{,} we can collect warning messages from vehicles to determine the traffic status. However, if in addition, a driver needs to know whether the route is safe to drive on based on authorities' recommendations, residents' preferences, {and so on,} then MCS can be useful (see Fig.~\ref{types}).

In~\cite{7355365}, Xiang \textit{et al.} used MCS to construct accurate outdoor received signal strength (RSS) maps using error-prone smartphones. In~\cite{7110391}, Wang \textit{et al.} proposed an energy-efficient cost-effective data uploading in MCS {via} providing incentives to participants to use the appropriate timing and network to upload the data. Data was offloaded to Bluetooth/WiFi gateways {via} using predictions on users' calls and mobility. To maintain relatively good performance of MCS applications, {a} sufficient number of participants need to {make contributions} to sensing. In~\cite{7065282}, the authors discussed about the different incentives for MCS from entertainment, service and monetary incentives, in which participants can be recruited in multiple sensing tasks.

\section{Big Data Storage and Communications}
\label{storage}
Caching CPS big data can lead to a reduction in the amount of traffic exchanged, which contributes to a better data management, lower latency and energy consumption. In this section, we discuss some of the solutions proposed for CPS big data caching and routing.
\subsection{Big Data Caching and Storage for CPS}
The high amount of traffic driven by the popular use of mobile video and online social media applications along with the scarcity of backhaul resources have pushed researchers and mobile operators to find solutions. Content caching in CPS is of high interest, especially that a big proportion of the traffic load originates from fetching data from different sources such as databases, cache servers and network gateways~\cite{7565185}. Rather than caching data from the cloud, performing the caching at the edge of mobile wireless networks, such as base stations and user equipments, offers the advantage of better data management~\cite{7150324,7524427}. For instance, in~\cite{7565185}, Zeydan \textit{et al.} proposed a big data proactive caching architecture that predicts popular content from users' behavior and network characteristics to perform caching at the base stations. This has the advantage of backhaul offloading to the edge, so that data {get} closer to users, thereby enhancing users' quality of experience and reducing latency. The proposed architecture is validated using dataset from a Turkish mobile operator, where it was shown that proactive caching can yield 100\% user satisfaction by offloading 98\% of the backhaul to the edge. The proactive caching at the edge is envisioned to solve big data management in future 5G networks, especially that base stations densification and acquiring new spectrum do not seem quite effective in terms of cost, scalability and flexibility~\cite{7087016,7565185}. A new approach to reducing network traffic in telehealth systems was suggested in~\cite{7336366}, where a filter is used inside the sensors which associates a scale to each record, to determine the data fields that need to be sent to the server.

Different works have been proposed to deal with challenges facing Hadoop for handling big data storage and processing~\cite{7284356,7492645,6733207}. For instance, in~\cite{7492645}, the authors proposed a novel cache to store the intermediate data, which helps eliminate redundancy in storage and processing of the big data set, in addition to speeding up the performance of the system by fetching the data from the cache rather than running mapper functions. Dache, a data-aware cache framework for big-data applications, was suggested in~\cite{6733207} to accelerate the execution of MapReduce tasks. {The transmitters of the tasks} send their intermediate data to a cache manager, which is then used to fetch potential processing results.

Cache memory can help speed up CPS communications via increasing the execution speed and decreasing the time spent on memory access to fetch the required data. For instance, in medical CPS, a timely response from medical sensors to servers' requests is required, where caching can be very useful. However, as mentioned in~\cite{suh2014applied}, cache misses are highly likely in CPS, especially when a task is interrupted by a higher priority task, triggering a cache interference cost. A potential solution to this problem is cache partitioning where different tasks with shared resources are isolated and assigned reserved partitions of a cache memory. This can be very useful for real-time CPS applications such as mobile-health, environmental monitoring, traffic surveillance, aerospace and so on, where reliability and predictability are of high importance. Moreover, with the large-scale network of CPS, {temporal interferences from a} large number of devices contending on shared resources, such as processor cores, buses, I/O devices, can significantly reduce the runtime of CPS and lead to unpredictable systems. Several algorithms and component-based approaches have been proposed to help reduce the temporal {interferences} of CPS~\cite{Wahler:2015:RMC:2737166.2737176,5763145,herkersdorf2013multicore}.

\subsection{Big Data Communications for CPS}
To accommodate the big data environment for CPS, building a resilient network infrastructure for the data-intensive applications and services has become essential~\cite{7553025}. The data collections, the data chunks distribution to data centers and the data delivery to intended users, necessitate a fast and reliable networking to bridge these stages together. For instance, machine-type communications (MTC), a class of technologies related to IoT and defined by the 3rd Generation Partnership Project (3GPP), are becoming more popular as we move towards a smart city. In an attempt to make homes smarter, Apple introduced the HomeKit, which uses Siri to control different things inside the home remotely using the iPhone, iPad and even AppleTV~\cite{Mone:2014:IL:2692965.2676393}. Samsung, on the other hand, introduced SmartThings that allow the automation of many home tasks using WiFi, Zigbee, Bluetooth low energy (BLE), and Z-Wave~\cite{tibken2015samsung}. These new technologies allow all the devices to be interconnected and to be communicated together.  Many of the MTC communications are envisioned to run over current cellular networks~\cite{3GPP}, providing MTC devices with ubiquitous coverage, global connectivity, reliability and security. However, this creates a set of challenges for cellular network providers~\cite{6845044}, such as the inability to handle MTC traffic on networks optimized for human communications, excessive congestions due to signaling overhead, packet scheduling problems due to MTC traffic requiring a number of radio resources below the minimum allocated to a cellular device;, and the large interferences generated from MTC devices~\cite{6678832}. Different relevant works, such as~\cite{7343681,7248779,6906487,8241868}, suggested offloading the MTC traffic onto device-to-device (D2D) communications links. As a matter of fact, Google OnHub router can support direct D2D communications, as well as WiFi, BTL and 802.15.4 using Brillo as a stripped OS version of Android. The D2D technology provides an ideal solution to support the massive communications of MTC devices, especially that these devices are anticipated to be located close to each other~\cite{7051287}. The paper\cite{J_ISJ_CZhu_09_2016} discussed the integration of wireless sensor networks (WSNs) and mobile cloud computing (MCC) and proposed a sensory data processing framework to transmit desirable sensory data to the mobile users in a fast, reliable, and secure manner.

In~\cite{7562084}, the authors proposed two different sustainable routing designs called SustainMe. The first model uses a dedicated backup protection in which network components are turned off after data is fully delivered to help save energy. The second model uses shared backup protection to achieve a trade-off between energy efficiency and capacity usage efficiency. The latter is shown to consume less capacity than the the first model, but at the expense of an increase in energy expenditure.  In~\cite{7417577}, the authors proposed a software defined network (SDN) routing for Hadoop to accelerate the speed of big data delivery through speeding up the MapReduce data shuffling. This is useful for time-sensitive big data applications. As a matter of fact, SDN can significantly improve big data applications, especially {which} separates the control and data planes, allowing the control plane to be logically centralized so that decisions and network optimization are performed efficiently {via} having a global view of the network. All these SDN's features can make the big data acquisition, processing, transmission and storage much easier and faster. On the other hand, big data can also assist in efficiently designing and optimizing SDN functions through traffic engineering, cross-layer design, security and inter- and intra-data center communications~\cite{7389832}.

In~\cite{7553025,6659463}, taking advantage of social and geographical community structures, routing, navigation services, and data delivery performance can all be improved. In~\cite{7473821}, different big data routing algorithms for energy efficiency {were reviewed}, such as selecting routing paths for data centers that achieve the lowest energy costs, parallelizing data transfers by using multiple cores in data routing, and balancing big data traffic through a distributed adaptive routing algorithm that minimizes packet delivery.

In massive high-density environment, data collections and transmissions can be challenging tasks especially with the different data flow information, quantities and characteristics. In~\cite{7491630}, cyclists' data collections and transmissions were studied. Autonomous data collection using automated video analysis can be performed in order to overcome the need for costly manual data collection. For instance, a computer vision system can group users' features based on spatial proximity data, speed, trajectory and others. The obtained information can be input to an algorithm that obtains valuable information such as traffic stream, number of cyclists, lane density and so on.

In high-density environment, it is important for data to be efficiently collected in a short period of time to ensure up-to-date information. For this purpose, concurrent data collection trees can be valuable where multiple data collection streams can be initiated by different users on same set of devices~\cite{7569062}. This can be extremely helpful for  delay-sensitive applications. In~\cite{6330525}, the authors proposed concurrent massive data transmission for remote real-time health monitoring system. An input/output(I/O) Completion Port (IOCP) main server dynamically binds Internet protocol (IP) addresses to massive sensors before they can send their data. The main server assigns one of the IOCP servers with the least number of concurrent connections to micro sensors using  Oracle  Real Application Cluster (RAC). Finally, data storage is separated from Oracle server instances using Network Attached Storage (NAS) technology to allow for greater I/O performance for massive concurrent connections. The paper \cite{chao2016distribution} firstly analyzed the objective of massive access control for machine-type communications (MTC) or  machine-to-machine communications (M2M), another terms of CPS used in 3GPP standardizations for wireless cellular networks, and proposed distribution reshaping to enable data arrival distribution with highly coordinated manner to notably reduce multiple access congestions.

{Real-time data fusions are important for heterogeneous IoT/CPS data streams and unreliable networks with increasing data size. To support locally available computational values to help real-time analytics, the work \cite{J_IA_SJab_04_2018} proposed the fusion of three different data models with relational, semantical, and big data based data and metadata. The paper \cite{J_IA_AAkb_02_2018} proposed two-layer architecture for IoT data analysis, where the first layer is with the service oriented gateway based generic interface to obtain data from multiple interfaces and IoT systems and store them scalably and make relevant real-time analysis to extract high-level activities, while the second layer works for the probabilistic fusion of these high-level activities.}

\subsection{Summary and Insights}
To summarize, different strategies can be employed to enhance CPS performance {via} speeding up data collection, processing and distribution.
At the CPS traffic level, data collection and processing can be accelerated through caching at the edge,
filtering the data to make it more manageable in size, caching execution tasks, and using cloudlets whenever possible.
At the CPS devices level, the use of on-fly D2D, SDN, backup protection, and social/geographical community structures along
with navigation services can help accelerate data delivery and reduce latency. The use of these technologies and solutions
 can significantly speed up data handling while at the same time extending the communication range of CPS traffic.

\section{Big Data Processing for CPS}
\label{processing}
Cloud computing along with data clustering facilitates the parallel processing and execution of tasks and queries. Mapping and scheduling workflows in a multi-cloud environment speeds up the processing and allows for a better big data management. In this section, we discuss cloud computing, big data clustering, NoSQL and fog computing for big data workflows processing.
\subsection{Cloud Data Processing}
With the large volume of data in the order of exabyte, it becomes almost impractical to process the data on individual machines, no matter how powerful they are. Parallel processing of the data chunks on dedicated servers, such as MapReduce tool proposed by Google, offers advantages over conventional processing methods; however it is still not very effective {to handle a} large amount of data, mainly due to scalability, latency, availability, and inefficient programming techniques, including but not limited to database management systems~\cite{6415919,Dewitt}. One attractive solution to dedicated servers is the processing on cloud centers, which offers users the ability to rent computing and storage resources in a pay-as-you-go manner~\cite{7293302}. In  addition, even though users will be sharing a common hardware, the shared resources appear exclusive to them through machine virtualization {via} hiding the platform details~\cite{Armbrust:2010:VCC:1721654.1721672}. However, this approach can create problems in the pay-as-you-go environment due to untruthfulness, unfairness and inefficiency of resources and workload transactions~\cite{7037728}.

The main difference between parallelizing tools, such as MapReduce and cloud computing, is that MapReduce uses mappers and reducers to produce intermediate results and final results, respectively. However, public clouds offers users with virtual machines (VMs) with a highly elastic resource allocation~\cite{7293302}. The function of Map is in charge of processing input key-value pairs and generating intermediate key-value pairs; while Reduce function is used to further compress the value set into a smaller set based on the intermediate values with the same keys~\cite{Blanas:2010:CJA:1807167.1807273}. To maximize the query rate of remotely located data in an attempt to maximize system performance, the authors in~\cite{7524388} designed a dynamic resource allocation algorithm that takes into account the computations of query streams across the nodes and the limiting number of resources available.

Due to the volume and velocity characteristics of big data, streaming data processing and storage might require different compression techniques to ensure efficiency and scalability. Yang \textit{et al.} proposed a novel low data accuracy loss compression technique for cloud data processing and storage. A similarity check was performed on partitioned data chunks and a compression is conducted over the data chunks rather than the basic data units~\cite{7412709}. Another similarity check-based compression technique was proposed in~\cite{7066939} using weighted fast compression distance.
\begin{figure}[htbp!]
  \begin{center}
    \includegraphics[scale = 0.3]{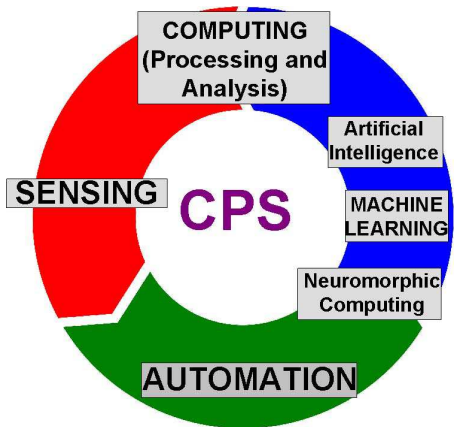}
    \vspace{-0.2cm}
    \caption{\label{cps} CPS cycle to automation.}
  \end{center}
\end{figure}

Integrating cloud computing in IoT can take the processing of sensing data streams to the next level to provide ubiquitous sensing services beyond the capacities of individual things. When combined with artificial intelligence, machine learning, and neuromorphic computing techniques, it is envisioned that new applications will be developed with automated decision making, which would revolutionize the field of smart cities, industrial plants, environmental monitoring and others (see Fig.~\ref{cps}). Cloud computing will enable IoT applications to have a reduced latency, power consumption and enhanced scalability. Examples of such applications include but not limited to healthcare, where patients' information can be accessed using a cloudlets-based infrastructure~\cite{7576619}. Cloudlets are clouds that are closer to users to help overcome the high latency and power consumption of distant clouds~\cite{7295892}. Other cloud application examples include vehicles traffic control system~\cite{7578484}, genome analysis~\cite{7364117}, earth surface analysis~\cite{7116525} and many others.

\subsection{Multi-Cloud Data Processing}
In many IoT scientific applications, the data collection and generation, computation, processing and analysis are broken down into workflows, consisting of interdepending computing entities. Due to the data-intensive nature of IoT applications, the large-scale workflows need to be distributed across multiple cloud centers~\cite{7557446}. To allow the support of multiple applications and to overcome the limitations of current frameworks that are dedicated to a unique type of applications, the authors in~\cite{7049894} proposed a distributed application management framework in multi-cloud environment by using a domain specific language (DSL) to describe applications in a hierarchical manner. However, the inter-cloud communications constitute a big deal of the financial costs of processing workflows due to their large volume. In~\cite{7557446}, to optimize system performance, Wu \textit{et al.} proposed a budget-constrained workflow mapping in multi-cloud environment. An efficient pay-on-demand pricing strategy for streaming big data processing in multi-cloud environment was proposed in~\cite{7383293}, to offer a low price for data load processing, while maximizing the revenues of cloud service providers. As for building trust across multiple cloud centers that collaborate together on data storage and processing, Li \textit{et al.} in~\cite{7236926} proposed a trust-aware monitoring architecture between users and cloud centers with hierarchical feedback mechanism to enhance the robustness and reliability of the quality of service of cloud providers, which helps provide them with a rating based on their trust reputation. This allows users to select services from different cloud providers based on feedback and past service records.

In~\cite{7353181}, Wang \textit{et al.} optimized virtual machine (VM) placement in national cloud data centers to help minimize the energy consumption of data-intensive services, such as planet analysis. In these cloud centers, VMs are assigned to physical servers, which helps provide users with a high quality of service but at the expense of greater energy consumption. The trade-off between energy consumption and quality of service is taken into consideration in the optimization problem. The cost of data access and storage limitations of national cloud centers was considered in~\cite{7590316}, where the authors used Lagrangian relaxation based heuristics algorithm to obtain the optimal data centers placement that can reduce data access costs.

The distribution of users' tasks on geographically distributed cloud centers was addressed in~\cite{7225163}, where the authors proposed a big data management solution to maximize system throughput such that fairness of limited resources usage by users is guaranteed and the operational cost of service providers is reduced. To allocate resources of multi-cloud centers to users, a multi-round combinational double auction based mechanism was proposed in~\cite{7510724}, where auctions on different VMs were performed by both users and data centers in multiple rounds to maintain a high quality of service level.

\subsection{Big Data Clustering}
\label{clustering}
\begin{figure}[htbp!]
  \begin{center}
    \includegraphics[width =3in]{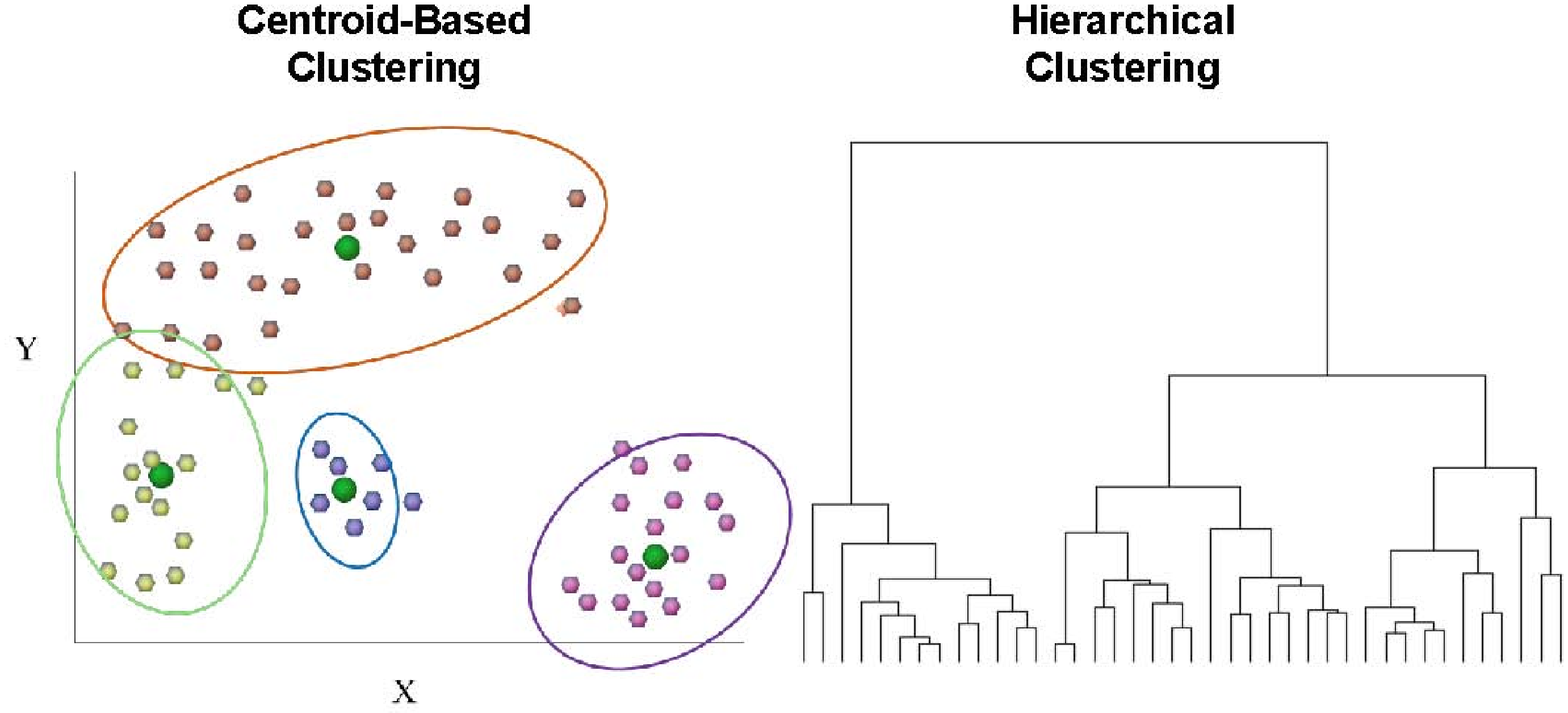}
    \vspace{-0.2cm}
    \caption{\label{clust} An illustration comparing centroid-based clustering and hierarchical clustering (adapted from~\cite{Visuali} and~\cite{centro}).}
  \end{center}
\end{figure}

Data clustering refers to partitioning a set of objects comprising of attributes into different groups of similar objects and features~\cite{7283582}. Data clustering becomes very useful in big data applications, where there is a high need to process and analyze large volume of data. Estimating the number of clusters becomes important as clustering facilitates the distribution of the data storage, tasks execution, parallel computing, and queries requests~\cite{7546221}. Fig.~\ref{clust} shows two groups of clustering methods researched in depth in literature: i) hierarchical clustering, and ii) centroid-based clustering. In hierarchical clustering, nearby objects have higher probability of being grouped together than far away objects. On the other hand, in centroid-based clustering, objects closer to the cluster center are grouped together. The paper \cite{J_TII_YZhao_n1_2018} discussed a flexible multiple clustering analytic and service framework, and a novel tensor-based multiple clusterings (TMC) approach.

One of the most popular centroid-based clustering is k-means due to its computational efficiency and low-complexity implementation. However, as the number of clusters increases, k-means clustering suffers from the empty clustering problem and the increase in number of iterations for convergence. This means that traditional k-means is not suitable for big data applications. Different works have suggested enhanced versions of k-means clustering for purposes of improving clustering quality, execution time, and accuracy. For instance, in~\cite{7456910}, the authors used an enhanced version of the k-means clustering, where the initial centroids of the cluster are not selected randomly but based on averaging the data points. This achieves higher accuracy than conventional k-means. Another enhanced version of k-means clustering was suggested in~\cite{7363909}, to help eliminate the empty clustering problem of traditional k-means. The clustering approach was based on a combination of Fireworks and Cuckoo-search algorithms with representative points being selected as the centroids. In~\cite{7207256}, the authors also used a centroid calculation heuristics to help enhance the clustering performance as the number of clusters increases.

In~\cite{7283582}, for instance, the authors used hierarchical clustering in their proposed algorithm, where the number of clusters was estimated visually based on a reordered distance matrix. The data samples were clustered using single linkage (SL), which cuts large edges of the clustering tree in a minimum spanning tree (MST). The algorithm was shown to be superior in terms of clustering speed and accuracy when compared to other clustering algorithms like k-means. Another hierarchical clustering was suggested in~\cite{7207373}, where users themselves define the number of clusters based on different similarity measures such as homogeneity and the relative population of each cluster. This helps improve user satisfaction of the clustering algorithm. A hierarchical k-means clustering algorithm was proposed in~\cite{7272099} to find high quality initial centroids. The centroids were obtained based on a hierarchical structure of k-means that consists of several levels, where the first level is the original dataset. Subsequent levels are compromised of smaller size datasets that consist of similar patterns as the original dataset. Fuzzy clustering, another clustering technique, is similar to k-means; however an object can be associated with more than a single cluster depending on its degrees of membership that are usually calculated based on Euclidean distances between the object and the data center~\cite{7555984}.

Another big data clustering is clustering features (CF), where a CF-tree includes a summary of data patterns in a cluster. It uses a threshold preset by users to set the size of micro clusters in a CF-tree. However, this method can have time and scalability complexities, especially with the large number of data that needs to be scanned and assessed to construct the tree. Clustering based on summary statistics allows the threshold to be different for different micro clusters by dynamically adjusted it based on regions' densities~\cite{7473093}. Graph clustering attempts to cluster data points based on network's structure and nodes' connections, where similarities between nodes allow them to be connected to build a community. An example would be building a Tweet graph for Twitter online social application where Tweets (nodes) with similarities (similar URL count, similar hashtag count, or similar username count) are connected together with edges~\cite{7359276}. Finally, incremental mining helps deal with the increasing network size and data growth challenges by incrementally updating the clusters without the need for reconstructing them from scratch.

In the context of CPS applications, \cite{5319232} attempted to perform structuralized clustering for sensor-based CPS for purposes of reducing the energy consumption of the sensor network. The sensor network is basically structured into small equal units, each consisting of a set of clusters, whose number is carefully selected based on the cluster head workload and its distance to the base station to achieve energy efficiency. In~\cite{6863675}, the authors tried to solve the cascaded subnetworks' failures problem in CPS by obtaining an upper bound on the small clusters' size to mitigate networks' failure, especially when they are tightly coupled. In~\cite{Bali2016476}, the authors proposed a secure clustering method for vehicular CPS, where a trust metric is calculated for each vehicle based on their transmission characteristics in order to create secure clusters. Another secure clustering for vehicular CPS was proposed in~\cite{huo2017coalition}, where the clustering problem is formulated as a coalition game taking into consideration the relative velocity, position and bandwidth of vehicles. Furthermore, incentives and penalty mechanisms are suggested to prevent selfish nodes from degrading the communication quality performance. A density-based stream data clustering for real-time monitoring CPS applications was suggested in~\cite{SPEZZANO20151016}, where the authors used FlockStream algorithm to group similar data streams. Each data point is associated with an agent,
and similar agents within a visibility range of each other in the virtual space, form a flock allowing for real-time stream clustering.
A summary of the different big data clustering techniques for CPS is provided in Table~\ref{tableclust}.

\begin{table*}[htbp]
\centering
\caption{Summary of big data clustering techniques for CPS}
\label{tableclust}
\begin{tabular}{l|l|l|}
\cline{2-3}
                                                                        & Relevant Research Works                                                                                                                                                                                                                                      & Big Data Clustering Techniques                                                                                                                                                                                \\ \hline
\multicolumn{1}{|l|}{Hierarchical clustering}   & \begin{tabular}[c]{@{}l@{}}\cite{7207373}, \cite{6967763}, \cite{7558012}\\\hline\\ \cite{7272099}, \cite{6960906}\\\hline\\ \cite{7555984}, \cite{7000907}, \cite{7292782}\\\hline\\ \cite{7473093}\\\hline\\ \cite{7743854}, \cite{7511386}\end{tabular} & \begin{tabular}[c]{@{}l@{}}Clustering based on similarity measures\\\hline\\ Hierarchical k-means solutions\\\hline\\ Fuzzy clustering\\\hline\\ Clustering based on summary statistics\\\hline\\ Incremental mining\end{tabular} \\ \hline
\multicolumn{1}{|l|}{Centroid-based clustering} & \begin{tabular}[c]{@{}l@{}}\cite{7283582}, \cite{7578325}\\\hline\\ \cite{6740862}\\\hline\\ \cite{7359276}, \cite{7551774}\end{tabular}                                                                                                                     & \begin{tabular}[c]{@{}l@{}}Minimum Spanning Trees\\\hline\\ Consensus clustering statistics\\\hline\\ Graph clustering/Spectral clustering\end{tabular}                                                                 \\ \hline
\end{tabular}
\end{table*}

\subsection{NoSQL}
Conventional relational database management systems are not suitable for heterogeneous big data processing, as they consist of strict data model of pre-defined data structures and constraints with a fixed schema~\cite{7279070}. NoSQL (Not Only Structured Query Language) relaxes many of the relational databases' properties such as ACID transactional properties to allow for greater querying flexibility, operational scalability and simplicity, higher availability and faster read/write operations of unstructured big data through replicating and partitioning the data across several nodes~\cite{Cattell:2011:SSN:1978915.1978919,7452296}.

NoSQL databases can store data in three different forms: key-value stores, document databases, and column-oriented databases~\cite{strauch2011nosql}. In the document databases form, the data is stored in a complex structure form such as XML documents. Column-oriented databases store columns of data in data tables, allowing greater ease of adding and deleting columns compared to row-oriented databases.

Furthermore, two main classes for NoSQL systems:  operational NoSQL systems (Cassandra, MongoDB, Oracle NoSQL), and analytical NoSQL systems which are based on MapReduce, Hadoop, and Spark. Operational NoSQL systems include  online transaction processing (OLTP) systems, while analytical NoSQL systems include decision support systems (DSSs). The main difference between OLTP and DSSs is that the latter involves processing over larger tables and hence, involves complex queries processing (scanning, joining, and aggregating), while OLTP performs read/write operations for a smaller number of entries in the database~\cite{7452296}. For instance, in~\cite{7584920}, the authors used NoSQL for big data workflows execution to improve the scalability, parallelism and execution compared to traditional MapReduce framework.

\begin{table*}[htbp!]
\centering
\caption{Comparisons among different parallel processing techniques}
\label{processing_summary}
\begin{tabular}{
l |
l |
l |}
\cline{2-3}
& Advantages &Disadvantages                                                                                                                                                                                                                   \\ \hline
\multicolumn{1}{|l|}{MapReduce parallel processing}    & dedicated resources to users                                                                                                                                                 & inefficient at handling large data                                                                                                                                                                                              \\ \hline
\multicolumn{1}{|l|}{Single cloud parallel processing} & \begin{tabular}[c]{@{}l@{}}i) scalability and availability in handling\\ big data; ii) elastic resource allocation\end{tabular}                                                     & \begin{tabular}[c]{@{}l@{}}i) shared resources; ii) users need to\\ pay to rent VMs; iii) untruthfulness\\ and unfairness in using resources\end{tabular}                                                                                   \\ \hline
\multicolumn{1}{|l|}{Multi-cloud parallel processing}  & \begin{tabular}[c]{@{}l@{}}i) Better quality of service; ii) overcomes \\ storage limitations; iii) maximizes system\\ throughput; iv) achieves fairness of limited\\ resources\end{tabular} & \begin{tabular}[c]{@{}l@{}}i) requires VMs placement\\ optimization; ii) might be costly to\\ implement; iii) increases energy \\ consumption, iv) complicates data \\ access and management; v) requires\\ robust security solutions\end{tabular} \\ \hline
\end{tabular}
\end{table*}

\subsection{Fog Data Computing}
Cloud computing and services can be extended to the edge of network via the fog computing paradigm. Though both cloud and fog provide data computations, storage and application services to end-users, the distinguished features of fog from cloud include its proximity to end-users, the dense geographical distribution and its mobility support~\cite{Bonomi}.
These features facilitate fog computing for latency-sensitive applications. Consider an example~\cite{FedCSIS2014503} of smart traffic lights and connected vehicles, where an ambulance flashing lights sensed by a video camera could automatically trigger street lights to open lanes for the ambulance to pass through the traffic. In smart grid scenario~\cite{Bonomi}, the fog devices at the edge collect the data generated by grid sensors, real-time process the data, and issue control commands to the actuators.  A fog-based intelligent decision support system was proposed in~\cite{Roy} for driver safety and traffic violation monitoring based on the IoT. A smart city speeding traffic surveillance scheme using fog computing paradigm was provided in~\cite{7545004}, and its effectiveness was validated by intensive experiments conducted using real-world traffic surveillance video streams. Fog computing could make possible the early predictions of biomarkers to enable automated decisions making in a connected health scenario~\cite{7255196}. A fog computing system for e-Health applications was implemented in~\cite{7421170}, where fog nodes were installed in home to achieve the smallest processing time. Disaster decision support systems, deployed with fog nodes, could process acquired real data and trigger alarms in case of an emergency~\cite{BRZOZAWOCH20152387}. An augmented brain computer interaction game based on fog computing and linked data was developed in~\cite{6910482}. Other exemplary scenarios~\cite{FedCSIS2014503} for fog computing could be wireless sensor and actuator networks, decentralized smart building control, software defined networks, and so on.

Beyond the latency-sensitive applications, fog computing could also offload the core network traffic and keep the sensitive data inside the network~\cite{cisco}. In~\cite{7363093}, fog computing at smart gateways  was shown to enhance the health monitoring system through the usage of advanced techniques such as embedded data mining, distributed storage and notification service at the edge of network. In~\cite{Hong:2013:MFP:2491266.2491270}, the authors proposed a high level programming model, called mobile fog, for future Internet applications that are geospatially distributed, large-scale and latency-sensitive. The placement and migration method for infrastructure providers was proposed in~\cite{Ottenwalder:2013:MOM:2488222.2488265} to incorporate cloud and fog resources. Specifically, the network intensive operators were placed on distributed fog devices while computationally intensive operators were placed in the cloud. Some design goals and challenges in fog computing platform were described in~\cite{7372286}, which also suggested several important components of a fog computing platform, i.e., authentication and authorization, offloading management, location services, system monitor, resource management, virtual machine scheduling.

The security and privacy issues were examined in the context of smart grids~\cite{Wang:2013:SCS:2459506.2459606} and machine-to-machine communications~\cite{5741143}, and so on. These security solutions for cloud computing may not be directly applicable to fog computing, as fog devices are placed at the edge of networks. One security issue of fog computing is the authentication at different levels of gateways. The compromise of gateways serving as fog devices could lead to the Man-in-the-Middle attack~\cite{5471501}. Since a certain amount of fog networks are connected through wireless, typical wireless attacks (e.g., jamming attacks, sniffer attacks, and so on) could be possible threats for fog computing~\cite{SYI}. Furthermore, the leakage of private information, such as data, location or usage could be the main concerns of end users.

\subsection{Summary and Insights}
From Table~\ref{processing_summary}, we can observe that users' workloads can be more efficiently processed in a multi-cloud environment, especially that most CPS applications deal with a large volume of data. However, such a solution requires additional optimization techniques to optimize VMs placement for energy, security, fairness, and costs. { We have discussed relevant issues in fog computing relevant issues, while there would be also more relevant big data processing issues in cloud computing and edge computing, such as relevant architectures and applications, distributed data analytical frameworks, cyber defense and cyber intelligence as well as convergence and complexity issues.}

\section{Big Data Analytics}
\label{analysis}
Big data analytics constitutes one of the most important arenas in big data systems, as it allows to uncover hidden patterns, unknown correlations and other useful information, which in turn assist in boosting the revenues for many businesses. In this section, we present an overview of relevant big data analytics techniques and tools. Readers may find some introductions of big data analytics in \cite{J_CAIS_HJWat_2014}.

\subsection{Data Mining}
\label{miningg}
One of the interesting features of CPS is the automated decision making. This means that CPS objects are supposed to be smart in sensing, identifying events and interacting with others~\cite{5621980}. The massive data collected by CPS needs to be converted into useful knowledge to uncover hidden patterns to find solutions, enhance system performance and quality of services. The process of extracting this useful information is referred to as data mining. One solution to facilitate the data mining process is to reduce data complexity {via} allowing objects to capture only the interesting data rather than all of it. Before data mining can be applied to the data, some processing steps need to be completed such as key features selection, preprocessing and transformation of data. Dimensionality reduction is one potential method to reduce the number of features of the data~\cite{6755512}. For instance, in~\cite{5170925}, Chen \textit{et al.} used neural network with k-means clustering via principal component analysis (PCA) to reduce the complexity and the number of dimensions of gene expression data to extract disease-related information from gene expression profiles. Knowledge discovery in databases (KDD) is also used in different CPS scenarios to find hidden patterns and unknown correlations in data so that useful information  can be converted into knowledge~\cite{Fayyad:1996:DMK:257938.257942}.  One such use of KDD is in smart infrastructures systems, where these systems need to answer queries and make recommendations about the system operation to the facility manager~\cite{7515055}.

Tsai \textit{et al.}~\cite{6674155} broke down the core operations of data mining into three main operations: data scanning, rules construction and rules update. Data scanning is selecting the needed data by the operator. Rules construction includes creating candidate rules by using selection, construction and perturbation. Finally, candidate rules are checked by the operator, then evaluated to determine which ones will be kept for the next iteration. The process of scanning, construction and update operations is repeated until the termination criteria is met. This data mining framework works for deterministic mining algorithms such as k-means, and the metaheuristic algorithms such as simulated annealing and genetic algorithm.

\begin{figure}[htbp!]
  \begin{center}   \includegraphics[scale=0.34]{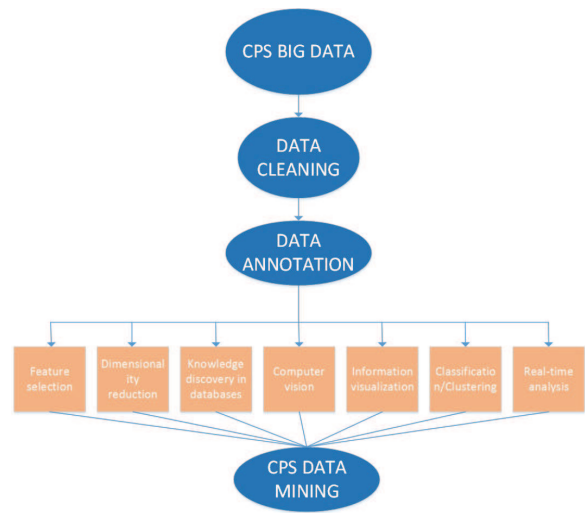}
    \vspace{-0.2cm}
    \caption{\label{tree1} CPS big data mining process.}
  \end{center}
\end{figure}

Clustering, classification and frequent pattern are different mining techniques that can be used to make CPS smarter. Clustering methods have already been discussed in Section~\ref{clustering}. Tsai \textit{et al.}~\cite{6674155} discussed about two different purposes for clustering: i) clustering for infrastructure of IoT, and ii) clustering for services of IoT. Clustering for infrastructure of IoT helps enhance system performance in terms of identification, sensing and actuation, such as in~\cite{60}, where nodes can exchange information between each other to identify whether they can be grouped together depending on the needs of the IoT applications. As for services of IoT, clustering can help provide higher quality services such as in smart homes~\cite{5560652}. On the other hand, classification does not require prior knowledge to complete the partitioning of objects into clusters, also known as unsupervised learning. Classification tools include decision trees, k-nearest neighbor, naive Bayesian classification, adaboost and support vector machines. Classification can also be done to improve infrastructure as well as services of IoT. Finally, frequent pattern mining is about uncovering interesting patterns such as which items will be purchased together with previously purchased items, or suggest items for customers to purchase based on customer's characteristics, behavior, purchase history, and so on. Fig.~\ref{tree1} illustrates the CPS big data mining process for useful information extraction.

The paper \cite{J_TCOMPU_DZhang_n2_2015} proposed NextCell algorithm to improve the location prediction via utilizing the social interplay mined in cellular call records.

\subsection{Real-time Analytics}
Real-time analysis is another approach to produce useful information from massive raw data. Real-time streams data are first converted to a structured form data before being analyzed by big data analysis tools such as Hadoop. Many application domains such as healthcare, transportation systems, environmental monitoring, and smart cities will require real-time decision making and control~\cite{7406686}. For example, Twitter data can be real-time analyzed to enhance the prediction process and to provide useful recommendations to users~\cite{7592729}; terrorist incidents data can be real-time analyzed to predict future incidents~\cite{7568906}; big data stream in healthcare can be analyzed to help medical staff make decisions in real-time, which can help save patients' lives and improve the healthcare services provided, while reducing medical costs~\cite{7529531}. Near real-time big data analysis architecture for vehicular networks was proposed in~\cite{7292404}, which consists of a centralized data storage for data processing and a distributed data storage for streaming processed data in real-time analysis.

{In \cite{C_ICCC_KWang_07_2016}, a real-time hybrid-stream big data analytics model was proposed for big data video analysis. The paper \cite{C_GLOBECOM_BZhou_12_2017} considered the online network analysis as a stream analysis issue and proposed to utilize Spark Streaming to monitor and analyze the high-speed Internet traffic data in real-time. The work \cite{C_ICBD_HCao_12_2017} proposed mobile edge computing nodes deployed on a transit bus with descriptive analytics to explore meaningful patterns of real-time data streams for transit. The paper \cite{C_ICBD_STri_12_2017} discussed the term and related concepts of Real Time Analytics (RTA) for industry big data analytical solutions. This paper \cite{C_ICBDA_ARAli_03_2018} provided a framework to efficiently leverage big data technology and allow deep analysis of large and complex datasets for real-time big data warehousing.}

Arranging the data in a representative form can provide information visualization, which makes the information extraction and understanding of complex large-scale systems much easier~\cite{7360233}. Geographical information systems (GIS) is one important tool of visualization~\cite{7338157}, as it can help real-time analysis of many applications such as in healthcare, urban and regional planning, transportation systems, emergency situations, public safety, and so on. In~\cite{7338157}, the authors proposed a large-scale system data visualization architecture called X-SimViz, which allows users for real-time dynamic data analytics and visualization. Computer vision is another approach to detecting security anomalies. Visualization can also be useful tool in predicting real-time cyber attacks. For instance, in~\cite{6967763}, the authors used computer vision to transform the network traffic data into images using a multivariate correlation analysis approach based on a dissimilarity measure called Earth Mover's Distance to help detect denial-of-service attacks. A computer vision deep learning algorithm for human activity recognition was proposed in~\cite{7520541}. The model is capable of recognizing twelve types of human activities with high accuracy and without the need of prior knowledge, which is useful for security monitoring applications.
\subsection{Cloud-Based Big Data Analytics}
Cloud-based analysis in CPS constitutes a scalable and reliable architecture to perform analytics operations on big data stream, such as extracting, aggregating and analyzing data of different granularities~\cite{7399283}. A massive amount of data are usually stored in spreadsheets or other applications, and a cloud-based analytics service, using statistical analysis and machine learning, helps reduce the big data to a manageable size so information can be extracted, hypothesis can be tested, and conclusions can be drawn from non-numerical data such as photos. Data can be imported from the cloud and users are able to run cloud data analytics algorithms on big datasets, after which data can be stored back to the cloud~\cite{DBLP}. For instance, in~\cite{7582986}, the authors used cloud computing using MapReduce algorithm to conduct analysis on crime rates in the city of Austin using different attributes like crime type and location to help build a design that prevents future crimes for public safety.

Even though cloud computing is an attractive analytics tool for big data applications{,} it comes with several challenges, mainly concerning security, privacy and data ownership, which will be discussed further in Section~\ref{security}. In~\cite{7576619}, the authors extended the use of clouds to mobile cloud computing to help overcome the challenge of resources limitations such as memory, battery life and CPU power. A mobile cloud computing architecture was suggested for healthcare applications with discussion on various big data analytic tools available. In~\cite{7431422}, the authors suggested using a hybrid cloud computing consisting of public and private clouds to accelerate the analysis of massive data workloads on MapReduce framework without requiring significant modifications to the framework. In a private cloud, cloud services delivered over the physical infrastructure are exclusively dedicated to the tenant. The hybrid cloud uses a set of virtual machines {running} on the private cloud, which take advantage of data locality, and another set of virtual machines run on a public cloud to run the analysis at a faster rate.

To optimize the utilization of cloud computing resources, predicting the expected workload and the amount of resources needed becomes important to reduce waste. In~\cite{6877244}, the authors developed a system that predicts the resources requirements of a MapReduce application to optimize bandwidth allocation to the application{, while,} in~\cite{Islam:2012:EPM:2304777.2304880}, the authors used linear networks along with linear regression to predict the future need of new resources and VMs. When the system fails short in predicting the right amount of resources needed, it becomes incapable of accommodating a high workload demand, leading to anomalies. Anomalies detection is an essential part of big data analytics, as it helps improve the quality of service {via} checking {whether} the measurements of the workload observed and the baseline workloads diverge by a specific margin, where the baseline workloads provide a measure on how the demand changes during a period of time based on historical records~\cite{7384281}.

\subsection{Spatial-Temporal Analytics}
Massively data obtained from widely deployed spatio-temporal sensors have caused grand challenges on data storage, process scalability, and retrieval efficiency. The paper \cite{C_IGARSS_YZhong_07_2013} proposed distributed composite spatio-temporal index approach VegaIndexer for efficiently processing the large amount of spatio-temporal sensor data. The paper \cite{C_ACBD_DZheng_11_2014} investigated the big data issues in Internet of Vehicles (IOV) applications and proposed to use clouding based big data space-time analytics to enhance the analysis efficiency. The paper \cite{C_DASC_MSin_11_2017} proposed STAnD to determine anomaly patterns for potential malicious events within these spatial-temporal data sets. Massively data obtained from widely deployed spatio-temporal sensors have caused grand challenges on data storage, process scalability, and retrieval efficiency. The spatial distributed CPS nodes also can be used to analyze location information. The paper \cite{J_IEICE_TCOM_GDing_08_2014} proposed an efficient indoor positioning based on a new empirical propagation model using fingerprinting sensors, called regional propagation model (RPM), which is based on the cluster based propagation model theory, and then the paper \cite{J_IEICETCOM_GDing_03_2015} used particle swarm optimization (PSO) to estimate the location information via Kalman filter to update the initial estimated location.

\subsection{Big Data Analytical Tools}
In this section, some typical tools are briefly introduced for aforementioned three methods of big data analytics, namely data mining, real-time big data analytics and cloud-based big data analytics.

\subsubsection{Tools for Data Mining}
Hadoop~\cite{Acharjya} is an open source managed by the Apache Software Foundation. There are two main components for Hadoop, namely HDFS~\cite{MChenn} and MapReduce~\cite{Derbeko20161}. HDFS is developed from an inspiration of GFS~\cite{6974788}, and it is a scalable and distributed storage system, which is an appropriate solution for data-intensive applications, such as Gigabyte and Terabyte scale. Rather than just being a storage layer of Hadoop, HDFS is also beneficial to throughput improvement of the system and it supplies efficient fault detection and automatic recovery. MapReduce is a framework which is used to analyze massive data sets in a distributed fashion by means of numerous machines~\cite{6779407}. There are two functions in the mathematical model of MapReduce including Map and Reduce, both of which are available to be programmed. $R$ is also an open-source software environment for data mining developed by AT\&T Bell Labs~\cite{Fan:2013:MBD:2481244.2481246}. Actually, $R$ is a realization of the S language used to explore data, implement statistical analysis and draw plots. Compared {with $S$, $R$ is more popular and supported by a large number of database manufacturers, such as Teradata and Oracle.}

\subsubsection{Tools for Real-Time Big Data Analytics}
Storm~\cite{Fan:2013:MBD:2481244.2481246,6842585} is a distributed real-time computing system for big data analysis. Compared with Hadoop, Storm is easier to operate and more scalable to provide competitive and efficient services. Storm makes use of distinct topologies for different storm tasks in terms of storm clusters, which are composed of master nodes and worker nodes. The master nodes and worker nodes play two kinds of roles in the fields of big data analysis, namely nimbus and supervisor, respectively. The functions of these two roles are in agreement with jobtracker and tasktracker of the MapReduce framework. Nimbus takes charge of code distribution across the storm cluster, the schedule and assignment of worker nodes tasks, and the whole system surveillance. The supervisor {compiles} tasks given by nimbus. Splunk~\cite{zadrozny2013big} is also a real-time platform designed for big data analytics. Based on the web interface, Splunk is available to search, monitor and analyze machine-generated big data, and the results are exhibited in different varieties including graphs, reports, alerts and so on. Compared with other real-time analytical tools, Splunk provides various smart services for commercial operations, system problem diagnosis, and so on.
\begin{figure}[htbp!]
  \begin{center}   \includegraphics[width=45cm,height=10cm,keepaspectratio]{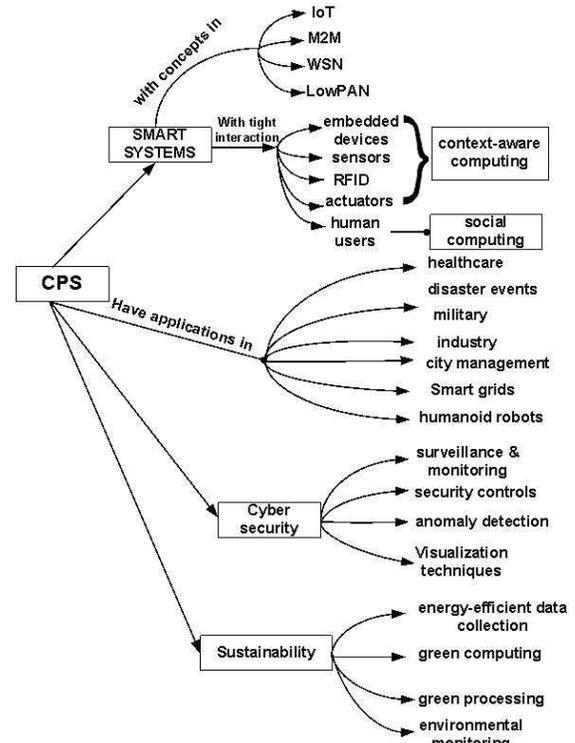}
    \vspace{-0.2cm}
    \caption{\label{tree2} An illustration of CPS taxonomy.}
  \end{center}
\end{figure}
\subsubsection{Tools for Cloud-Based Big Data Analytics}
As the most popular tool for cloud-based big data analytics, Google's cloud computing platform~\cite{Hashem:2015:RBD:2946158.2946407} consists of GFS~\cite{BRao} (big data storage), BigTable~\cite{2016:SPA:2974459.2974485} (big data management) and MapReduce (cloud computing), which was discussed in the previous section. GFS is a distributed file system and it is enchanced to meet the requirements of big data storage and usage demands of Google Inc. In order to deal with the commodity component failure problem, GFS facilitates continuous surveillance, errors detection and component faults tolerance. GFS adopts clustered approach that divides data chunks into 64-KB blocks and stores a 32-bit checksum for each block. BigTable supplies highly adaptable, reliable, applicable and dynamic control and management in the field of big data placement, representation, indexing and clustering for enormous and distributed commodity servers, and it constitutes of a row, column, record tablet and time stamp.

\subsection{Summary and Insights}
To better extract information from big data, it is of utmost importance to enhance cloud's analysis performance. A combination of different techniques discussed in this section can be used to optimize cloud computing resources. If VMs and cloud's resources and requirements can be predicted beforehand, then workloads can be efficiently processed and analyzed by taking advantage of cloud {analytical} tools previously discussed. Furthermore, using a hybrid cloud can further speed up the analysis of workloads, leading to reduced latency and efficient data mining.

\section{Big Data Cybersecurity and Privacy}
\label{security}
In the realm of cyber physical systems, the tight interaction among physical objects which collect and transmit {a} large volume of data place security threats under the spotlight of attention. With this enormous amount of data that are constantly flowing through the network, it becomes essential to protect the system from cyber attacks\cite{IETCPS_RAtat_04_2017}. In this section, we provide an overview of the different security solutions proposed for big data storage, access and analytics (see Table~\ref{tablesec} for a summary).
\subsection{Security in Big Data Storage and Access}
While data storage in the cloud offers several advantages in terms of data storage, availability, scalability and processing, it increases the chance of malicious attacks, that in addition to potential privacy invasion by cloud operators who can have accesses to sensitive data. All this puts a question mark on whether cloud data storage is feasible, especially for governmental agencies and financial industries. Several works have attempted to solve the security challenges of cloud storage. For instance, Gai \textit{et al.} proposed a method that splits files into encrypted parts and store them in distributed cloud servers without users' data being directly reached by cloud service operators~\cite{7502279}. In~\cite{7502311}, the authors optimized the data placement on cloud servers that minimizes the retrieval time of data files while guaranteeing their security based on the distance between nodes that store the data chunks, such that the malicious attacker cannot guess the locations of all the data chunks. In~\cite{7724531}, the authors suggested that data should be encrypted and decrypted before being sent to clouds.  The paper \cite{J_IoTJ_CJDo_04_2017} addressed iOS devices as case studies and stressed the possible applications for pairing mode in iOS devices, which supports a trusted relationship between an {iOS} device and a personal computer, for covert data exfiltration.

When data need to be transferred from one cloud to another, data privacy and integrity become important. In~\cite{7636866},
Ni \textit{et al.} proposed a secure data transfer scheme where users encrypt the data blocks before uploading to the cloud.
When transferring from one cloud to another, a security protocol was described using secret keys and a signature checking
with polynomial-based authentication was performed without retrieving data from the source cloud. While the mentioned works considered encryption
 as a way to protect the data from privacy violations, encryption introduces a new challenge: cloud data deduplication,
 especially when data is shared among many users. Even though deduplication can save up to 95\% in terms of needed storage for
 backup applications~\cite{opendedup}, and 68\% for standard file systems~\cite{Meyer:2012:SPD:2078861.2078864}, it wastes resources,
 and consumes energy, and makes the data management very complicated. In~\cite{7511769}, Yan \textit{et al.} attempted
 to solve the deduplication problems via proposing a scheme, where the users upload the encrypted data to the cloud along
 with a token for data duplication check, which is then used by the cloud service providers to check whether the data has already been stored.
 A scheme to verify data ownership was presented to ensure secure data management.
\subsection{Security in Big Data Analytics}
\label{sec_analy}
Enabling security and privacy aspects of big data analytics has attracted a great attention from the scientific community mainly due to different reasons. First, the data {are more likely} stored, processed and analyzed in several cloud centers leading to security issues due to the {the random locations} of data. Second, big data analytics treats sensitive data in similar way to other data without taking security measures such as encryption or blind processing into consideration~\cite{7543859}. Third, big data computations need to be protected from malicious attacks in order to preserve the integrity of the extracted results. In the realm of CPS, an enormous amount of data make the {surveillance} of security-related information for anomaly detection a challenging task for analysts. In healthcare, for instance, the security issues of information extraction from massive amount of data and accurate analytics are of high importance. Sensitive data recorded in databases need to be protected {via} monitoring which applications and users get accesses to the data~\cite{7306738}. In order to guarantee a strong secure big data analytics, the following tasks can be performed~\cite{6725337}:
\begin{itemize}
\item Surveillance and monitoring of real-time data streams,
\item Implementation of advanced security controls such as additional authentication and blocking suspicious transactions,
\item Anomaly detection in behavior, usage, access and network traffic,
\item Defending the system against malicious attacks in real-time,
\item Adoption of visualization techniques that give a full overview of network problems and progress in real-time.
\end{itemize}
 In~\cite{7429688}, the authors tackled the big data analytics in mobile cellular networks based on random matrix theory, where big data is represented in matrix form of size $n\times N$, where $n$ is the number of data samples of a random vector $x$, and $N$ is the number of independent realizations of $x$. For cellular networks, big data manifests as big signaling data consisting of
\begin{enumerate}
  \item a large number of control messages to ensure reliability, security and efficiency of communications,
  \item big traffic data which require traffic monitoring and analysis to balance network load and optimize system performance
  \item big location data generated by GPS sensors, Bluetooth, WiFi and so on, to assist in different areas such as transportation systems, public safety, crime hot spots analysis and so on,
  \item big radio waveforms data emanating from 5G massive MIMO systems to estimate users' moving speed for purposes of finding correlation among transmitted signals as well as assist in channel estimation,
  \item big heterogeneous data such as data rate, packet drop, mobility and so on that can be analyzed to ensure cybersecurity.
\end{enumerate}
The work \cite{J_TII_JFeng_EA_2018} proposed a secure high-order clustering algorithm via fast search density peaks on hybrid cloud for industrial Internet of Things.

Machine learning, among other tools, offers a promising solution to automate many of the above mentioned security-related tasks, especially with the continuous growth of the flowing data in terms of scale and complexity. Through the process of training datasets, machine learning makes possible the detection of future security anomalies via detecting unusual activities in the network traffic. To achieve a higher accuracy, a large volume of training datasets are needed, but this would be at the cost of added overhead and storage constraints. The process of training can be supervised, unsupervised or semi-supervised, depending on whether the outcome of a particular dataset is already known. {Particularly,} the system starts via classifying similar datasets into clusters to determine their anomaly. A human analyst can then explore and identify any unusual data. The outcome found by the analyst can then be fed back to the training system in order to make it more ``supervised''\cite{he}. This has the potential in enabling the training system adapt to new forms of threats without human intervention, so actions can be immediately taken before actual damages occur.

Different approaches for anomaly detection exist in literature such as discretizing the continuous domain into different dimensions such as in the surveillance system in~\cite{ristic}, where the author partitioned the surveillance area into a square grid where the positions and velocities of the moving objects falling in each cell are modeled by a Poisson point process. Another approach is the multivariate Gaussian analysis in which data are flagged as abnormal when they lie a number of standard deviations away from the mean. For instance, in~\cite{rocha}, the authors used multivariate Gaussian analysis to detect Internet attacks and intrusions via analyzing the statistical properties of the IP traffic captured. In clustering methods such as k-means clustering, data points can be grouped into clusters based on their distance to the center of the cluster. Then, if some data point lies outside of the group cluster, it is considered as an anomaly. The authors in~\cite{jia} used kernel k-means clustering with local-neighborhood information to detect a change in an image by optimally computing the kernel weights of the image features such as intensity and texture features. As for the artificial neural network approach, one implementation of such a model is the autoencoder, also known as replicator neural network, which flags anomalies based on calculations of the difference between the test data and the reconstructed one. This means that if the error between test and reconstructed data exceeds a specified threshold, then it is considered far away from a healthy system distribution~\cite{murphree}. An example of such an approach is given in~\cite{yuan}, where the authors used the autoencoder as a high accuracy and low-latency model to detect anomalies in the energy consumption and operation of smart meters.

{ Privacy preserving data analytics and mining can be quite challenging tasks since analyzing encrypted data is an inefficient, costly and non-straightforward solution. Homomorphic encryption is one of the solutions proposed to enable analytical operations to be performed on ciphertexts using multiple mathematical operations~\cite{6650119,7478543}. For instance, in~\cite{8005854}, the authors used Efficient Privacy-preserving Outsourced calculation with Multiple keys (EPOM) homomorphic encryption to encrypt data before sending to cloud. The cloud then uses ID3 algorithm to perform data mining on encrypted data. The algorithm uses a hierarchical tree decision to determine which attributes from a set of samples provide the best prediction or information gain. Another approach for processing mining on encrypted data in cloud using homomorphic encryption was suggested in~\cite{8010376}. A cloud service provider (CSP) collects and stores encrypted data, while a server, referred to as Evaluator, collaborates with CSP to perform mining over encrypted data. A miner submits encrypted mining queries to CSP which in turn computes inner product between vectors to determine the frequency of the mining itemset without CSP and Evaluator having access to the sensitive data. However, homomorphic encryption can be computationally expensive and impractical for big datasets~\cite{6863131}.

One potential approach to protect private data during the analytic processes is the k-anonymization proposed in~\cite{6565224}. First, users who access the data need to be authenticated and authorized based on the level of shared results' privacy. Then, a list of hashed and primary identifiers is generated to act as a data filter for the information that can be accessible by the authorized user. K-anonymization is then applied on the personally identifiable columns in the dataset in order to generalize or suppress values in the output dataset. The result is an k-anonymized list on which analysis and mining can be performed on the authorized accessible data.

Although k-anonymization appears to be a promising adaptable approach to privacy preserving data analytics independent of the underlying processes, its feasibility in the big data contexts is possibly not further evaluated for data types and computational time. Furthermore, k-anonymization can be problematic if different datasets contain same sensitive values~\cite{1623889}. One approach that attempts to solve this problem is the cosine similarity computation protocol suggested in~\cite{6863131,7431421}. The proposed approach allows for larger datasets scalability for both binary and numerical data types in a time-efficient manner. The idea is to allow data to be shared without disclosing the sensitive information to unauthorized users. This can be done via computing the scalar product between different vectors of numerical values, such as calculating the cosine of the angle between them. Having the result closer to 1 indicates that vectors are more similar to each other.

On the other hand, many of the big data applications have hierarchical structures in nature, and thus require hierarchical privacy preserving solutions. For instance, in~\cite{8013260}, hierarchical cloud and community access control can be implemented to strengthen privacy preserving in smart homes and smart meters. The home controller, which protects household personal data, is connected to a cloud platform through a community network, which provides privacy preserving solutions to homes through data separation, aggregation and fusion. The cloud combines the access control schemes for homes and community in more complex and stronger privacy protection process.}

\subsection{Summary and Insights}
From Table~\ref{tablesec}, it is clear that a single security solution is not sufficient to ensure a robust system against attackers. It is quite necessary to incorporate different strategies to face security flaws that stem from poor systems designs. For instance, it makes sense to combine advanced security controls with cryptography to guarantee that only legitimate users have accesses to data. However, such a solution might not work well for delay-sensitive CPS applications such as e-health systems, especially in the cases that the small sensors have limited computational capabilities. For such applications, it is more efficient to employ visualization detection along with machine learning-based anomaly detection techniques to protect users' sensitive data.

\begin{table*}[htbp!]
\centering
\caption{Summary of security solutions proposed for CPS}
\label{tablesec}
\hspace{-1cm}
\begin{tabular}{l|
p{6em}|
p{6em}|
l |
l |}
\cline{2-5}
                                                                                                            &Literature                                                                                                                                                                                                                                              &Security Solutions                                                        &Advantages                                                                                                                                                                                                                                                      &Disadvantages                                                                                                                                                                                                                                                                                                                                                                                      \\ \cline{2-5}
\hline\begin{tabular}[c]{@{}l@{}}Key\\ Encryption \\ and Digital \\Signatures\end{tabular}    &\cite{BRao}\cite{7502279}\cite{7502311}&Secure data placement on cloud& \begin{tabular}[c]{@{}l@{}}i) ensures users' data protection on\\  clouds; ii) does not require complex \\ signatures and keys; iii) provides data\\ privacy at users' side\end{tabular}                                                                        & \begin{tabular}[c]{@{}l@{}}i) data might need to be encrypted/decrypted\\ before transferring to clouds;\\ ii) can have a long data retrieval time;\\ iii) not suitable for making real-time network\\ decisions; iv) not very robust as attackers can\\ guess data's locations; v) service providers can \\ have access to users' data; vi) complicated data\\ management and access\end{tabular} \\ \cline{2-5}
                                                                                                            &\cite{7502279}\cite{7724531}\cite{7636866}\cite{7511769}\cite{7414062}\cite{7478543}                                                                                                    & Cryptographic Solutions                                                   & \begin{tabular}[c]{@{}l@{}}i) ensures information protection, data\\ privacy and integrity; ii) works in \\ conjunction with other security\\ solutions for more robust security\end{tabular}                                                                   & \begin{tabular}[c]{@{}l@{}}i) costly in terms of computational overhead;\\ ii) can have a large power consumption;\\ iii) increases latency;\\ iv) inefficient for many CPS devices with limited\\ computational capabilities; v) works only for\\ fixed malicious behaviors\end{tabular}                                                                                                          \\ \cline{2-5}
\hline\begin{tabular}[c]{@{}l@{}}Data Access\\ Permission\\Restrictions\end{tabular}  &\cite{7097702}\cite{7306738}\cite{7737997}\cite{7347956} & Advanced security controls& \begin{tabular}[c]{@{}l@{}}i) selective access control;\\ ii) can be used in conjunction with\\ cryptography for a more robust security;\\ iii) good for real-time data streams;\\ iv) provides additional layer of protection\end{tabular}                     & \begin{tabular}[c]{@{}l@{}}i) difficult to access for legitimate users;\\ ii) needs to be combined with other solutions;\\ iii) not sufficient by itself to protect users' data;\\ iv) might increase latency from accessing the data\end{tabular}                                                                                                                                                 \\ \cline{2-5}
\hline\begin{tabular}[c]{@{}l@{}}Visualization \\ Detection of\\ Anomalies\end{tabular} &\cite{6967763}\cite{he}\cite{rocha}\cite{jia} \cite{murphree}\cite{yuan}\cite{7737997}\cite{7479069}\cite{6882174}\cite{7232339}\cite{7762123}\cite{7510811}& Anomaly Detection                                                         & \begin{tabular}[c]{@{}l@{}}i) does not require costly solutions\\ such as signatures and keys;\\ ii) works for a variety of attacks with\\ varying malicious behaviors;\\ iii) can detect abnormal traffic patterns\\ with no associated signature\end{tabular} & \begin{tabular}[c]{@{}l@{}}i) requires a large dataset for training;\\ ii) costly in terms of overhead and storage of\\ training data; iii) might require human analysts'\\ intervention to make the detection more\\textbackslash supervise\end{tabular}                                                                                                                                          \\ \cline{2-5}
                                                                                                            &\cite{7520541}\cite{7379452}\cite{6967763}\cite{ristic}\cite{jia}\cite{7247576}& Visualization Techniques                                                  & \begin{tabular}[c]{@{}l@{}}i) ideal for monitoring real-time traffic;\\ ii) provides global overview of network\\ problems\end{tabular}                                                                                                                         & \begin{tabular}[c]{@{}l@{}}i) not accurate if traffic patterns are misclassified;\\ ii) requires complex statistical analysis of traffic;\\ iii) might need to be supervised by humans;\\ iv) might require prior knowledge for a better \\ intrusion detection\end{tabular}                                                                                                                       \\ \cline{2-5}
\end{tabular}
\end{table*}

\section{Big Data Meet Green Challenges for CPS}
\label{green}

There are two view directions for big Data meeting green challenges for CPS. Firstly, we address greening big data systems for CPS, including CPS big data collection/storage, computing and processing. Secondly, we discuss big data with CPS toward green applications. Table~\ref{sustainable} shows different sustainable applications, along with challenges and solutions for a greener CPS. Table~\ref{tablegreen} provides a summary of these solutions proposed in different research papers, while Table~\ref{approaches} groups them into three different categories: green data management, green architectures and green software solutions.


\begin{table*}[htbp]
{
\centering
\caption{Sustainable CPS: applications, challenges and solutions.}
\label{sustainable}
\begin{tabular}{l|ll}
\cline{2-2}
                                                                                                                                                                                                         & \multicolumn{1}{c|}{{ \textbf{Sustainable CPS}}}                                                                                                                          &                                                                                                                                                                                                                                                                                                                                                                                                                                                                                                                                                                                                                                                                                                                                                                                                                                                                                                                                                                                                                                         \\ \hline

\multicolumn{1}{|c|}{\textbf{Applications}}                                                                                                                                      & \multicolumn{1}{c|}{\textbf{Challenges}}                                                                                                                                                      & \multicolumn{1}{c|}{\textbf{Solutions}}                                                                                                                                                                                                                                                                                                                                                                                                                                                                                                                                                                                                                                                                                                                                                                                                                                                                                                                                                                         \\ \hline

\multicolumn{1}{|l|}{\begin{tabular}[c]{@{}l@{}}Environmental Monitoring\\ \\ - Air/Water pollution\\ - Oil spills pollution\\ - Noise pollution\end{tabular}}                   & \begin{tabular}[c]{@{}l@{}}- Limited communications \\ between subnetworks\\ \\ - Storage limitations\end{tabular}                                                                                                    & \multicolumn{1}{|l|}{\begin{tabular}[c]{@{}l@{}}- Provide relay incentives to extend  communication~ \cite{7065282,7110391,huo2017coalition},\\\cite{Bonomi,6800057}\\ \\ -Minimize the number of relay transmissions~\cite{6800057}\\ \\ - Data reduction techniques(aggregation, compression, network\\ coding)~\cite{7565185}, \cite{7150324,7524427}, \cite{7336366}, \cite{7284356,7492645,6733207},\cite{6877244}, \cite{7389318},\cite{179401},\cite{Perino},\\\cite{7152629}, \cite{7593221}\end{tabular}}                                                                                                                                                                                                                                                                                                                                                                                                                                                            \\ \cline{1-1}

\multicolumn{1}{|l|}{\begin{tabular}[c]{@{}l@{}}Power Management\\ \\ - Smart Grids\\ - Green buildings\\ - Green infrastructure\\ - Renewable energy sources\end{tabular}}      & \begin{tabular}[c]{@{}l@{}}- Redundant transmission \\ links\\ \\ - Deduplication\\ \\ - Redundant information in \\ data collection\end{tabular}                                                                     & \multicolumn{1}{|l|}{\begin{tabular}[c]{@{}l@{}}- Remove redundant transmission links using energy-efficient\\ topology control algorithm and clustering~\cite{5319232},\cite{6863675},\cite{Bali2016476},\cite{huo2017coalition},\cite{SPEZZANO20151016},\\\cite{1427709}, \cite{5351723}, \cite{7217796}, \\ - Employ data duplication checks~\cite{7511769}\\ \\ - Employ compressive sensing- based data collection~\cite{7588229},\cite{7478010}\end{tabular}}                                                                                                                                                                                                                                                                                                                                                                                                                                                                                           \\ \cline{1-1}

\multicolumn{1}{|l|}{\begin{tabular}[c]{@{}l@{}}Green Transportation\\ \\ - Fuel management\\ - Electric vehicles\\ - Sustainable transport\\ - Transport planning\end{tabular}} & \begin{tabular}[c]{@{}l@{}}- Increased energy expenditure\\ in cloud data centers\\ \\ Increased power consumption \\ of networked devices\\ \\ - Large volume of exchanged data\\ between cloud centers\end{tabular} & \multicolumn{1}{|l|}{\begin{tabular}[c]{@{}l@{}}- Efficient utilization of resources~\cite{7576619},\cite{6877244},\cite{Islam:2012:EPM:2304777.2304880}\\ \\ - Integrate energy harvesting from renewable sources~\cite{6475927},\cite{7828559}\\ \\ - Sleep scheduling for unused data centers~\cite{7399696},\cite{7828559}\\ \\ - Employ energy-efficient hardware(DVFS)~\cite{beloglazov2011a}, \cite{7590092}, \cite{Chao2015269}, \cite{6162477}\end{tabular}}                                                                                                                                                                                                                                                                                                                                                                                                                                                                                                  \\ \cline{1-1}

\multicolumn{1}{|l|}{{ \begin{tabular}[c]{@{}l@{}}Green Energy\\ \\ - Energy harvesting\\ - Energy conservations\end{tabular}}}                              & \begin{tabular}[c]{@{}l@{}}- Cloud data centers distant from\\ users\\ \\ -VM placement in cloud centers\\ \\ - Increased energy consumption\\ in case of hardware/software\\ failures\end{tabular}                   & \multicolumn{1}{|l|}{\begin{tabular}[c]{@{}l@{}}- Design energy-efficient cloud data centers~\cite{7399696}\\ - Activity scheduling (switch devices to sleep mode)~\cite{7828559}\\ - Power control and adjustment of networked devices~\cite{Chao2015269,6162477}\\ - Design architectures for power saving\cite{7399696},\cite{7590092}\\ - Energy-efficient routing and traffic engineering techniques\cite{7110391},\cite{7562084},\\\cite{Mahadevan}, \cite{6195471},\cite{Heller:2010:ESE:1855711.1855728},\cite{6648647},\cite{6996601},\cite{6888901}, \cite{7547281},\cite{7348690}\\ - Employ advanced communication technologies (MIMO)\\ - Employ cloudlets\cite{7576619}, \cite{7295892}\\ - Energy-efficient VM techniques (VM migration, VM placement,\\ VMallocation)~\cite{7353181},\cite{6689479}\\ - Reduce unnecessary executions using backups~\cite{Renault},\cite{Mereuta},\cite{6984236},\cite{6612229}\end{tabular}} \\ \hline
\end{tabular}
}
\end{table*}

\subsection{Greening big data for CPS}
\subsubsection{Green Data Collections and Communications}
With {a} massive number of interacting objects, gathering sensed data poses a challenge in terms of energy consumption, mainly due to the limited communication range between subnetworks, necessitating that objects act as relays for surrounding objects in order to extend their communications. This affects the lifetime of objects since each object needs to relay large volume of data generated by its neighbors~\cite{6800057}. In this context, different solutions have been proposed for different big data applications. In~\cite{6800057}, Takaishi \textit{et al.} proposed energy efficient solutions for data collection in densely distributed sensor networks. In an attempt to reduce the number of relay transmissions needed, sensor nodes transmit their data to a data collector node, the sink node, when they become close in proximity to it. Therefore, it becomes important to figure out the trajectory that the sink node needs to follow the nodes' information such as location and residual energy, as well as the cluster formation in order to reduce the energy consumption of data collection. Data compression technology is another solution to help deal with challenges of data storage, collection, transmission, processing, and analysis. For instance, in~\cite{7438894}, the authors proposed a highly efficient lossy data compression based on smart meters' load features, states and events with a small reconstruction error. ZIP-IO compression technique was proposed in~\cite{6412159} using FPGA as a potential implementation framework. Video compression is another important tool in big data for surveillance applications. The authors in~\cite{7723680} proposed a  background-based coding optimization algorithm that uses the residual gradient and the block edge differences to improve picture quality while achieving a high level of compression.

To achieve higher power savings for machine-to-machine (M2M) or machine type communications (MTC), this paper \cite{chao2011power} proposed improved approaches for M2M devices, radio access networks and core networks with simplified activities under optimized signaling flow without introducing negative impacts on legacy human-to-human (H2H) terminals. In wireless cellular networks, when mobile terminals can directly communicate each other without the aids of central stations or base stations, the communications in those scenarios are called device-to-device (D2D) communications~\cite{SSS_Hao}, which is another term of CPS in those scenarios. The paper \cite{yang2016energy} studied energy-efficient power control for D2D communications underlaying cellular networks with multiple D2D pairs and co-channel interference caused by resource sharing. The work \cite{xu2018towards} proposed an architecture using D2D multicast for energy efficient content delivery in cellular networks. Removing redundant transmission links can also reduce energy consumption while increasing network capacity. Topology control algorithms such as  local  minimum  spanning  tree (LMST)~\cite{1427709} and Local Tree-based Reliable Topology (LTRT)~\cite{5351723} have low computational complexities and can help obtain the best logical topology which can be beneficial for energy efficient data collections. In addition to removing redundant transmission links, removing redundant data can help save energy, such as in~\cite{7588229}, where a compressive-sensing-based collection framework was proposed for reducing data redundancy and saving energy. To implement this solution, an online learning module predicts the amount of data (principal data) that needs to be collected for compressive sensing. This means { that the principal data are} supposed to represent the whole big data using the compressive sensing technique. Then, each node can locally tweak the collection strategy dynamically depending on neighbors status, residual energy, and link quality. Nodes' clustering can also contribute to energy saving {via} reducing the number of data collection and transmissions, such as fan-shaped clustering proposed in~\cite{7217796} for large-scale networks with energy efficient selection of a cluster head and relay node. Other clustering algorithms have existed in literature, such as the well-known Low-Energy Adaptive Clustering Hierarchy (LEACH)~\cite{6388472}, which can be useful for big data wireless sensor networks. The work \cite{ComMag_KWang_12_2016} adopted an energy-efficient architecture for Industrial Internet of Things (IIoT) with a sense entities domain, RESTful service hosted networks, a cloud server, and user applications to balance the traffic load and support a longer lifetime of the whole system. The paper \cite{atat2018green} mainly studied two important issues: 1) performing cyber-physical systems (CPS) communications over cellular networks with ubiquitous coverage, global connectivity, reliability and security, 2) offloading a proportion of CPS traffic to small cells and freeing more network resources to other users.

\subsubsection{Greening big data computing}
While cloud centers are becoming an important aspect of big data computing to process data chunks in parallel, they contribute to a high energy expenditure, leading to increased costs and maintenance~\cite{Greenberg:2008:CCR:1496091.1496103}. Therefore, research efforts are shifting towards creating sustainable big data computing techniques. Shojafar \textit{et al.} proposed a job scheduler between servers called MMGreen to reduce energy consumption of computing in cloud data centers~\cite{7590092}. The MMGreen architecture is composed of physical servers hosting VMs connected to a front-end component that manages the incoming workload. Scheduler jobs that use dynamic voltage and frequency scaling (DVFS) technique for energy efficient servers include static scheduler whose power consumption is independent of clock rates and usage, and sequential schedulers which attempt to minimize reconfiguration costs via performing offline resource provisioning via predicting future workload information~\cite{beloglazov2011a}.   The paper \cite{J_TBD_HLiu_n2_2018} proposed 1) thermal-aware and power-aware hybrid energy consumption model jointly including the information of computing, cooling, and migration energy consumption, 2) a tensor-based task allocation and frequency assignment model to reflect the relationship among different tasks, nodes, time slots, and frequencies, 3) a  thermal-aware and DVFS-enabled big data task scheduling algorithm for the energy consumption reduction of data centers.

 In~\cite{7590092}, the authors described different techniques to reduce energy consumption such as DVFS to reduce VMs' frequencies and real-time adjustment of VMs frequencies processing and switching while maintaining quality of service to users. In~\cite{Chao2015269,6162477}, M2M power savings were optimized {via} reducing the execution frequency of some activities without negatively impacting the human-to-human communications.

Besides targeting the energy efficiency of cloud servers, network devices also need energy efficient solutions, since they also contribute to the total energy expenditure of the cloud data center. Traffic engineering is one solution for this problem, which takes advantage of the traffic prediction to turn off network devices such as switches during idle periods in order to reduce power consumption~\cite{Mahadevan,6195471,Heller:2010:ESE:1855711.1855728,6648647}. Traffic engineering techniques, such as the software defined networking (SDN)-based traffic engineering~\cite{6996601}, allow the network devices to dynamically adapt to current workload. One problem with traffic engineering techniques is that the predicted traffic pattern might not be accurate due to the variability of big data applications running in the data center. This makes the network configuration suffer from frequent oscillations, since the network configuration needs to be updated frequently leading to performance degradation~\cite{6888901}. To take into consideration this time-varying aggregate traffic load, one proposed solution is to include flow deadlines to measure the speed at which requests' responses are delivered to users. This allows the design of energy-efficient scheduling and routing for data center networks~\cite{6888901,7547281,7348690}. Another solution to enhance the inaccurate traffic engineering techniques is to take into consideration the unique features of data centers such as regularity of the topology, VM assignment, application characteristics. Such a framework was proposed in~\cite{6689479}, where the energy-saving problem was solved in two steps: i) a VM assignment algorithm that integrates application characteristics and network topology to better understand traffic patterns for energy efficient routing, and ii) an algorithm that minimizes the number of switches and balances traffic flow among them. Experimental results showed a 50\% energy savings using the proposed framework.

\subsubsection{Green Processing}
An energy-efficient orchestrator for smart grid applications was proposed in~\cite{7399696}. The green orchestrator coordinates sustainability between smart grids and big data enterprises from green infrastructure (data centers) to running green frameworks such as Hadoop MapReduce. The orchestrator's main components are: i) a green lesser to establish a per-job service-level agreements (SLA) that takes into account the available power, the power consumption statistics of jobs, the network and server states; ii) a pre-execution analyzer that executes jobs based on their power consumption statistics; iii) a network and server states predictors; iv) a network traffic analyzer which helps eliminate redundant traffic using traffic engineering techniques; v) a VMizer that intelligently places VMs such that some nodes are put to sleep; (vi) a pizer that schedules and places processes to a subset of clusters such that system resources are efficiently utilized; and (vii) a post-execution analyzer to analyze the energy profile of completed jobs.

Another green big data processing architecture is checkpointing aided parallel execution (CAPE), which uses checkpoints that save the sates of processes to avoid restarting unnecessary executions from beginning in case of hardware or software failures~\cite{Renault,Mereuta,6984236}. CAPE also allows threads of a shared-memory program to be executed in parallel on a distributed memory architecture rather than a shared memory architecture. The CAPE architecture can lead to energy saving since if the execution period of processors is short, they can go to idle mode rather than staying active for the whole execution of the program. This makes it beneficial in processing big data in CPS applications~\cite{6984236,6612229}.

Another approach to green big data processing is the efficient utilization of network resources by reducing the volume of communications that need to be exchanged in cloud data center networks. For instance, Asad \textit{et al.} proposed spate coding for the purpose of reducing the amount of exchanged data, but without compromising the rate of information exchange~\cite{7389318}. Spate coding incorporates both index coding and network coding, and uses side information originating from several processes sharing a physical node to encode packets. This coding technique was shown to reduce the volume of communications by 62\%, along with other advantages such as improving the utilization of system resources from disk utilization, queue size, and the number of bits transmitted during shuffle phase of Hadoop (200\%)~\cite{7389318}. Other approaches to reduce the burden on data centers from the information exchange include traffic flow prediction to reduce network transfers~\cite{6877244,179401}, and redundancy elimination scheme to remove redundant information data exchange, among others~\cite{Perino,7152629,7593221}.

\subsection{CPS-based Big Data toward Green Applications}
CPS-based big data applications can contribute to greening different sectors, environment, economic, and social/technical issues~\cite{7473821,7473815,7430133}. As for the environment, efforts are made to reduce air/water pollution as well as the impact of climate changes. For instance, sensors can be deployed to monitor air and water qualities. Using MapReduce or Spark programming frameworks, the concentration of pollutants can be monitored and studied~\cite{7474371}; the air quality not covered via monitoring stations can be estimated~\cite{7179453}, and the causalities of air pollutants can be identified using urban big data dynamics~\cite{7562036}. Pollution can also originate from oil spills. Predicting such catastrophes very early can help save beaches, coastlines and waters~\cite{7473821,7473815}. Marine oil spills can be detected using a large archive of remote sensing big data~\cite{7565634}; and a real-time warning can be generated from a quantitative data analysis using supervised oil system~\cite{Fingas20149}. Concerning water pollution, underwater sensors can be deployed to monitor water environment, such as water level, water flow, temperature, and pressure. The sensor network can be connected to a cloud platform via a wireless transceiver for analysis and visualization~\cite{7168518}. Noise pollution in cities is another contributing factor to environmental pollution, which can have negative impact on health, especially with the increasing number of circulating vehicles~\cite{7473821,7473815}. Noise pollution levels can be predicted via collecting four data sources: complaint data, social media, points of interests, and road network data~\cite{Zheng:2014}. These data can be gathered via deploying static municipal sensors, together with participatory sensing with smart phones in order to provide more accurate noise maps~\cite{7184864}.

As for green economics, CPS applications can be targeted for optimizing the energy use. One example is FirstFuel, which monitors temperature and lightnings inside buildings by checking the running status of the equipments, such as fans, heating and cooling units~\cite{7473821}. Power management can be realized using sensors monitoring the whole building, as well as using smart meters for electricity consumption measurements. {A solution in energy efficient smart meters} has been proposed in~\cite{7037178} using coalition game that maximizes the pay-off values of smart meters. An approach to predict daily electricity consumption inside buildings using data analysis was proposed in~\cite{5175339}. The authors used canonical variate analysis to group electricity consumption profiles into clusters in order to identify abnormal energy usage. Energy efficiency can also be employed in transportation systems to reduce fuel wastage. One example is~\cite{7113543}, where the authors used electric vehicles (EV) battery model to estimate the driver's behaviors and driving range to improve energy efficiency.

Social media and participatory sensing can also be used toward greener environments. For instance, in~\cite{Qin:2014}, the authors used social media to optimize smart grid management. In~\cite{7084108}, participatory sensing was used to implement a navigation service called GreenGPS to allow drivers to obtain customized routes that are the most fuel-efficient.

\begin{table*}[htbp!]
\centering
\caption{Summary of different green solutions proposed for CPS}
\label{tablegreen}
\begin{tabular}{
l |
l |
l |}
\cline{2-3}
                                                                      & Literature                                                                              & Green Solutions                                                                                                                                                                                                              \\ \hline
\multicolumn{1}{|l|}{Data Collection/Storage} & \begin{tabular}[c]{@{}l@{}}{\cite{6800057}}\\\hline {\cite{1427709}, \cite{5351723}, \cite{7217796}}\\\hline {\cite{7438894}, \cite{6412159}, \cite{7723680}, \cite{7588229}}\end{tabular}                     & \begin{tabular}[c]{@{}l@{}}Minimizing number of relay transmissions\\\hline Removing redundant transmission links\\\hline Data compression techniques\end{tabular}                                                                       \\ \hline
\multicolumn{1}{|l|}{Computing}               & \begin{tabular}[c]{@{}l@{}}{ \cite{beloglazov2011a}, \cite{7590092}, \cite{Chao2015269}, \cite{6162477}}\\\hline {\cite{7110391}, \cite{7562084}, \cite{Mahadevan}, \cite{6195471}, \cite{Heller:2010:ESE:1855711.1855728}, \cite{6648647}, \cite{6996601}, \cite{6888901}, \cite{7547281}, \cite{7348690}}\end{tabular}                               & \begin{tabular}[c]{@{}l@{}}Dynamic voltage and frequency scaling\\\hline Traffic engineering\end{tabular}                                                                                                                          \\ \hline
\multicolumn{1}{|l|}{Processing}              & \begin{tabular}[c]{@{}l@{}}{\cite{7399696}}\\\hline {\cite{Renault}, \cite{Mereuta}, \cite{6984236}, \cite{6612229}}\\\hline {\cite{7565185}, \cite{7150324,7524427}, \cite{7336366}, \cite{7284356,7492645,6733207},\cite{6877244}, \cite{7389318},\cite{179401}, \cite{Perino}, \cite{7152629}, \cite{7593221}}\\\hline {\cite{7576619}, \cite{7295892}}\\\hline {\cite{7353181}, \cite{6689479}}\end{tabular} & \begin{tabular}[c]{@{}l@{}}Energy-efficient orchestrators\\\hline Checkpointing aided parallel execution (CAPE)\\\hline Reducing the amount of exchanged data\\\hline Cloudlets\\\hline Virtual machine placement for energy efficiency\end{tabular} \\ \hline
\end{tabular}
\end{table*}

\begin{table*}[htbp!]
\centering
\caption{Different Green Solutions Approaches}
\label{approaches}
\begin{tabular}{|
l |
l |
l |}
\hline
Green Data Management Solutions                                                                                                                                                                                                  & Green Architectural Solutions                                                                                                                                                            & Green Software Solutions                                                                                                                                                                    \\ \hline
\begin{tabular}[c]{@{}l@{}}Minimizing number of relay transmissions\\\hline\\ Removing redundant transmission links\\\hline\\ Data compression techniques\\\hline\\ Traffic engineering\\\hline\\ Reducing the amount of exchanged data\end{tabular} & \begin{tabular}[c]{@{}l@{}}Dynamic voltage and frequency scaling\\\hline\\ Traffic Engineering\\\hline\\ Energy-efficient orchestrators\\\hline\\ Cloudlets\\\hline\\ Virtual machine placement\end{tabular} & \begin{tabular}[c]{@{}l@{}}Traffic Engineering\\\hline\\ Reducing the amount of exchanged data\\\hline\\ Data compression techniques \\\hline\\ Checkpointing aided parallel execution (CAPE)\end{tabular} \\ \hline
\end{tabular}
\end{table*}

\section{CPS Big Data Applications}
\label{CPS_app}
\begin{figure*}[htbp!]
  \begin{center}
    \includegraphics[scale = 0.16]{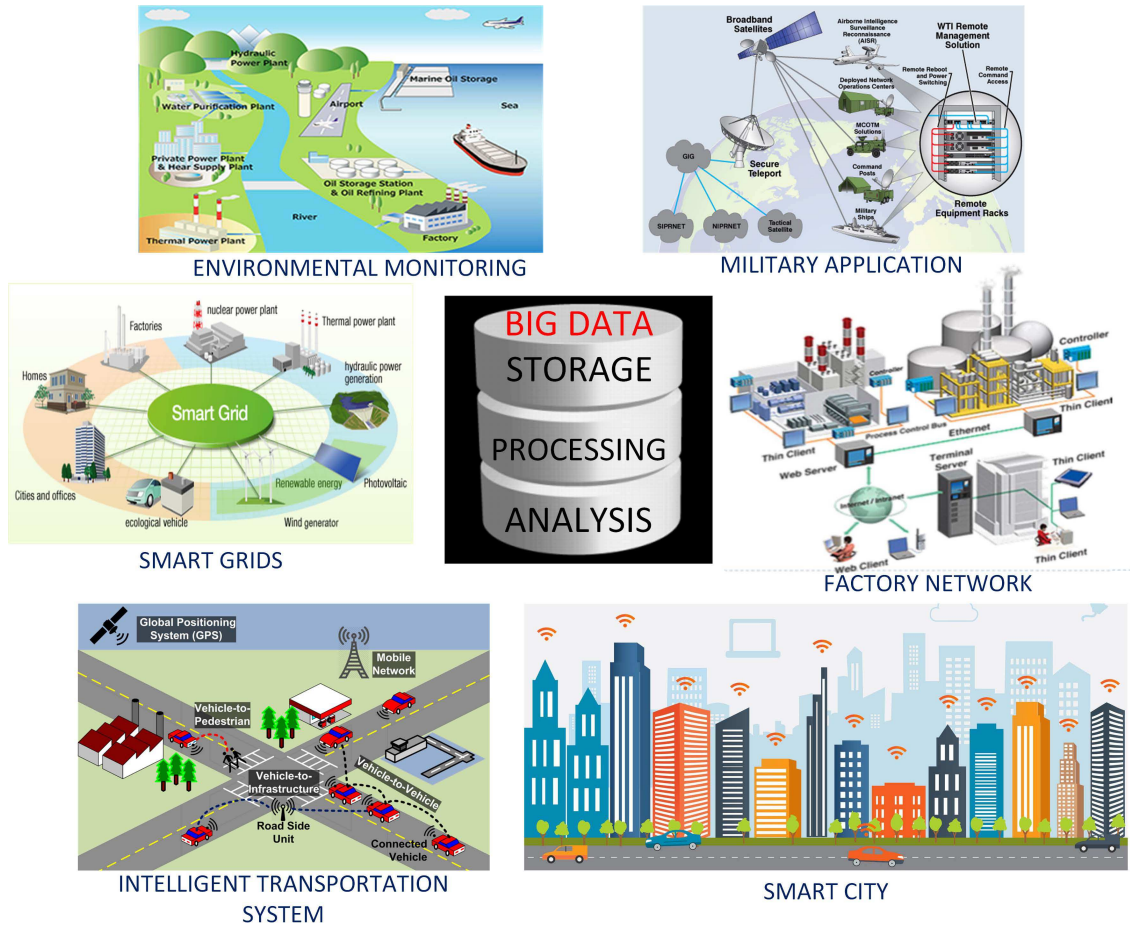}
    \vspace{-0.2cm}
    \caption{\label{cps_bigdata} Big data meet CPS applications (adapted from~\cite{hamida2015security,ge3s,defense,facilities,nanoworks,google}).}
  \end{center}
\end{figure*}

We now present some of the main CPS big data applications in different fields{, such as} energy utilization, city management, and disaster events applications, along with a public safety case study model. Fig.~\ref{cps_bigdata} shows different CPS applications and their big data generation. For instance, the intelligent transportation system would generate big data consisting of driver's behavior, passenger information, vehicles' locations, traffic signals management, accidents' reporting, automated fare calculations, and so on. Each one of the CPS applications {generate a} large amount of data that {need} to be stored, processed and analyzed in order to improve services and applications' performance.

\begin{figure*}[htbp!]
  \begin{center}
    \includegraphics[scale = 0.6]{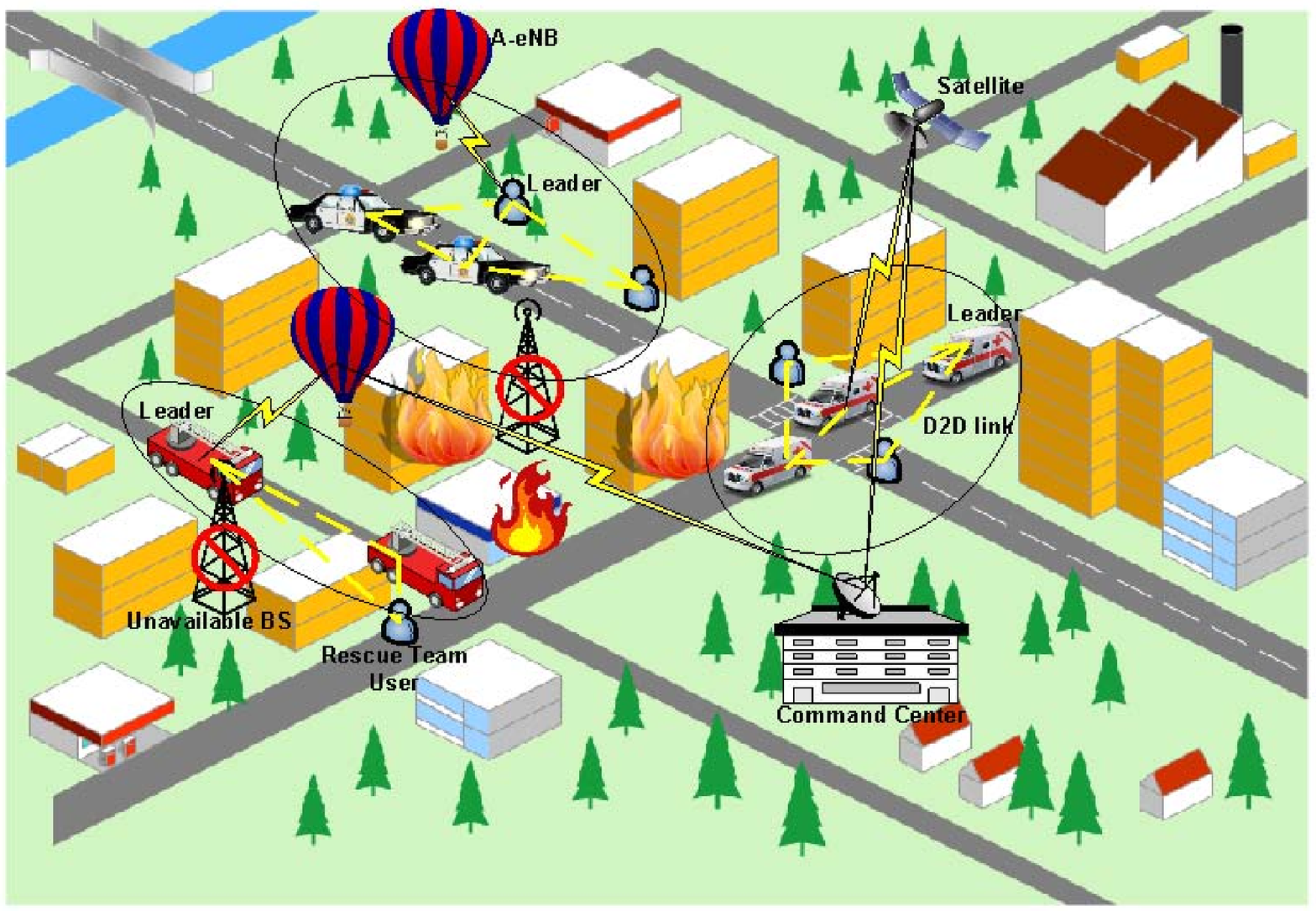}
    \vspace{-0.2cm}
    \caption{\label{public_safety} A public safety model for CPS.}
  \end{center}
\end{figure*}
\subsection{Smart Grids} Smart grids constitute an important aspect of sustainable energy utilization and are becoming more popular, especially with the advances in sensing and signal processing technologies. Automated smart decisions based on millions of data and control points play an important role in managing the energy usage patterns, understanding users' behaviors, reducing the need to build power plants, and addressing supply fluctuations by using  renewable resources~\cite{6475927}. The large number of embedded power generator sensors and their communications with different home sensors and appliances are expected to generate a large amount of data. These advanced sensing and control technologies used in smart grids are often limited to a small region such as a city; however they are envisioned to be deployed on a much larger scale such as the whole country. This will introduce several challenges, among which information management, processing and analysis are the main ones~\cite{6905754}. These big data tasks get even more complicated with the increasing number of transactions that need to be processed for millions of customers. For instance, one smart grid utility is expected to handle two million customers with 22 gigabytes of data per day~\cite{6905754}. That is why big data tools from cloud computing~\cite{6475927}, mining and analytics~\cite{6873776,7741444}, performance optimization~\cite{7069405} and others have been dedicated for smart grids applications.

Moreover, for reliable power grid, smart grids highly depend on cyber infrastructure. This poses several challenges such as exposing the physical operations of smart grid systems to cyber security attacks~\cite{6279584}. Furthermore, the collection of users' energy usage information such as the types of appliances they use, the eating/sleeping patterns, and so on, can be very beneficial in optimizing smart grids' performance; however users' privacy can also be affected. In~\cite{7342880}, Yassine \textit{et al.} proposed a game theoretic mechanism to balance between users' private information and the  beneficial uses of data. In~\cite{7154500}, a big data architecture for smart grids based on random matrix theory was proposed to conduct high dimensional analysis, identify data correlations, and manage data and energy flows among utilities. The proposed architecture not only allows large-scale big data analysis but also can be used as an anomaly detection tool to detect security flaws in smart grids.

For security threats that occur in short time, a security situational awareness technique was proposed in~\cite{7587350}, which uses fuzzy cluster analysis based on game theory and reinforcement learning to enhance security in smart grids. A big data computing architecture for smart grids was proposed in~\cite{7909159} with four main elements: 1) data resource where smart grids data is generated by different devices, companies and systems with complex correlated relationship among them, 2) data storage where only meaningful information is stored and processed with in stream mode or batch mode, 3) data analysis using demand-side management or load forecasting for purposes of categorizing the total demand response in a specific region, and 4) data transmission which bridges the previous elements together. Using this architecture  with energy scheduling scheme based on game theory and a generic algorithm-based optimization to obtain the optimal deployment of energy storage devices for each customer, the results show a significant reduction in total costs of consumers over long term. In \cite{J_TII_KHam_11_2017}, the combined temporal encoding, delayed feedback networks (DFNs) reservoir computing (RC) implementation, and a multi-layer perceptron (MLP) were used to execute effective attack detection for smart grid networks.

\subsection{Military Applications}
Big data can also be exploited to improve military experiences, services, and training. Real-time authentication of command and control messages in cyber-physical infrastructures is of high importance for military services to ensure security. In~\cite{6882222}, the authors developed a novel broadcast authentication scheme using special digital signatures for faster signature generation and verification, and packet loss tolerance. This can be useful to efficiently and rapidly secure military communications. In~\cite{7028577}, the authors used  Markov decision process to propose an approach to identify and reduce attacks' cost in military operations in order to protect important information through obtaining attack policies. Military satellite communications require being resilient to ensure missions' success. This can be achieved using matrix-based protection assessment approach based on traditional risk analysis, where an attack can be assessed in terms of both ease of attack and impact of attack~\cite{6735756}. Mitigating the following five core threats allows the satellite communications to be free from weak vulnerabilities that can be easily exploited by attackers: waveform, RF access to enemy, foreign presence, physical access, and traffic concentration~\cite{6735756}.

\subsection{City Management} Big data can facilitate daily activities by using smart infrastructures and services. With the increasing number of sensors deployed in in urban environment either indoor or outdoor, from smart phones, smart cards, on-board vehicle sensors and so on, the city is faced with a large amount of information that need to be exploited to detect different urban dynamic patterns~\cite{7558230}. For instance, traffic patterns can be analyzed and routes can be computed to allow people to reach their destinations faster. In~\cite{7400610}, the authors proposed to deploy road sensors to obtain information on the overall traffic, such as speed and location of individual vehicles. This information is then processed using graph algorithms by taking advantage of big data tools such as Giraph, Spark and Hadoop. This helps provide real-time intelligent decisions for smart efficient transportation. Since wireless sensor networks (WSN) are the main infrastructures for smart cities to monitor and gather information from the environment, several works have focused on extending network's lifetime. For instance in~\cite{7797384}, the authors proposed using software-defined networking (SDN) controller to reduce WSN traffic and improve decisions making. In~\cite{7478010}, compressive sensing and consensus filter were used to reconstruct signals from fewer sensor nodes, leading to power savings. Sleep scheduling with renewable energy resources such as solar harvesting were also suggested in~\cite{7828559. Rachad_Harvest} to prolong WSN lifetime. CPS based big data may also support localization applications\cite{J_IEICE_TCOM_GDing_08_2014,J_IEICETCOM_GDing_03_2015}.

An IoT-based general architecture for smart cities was presented in~\cite{7558230}, which consists of four different layers: 1) technologies layer consisting of self-configured and remotely-controlled sensors and actuators, 2) middleware layer where data from different sensors are collected to provide context information, 3) management layer where different data analytic tools are used to extract information, test hypothesis and draw conclusions, and 4) services layer consisting of services provided by smart cities based on the previous layers such as environmental monitoring, energy efficiency in buildings, intelligent transportation systems an so on.

Safety systems, such as deploying surveillance systems, are other city management big data applications. A computer vision deep learning algorithm for human activity recognition was proposed in~\cite{7520541}. The model is capable of recognizing twelve types of human activities with high accuracy and without the need of prior knowledge, which makes it useful for security monitoring applications. Crowd detection and surveillance is another safety system big data application. A target individual needs to be easily inferred from visual information. In~\cite{7379452}, the authors proposed such a framework that can easily detect target location and update the motion information to improve the detection.

\subsection{Medical Applications}
CPS health systems are foreseen to shape the future of tele-medicine in different areas such as cardiology, surgery, patients' health monitoring, which will significantly enhance the healthcare system by providing timely, efficient and effective medical decisions for a myriad of health applications such as diabetes management, blood pressure and heart rhythm monitoring, elderly support and so on~\cite{5306098}. With 774 million connected health-related devices~\cite{s141018009} by 2020, a large volume of data from small-scale networks, such as e-health systems or mobile-health systems, needs to be stored, processed and analyzed to enable timely intervention and better management of patients' health.

With e-health systems becoming widely deployed in hospitals and health centers, research has been focused on efficiently deploying medical body area networks (MBANs) to reduce interference on medical bands from other devices~\cite{Yuce2010116}. In MBANs, biomedical sensors are placed in the vicinity of patient's body or even inside her to sense health-related vital signals using short-range wireless technologies. The collected data {are} then multi-hopped to remote stations, so that medical staff can efficiently monitor patients' physiological conditions and disease progression~\cite{Yuce2010116}. For instance, in~\cite{5681102}, elderly patients' health tracking application was proposed, where a mixed positioning algorithm allows for 24-hour monitoring of patients' activities and transmits an alarm to medical staff through SMS, e-mail or telephone in case of an abnormal event or emergency. However, transmitting this health information in a timely and energy-efficient manner is of utmost importance for e-health systems. In~\cite{Martinez2011485}, a micro Subscription Management System ($\mu$SMS) middleware for e-health systems was presented. The $\mu$SMS platform allows sensor nodes to exchange information to provide event-driven services with dynamic memory and variable payload such as GPS coordinates, Home Context and so on. The designed architecture achieves lower memory overhead, lower software components load time and lower event propagation time than other similar proposals, which are all critical requirements for energy efficiency, reliability and scalability of e-health systems. The paper \cite{6775278} proposed locality sensitive hashing to learn sensor patterns for monitoring health conditions of dispersed users.

\subsection{Disaster Events Applications}
Network resilience and survivability are the utmost requirements for public safety networks. In case of a disaster or emergency event, the people who are first on scene are referred to as first responders, and they include law enforcement, firefighters, medical personnel and others~\cite{6599064}. Some of the major public safety requirements relate to the necessity of first responders to exchange information (voice and/or data) in a timely manner~\cite{6599064}. The big data can be used to support disaster events, such as analyzing big data from high resolution maps, floor plans and on-field video transmissions to transmit warning messages to authorities~\cite{7462716}. The remote sensing big data can be analyzed using a scalable hybrid parallelism approach to reduce the analytics execution time~\cite{7730540}. The large amount of data collected from previous earthquakes can be used to predict the future service availability areas, which can improve {preparation} and response to such events~\cite{7471282}. A disaster domain-specific search engine can be constructed using big data to make the understanding and {preparation} of disaster attacks easier and faster for authorities~\cite{6846628}. \\

\subsection{A Public Safety Case Study Model} Fig.~\ref{public_safety} depicts a public safety (PS) model, where the base station is unavailable due to network failure, indoor or underground communications or cell-edge locations. This fact necessitates that users form a device-to-device (D2D) network autonomously without assistance of any network infrastructure. In this model, each D2D cluster sends its PS-related data findings { such as maps, videos, pictures and so on} to a command center for data analysis. This can be done through satellites or by raising in the sky a balloon with attached 4G eNodeB (AeNB) to restore temporarily connectivity (Project ABSOLUTE~\cite{ABSOLUTE}). { LTE is suggested as a technology enabler for PS communications~\cite{6619579}. LTE provides high rate and very low latency IP connectivity. LTE can complement existing Private Mobile Radio (PMR) networks like TETRA/TETRAPOL/Project25. In terms of end-to-end latency requirement, messages are expected to be delivered within 5-10 milliseconds (ms) with a reliability of 99.99\% of packets successfully delivered within a time window~\cite{flynn5etsi,lucent2010government}. Moreover, the downlink peak data rates are expected to be over 50 Mbps with an uplink data rate of over 25 Mbps~\cite{witkowski2013effectively}.} The command center should be able to analyze different data types and then transmit timely, useful and accurate dynamic messages to each D2D cluster head. {The command center can transmit information such as assessment of the level of assistance needed, emergency notification messages, emergency service providers coordination, and so on.} Each D2D cluster head in turn multicasts this information to its members for decision making. This poses a challenge: some users might be far away from the multicast transmitter or their link quality suffers, which would decrease the coverage probability of the receiver and might require several re-transmissions to send the message successfully. Obviously, there is a trade-off between throughput (or end-to-end delay) and reliability due to the random transmission errors caused by the unpredictable behavior of the wireless channel.  It is evident that we need to minimize delay, while maintaining some good degree of reliability. In what follows, we briefly derive the end-to-end delay and reliability for public safety networks.

Let $\delta_i$ be the indicator variable defined as:
$$
\delta_i = \left\{
    \begin{array}{ll}
        1 & \mbox{if packet is successfully received by node }$$i$$ \\
        0 & \mbox{otherwise.}
    \end{array}
\right.
$$
The set of nodes that have received successfully in the $j^{\text{th}}$ round of multicasting can be defined as:
\begin{equation}
N^\star_j=\{i \mid \delta_i=1\}.
\end{equation}

In order to minimize the total delay, $D$, for distributing the data to all nodes, $L-1$ ($L$ is the total number of nodes), we need to select node $k^\star_j$ to multicast in the $j^{\text{th}}$ round of multicasting such that:
\begin{equation}
\label{eq1}
k^\star_j=arg\max_{i\in N^\star_j}\{\min_{l \notin N^\star_j} R_{i,l}\},
\end{equation}
where $ R_{i,l}$ is the transmission rate between nodes $i$ and $l$. Define the set of unvisited nodes as $\Pi_u$ and the set of the visited nodes as $\Pi_v$
Initially, we have $\Pi_\text{v}=1$, $\Pi_\text{u}= L-1$, and $D=0$. In the $j^{\text{th}}$ round of multicasting by node $k^\star_j$, $D$ gets updated by adding to it the time of last node receiving the packet successfully as:
\begin{equation}
D=D+\max\limits_{i\in \bar{N}_j}t_i,
\end{equation}
where $\bar{N}_j=\{i\in \Pi_u|\delta_i=0\}$, and $t_i$ is the minimum access time of node $i$. The multicasting rounds continue until either i) all unvisited nodes become visited, i.e., $\Pi_\text{u}=\Pi_\text{v}$, or ii) When $D$ exceeds a threshold $\theta_D$, in that case the packet is dropped.

Reliability can be defined as the ability of the network to perform its required functions for a certain period of time without service interruption. A link breakage can occur due to the following reasons:
\begin{itemize}
\item The receiver is outside of the transmission range of the transmitter
\item The packet is lost due to channel errors
\item The packet is lost due to total delay $D$ exceeding threshold $\theta_D$.
\end{itemize}

Assuming users are not moving, as in~\cite{6817780}, each D2D receiver is randomly and independently located around its corresponding transmitter with isotropic direction and Rayleigh distributed distance with probability density function (PDF):
\begin{equation}
\label{PDF}
f_\text{D}(r)=2\pi\lambda r e^{-\lambda \pi r^2},\text{  } r\geq 0.
\end{equation}

The probability that a user is outside the transmission range $R$ of the transmitter can be calculated as $p_\text{R}=1-\int_0^R{f_\text{D}(r)dr}=e^{-\pi\lambda R^2}$.

Considering log-normal shadow fading, the signal fades:  (1)  deterministically with a path-loss  exponent $\alpha$,  and  (2) stochastically  represented by  a  random  variable  with  zero  mean and a variance $\sigma$. Therefore, the probability $p_\text{e}$ that a packet
is not lost due to channel errors is given in~\cite{6093225}:
\begin{equation}
p_\text{e}=1-\frac{1}{2}+\frac{1}{2}\erf\left(\frac{\beta_\text{th}-\alpha\times 10\log(r)}{\sqrt{2}\sigma}\right),
\end{equation}
where $\beta_\text{th}$ is the minimum threshold required to deliver a packet between nodes.

Finally, we combine what we have discussed to get the link reliability as:
\begin{equation}
R_\text{L}=p_\text{e}+p_\text{R}+p_\text{drop}.
\end{equation}
where $p_\text{drop}=\mathcal{P}\left(D>\theta_\text{D}\right)$.

\section{Big Data Challenges and Open Issues for CPS}
\label{challenges}
While ongoing research is focusing on CPS enterprise development and applications, effective solutions to combat security flaws have not received the enough attention, which places question marks on the foreseen integration of CPS in critical infrastructures. This matter is made worse with CPS devices having i) computational challenges in ensuring data confidentiality and privacy protection; ii) the semi- or fully-autonomous security management~\cite{7345413}; and iii) the high computational costs of employing cryptography~\cite{7417841}. Low-complexity and lightweight ciphers such as PRESENT~\cite{6386942} have been developed; however research efforts should go beyond cryptography, especially that such solutions can be costly in terms of i) latency, ii) power consumption, and iii) key management complexity~\cite{7101220,6831320}. Furthermore, different CPS applications might have different security perimeters with a multitude of interactions among devices. This complicates further the access control decisions, the trustworthiness of entities, and their authentication management. Therefore, there is a high need to define secure interoperation protocols and strategies for dynamic CPS, where devices from different applications interact together to complete specific tasks~\cite{rayaccess}.

{ Privacy preserving is an important aspect of big data analytics and mining. Homomorphic encryption techniques, as mentioned in Section~\ref{sec_analy}, have been suggested to perform analysis on encrypted data in cloud. However, they turned out to be impractical and inefficient in terms of computational time and overhead for large amount of datasets. {Although anonymization techniques constitute a more efficient solution for privacy protection big data analytics, further research efforts are} needed to determine their applicability for different big data types, and their performance evaluation and endurance in different mining techniques as outlined in Section~\ref{miningg}. Moreover, future solutions should be based on the hierarchy property of many CPS big data applications~\cite{8013260}. This means that future research should be directed towards proposing integrated privacy preserving protocols in the modules of the CPS hierarchical structure that can perform analytical processes on encrypted variety of big data types in a time- and cost-efficient manner.}

Correctness of CPS is another research area under the spotlight of attention. Due to dynamic nature of physical environment, CPS need to constantly be adapted to new situations while operating and functioning properly with little or no human supervision~\cite{7562128}. The use of models can allow the early detection of failures by simulating different components of complex designs to verify the integration of the whole system~\cite{Cancila,4606742}. CPS components verification to ensure they are working properly or that they meet execution time requirements is another approach to ensure correctness~\cite{6843708,7167224,7423131}. Future research shall focus on creating robust real-time anomaly detection and correctness techniques that have low overhead and implementation costs.

With CPS being employed on a large scale, data access, routing, transmissions, and processing might consume a significant amount of time, leading to real obstacles in the way of realizing  ultra-reliable and low latency communications (URLLC), a key feature for emerging 5G technologies. In URLLC, stringent requirements are placed on latency, throughput, and availability. Future research should be directed towards faster data access from the clouds with fewer data re-transmissions in order to free up resources and reduce latency. Obviously, there is a trade-off between throughput (or end-to-end delay) and reliability due to transmission errors that might be caused by links failure, security policies, caching misses, scarcity of resources and so on. For instance, cross-layer designs protocols between different CPS layers (sensing, middleware, transmission, management and services layers) can be used in a similar way to the wireless communication protocols to help optimize different parameters from latency to reliability and security. An example would be the transmission layer communicating with the sensing layer about integrating data filtering aimed at prioritizing the sensed data in the case of too many re-transmissions.

For faster data collection, research efforts need to shift towards devices that do not need to preconfigure to a network with dynamic on-the-fly D2D connectivity, and without the need of controllers or infrastructure deployment. Furthermore, to extend the communication range among devices, incentivized devices with tokens will replace dedicated relays~\cite{Mastronarde_D2D}. In terms of CPS computing, research directions should shift towards faster data processing via moving data processing closer to the sources and speeding-up big data handling. For a faster and efficient CPS analysis, research activities need to be further conducted on information fusion techniques, especially that information originates from heterogeneous devices with varying capabilities. Through perfecting the fusion algorithms, related information can be aggregated for analysis leading to higher quality information~\cite{Goodman:1997:MDF:549931}.

As mentioned in this survey, big data analytics is one of the most important aspects of CPS, as it helps to pave the way for new opportunities and services, in addition to enhancing or even optimizing systems' performance. However, as discussed in Section~\ref{data_sources}, the sensed data can originate from a variety of sources with unstructured formats. Merging these different data types into single homogeneous structures is vital to ease and speed up the analysis process so meaningful conclusions can be drawn and automated decisions can be made~\cite{Ali2016}. How to homogenize different data sources remains an open research issue for CPS. Moreover, different CPS applications might need to collaborate together to achieve a specific mission. For instance, the mobile-health application might need to collaborate with the transportation system to get an ambulance as fast as possible in the case of patient's biomedical sensors revealing an emergency situation. In such scenarios, data analytics becomes a challenging task as it requires to be able to put the different analysis fragments from different CPS applications together to provide broader conclusions and decisions making.

\section{Conclusion}
\label{conclusion}
The emerging CPS technologies mutually benefit the technological advancements in big data processing and analysis. When combined with artificial intelligence, machine learning, and neuromorphic computing techniques, CPS will bring about new applications, services and opportunities, all envisioned to be automated with little or {without} human intervention. This will help revolutionize the ``smart planet'' concept, where smarter water management, smarter healthcare, smarter transportation, smarter energy, and smarter food will create a radical shift in our lives. In this survey paper, we have provided a broad overview of CPS big data collection, storage, access, caching, routing, processing, and analysis to support understanding and discovering the challenges facing CPS, the existing proposed solutions, and the open issues that are yet to be addressed. Then, we have discussed the security vulnerabilities of CPS and the different security solutions, as well as how big data meet green challenges for CPS systems to address sustainability and environmental concerns. Finally, some challenges and open issues for CPS have been addressed.


%





\ifCLASSOPTIONcaptionsoff
  \newpage
\fi


%



%
\bibliographystyle{IEEEtran}
\bibliography{IEEEabrv,references}

\begin{thebibliography}{100}
\providecommand{\url}[1]{#1}
\csname url@samestyle\endcsname
\providecommand{\newblock}{\relax}
\providecommand{\bibinfo}[2]{#2}
\providecommand{\BIBentrySTDinterwordspacing}{\spaceskip=0pt\relax}
\providecommand{\BIBentryALTinterwordstretchfactor}{4}
\providecommand{\BIBentryALTinterwordspacing}{\spaceskip=\fontdimen2\font plus
\BIBentryALTinterwordstretchfactor\fontdimen3\font minus
  \fontdimen4\font\relax}
\providecommand{\BIBforeignlanguage}[2]{{%
\expandafter\ifx\csname l@#1\endcsname\relax
\typeout{** WARNING: IEEEtran.bst: No hyphenation pattern has been}%
\typeout{** loaded for the language `#1'. Using the pattern for}%
\typeout{** the default language instead.}%
\else
\language=\csname l@#1\endcsname
\fi
#2}}
\providecommand{\BIBdecl}{\relax}
\BIBdecl

\bibitem{RAtat_thesis_2017}
R.~Atat, ``Enabling cyber-physical communication in {5G} cellular networks:
  Challenges, solutions and applications,'' PhD Thesis, University of Kansas,
  USA, 2017.

\bibitem{cisco}
{CISCO}, ``Fog computing and the internet of things: extend the cloud to where
  the things are,'' white paper, {CISCO}, Tech. Rep., 2015.

\bibitem{lee2015cyber}
J.~Lee, B.~Bagheri, and H.-A. Kao, ``A cyber-physical systems architecture for
  industry 4.0-based manufacturing systems,'' \emph{Manufacturing Letters},
  vol.~3, pp. 18--23, 2015.

\bibitem{muhonen2015standardization}
T.~Muhonen, ``Standardization of industrial internet and iot (iot--internet of
  things)--perspective on condition-based maintenance,'' \emph{University of
  Oulu, Finland}, 2015.

\bibitem{6674155}
C.~W. Tsai, C.~F. Lai, M.~C. Chiang, and L.~T. Yang, ``Data mining for internet
  of things: A survey,'' \emph{IEEE Communications Surveys Tutorials}, vol.~16,
  no.~1, pp. 77--97, First 2014.

\bibitem{7473821}
J.~Wu, S.~Guo, J.~Li, and D.~Zeng, ``Big data meet green challenges: Greening
  big data,'' \emph{IEEE Systems Journal}, vol.~10, no.~3, pp. 873--887, Sept
  2016.

\bibitem{wu2018information}
J.~Wu, S.~Guo, H.~Huang, W.~Liu, and Y.~Xiang, ``Information and communications
  technologies for sustainable development goals: State-of-the-art, needs and
  perspectives,'' \emph{IEEE Communications Surveys \& Tutorials}, vol.~20,
  no.~3, pp. 2389--2406, 2018.

\bibitem{6829939}
J.~Wu, I.~Bisio, C.~Gniady, E.~Hossain, M.~Valla, and H.~Li, ``Context-aware
  networking and communications: Part 1 [guest editorial],'' \emph{IEEE
  Communications Magazine}, vol.~52, no.~6, pp. 14--15, June 2014.

\bibitem{6512846}
C.~Perera, A.~Zaslavsky, P.~Christen, and D.~Georgakopoulos, ``Context aware
  computing for the internet of things: A survey,'' \emph{IEEE Communications
  Surveys and Tutorials}, vol.~16, no.~1, pp. 414--454, First 2014.

\bibitem{7565634}
M.~Chi, A.~Plaza, J.~A. Benediktsson, Z.~Sun, J.~Shen, and Y.~Zhu, ``Big data
  for remote sensing: Challenges and opportunities,'' \emph{Proceedings of the
  IEEE}, vol. 104, no.~11, pp. 2207--2219, Nov 2016.

\bibitem{7470600}
B.~Zhang, Z.~Zhang, Z.~Ren, J.~Ma, and W.~Wang, ``Energy-efficient
  software-defined data collection by participatory sensing,'' \emph{IEEE
  Sensors Journal}, vol.~16, no.~20, pp. 7315--7324, Oct 2016.

\bibitem{7406686}
Y.~Sun, H.~Song, A.~J. Jara, and R.~Bie, ``Internet of things and big data
  analytics for smart and connected communities,'' \emph{IEEE Access}, vol.~4,
  pp. 766--773, 2016.

\bibitem{7402272}
B.~Guo, C.~Chen, D.~Zhang, Z.~Yu, and A.~Chin, ``Mobile crowd sensing and
  computing: when participatory sensing meets participatory social media,''
  \emph{IEEE Communications Magazine}, vol.~54, no.~2, pp. 131--137, February
  2016.

\bibitem{MChenn}
M.~Chen, S.~Mao, and Y.~Liu, ``Big data: A survey,'' \emph{Mobile Networks and
  Applications}, vol.~19, no.~2, pp. 171--209, Apr 2014.

\bibitem{Derbeko20161}
P.~Derbekoa, S.~Dolevb, E.~Gudesb, and S.~Sharmab, ``Security and privacy
  aspects in mapreduce on clouds: A survey,'' \emph{Computer Science Review},
  vol.~20, no.~1, pp. 1--28, May 2016.

\bibitem{Fan:2013:MBD:2481244.2481246}
\BIBentryALTinterwordspacing
W.~Fan and A.~Bifet, ``Mining big data: Current status, and forecast to the
  future,'' \emph{SIGKDD Explor. Newsl.}, vol.~14, no.~2, pp. 1--5, Apr. 2013.
  [Online]. Available: \url{http://doi.acm.org/10.1145/2481244.2481246}
\BIBentrySTDinterwordspacing

\bibitem{6842585}
H.~Hu, Y.~Wen, T.~S. Chua, and X.~Li, ``Toward scalable systems for big data
  analytics: A technology tutorial,'' \emph{IEEE Access}, vol.~2, pp. 652--687,
  2014.

\bibitem{zadrozny2013big}
P.~Zadrozny and R.~Kodali, \emph{Big data analytics using {Splunk}: Deriving
  operational intelligence from social media, machine data, existing data
  warehouses, and other real-time streaming sources}.\hskip 1em plus 0.5em
  minus 0.4em\relax Apress, 2013.

\bibitem{6974788}
M.~Wang, B.~Li, Y.~Zhao, and G.~Pu, ``Formalizing google file system,'' in
  \emph{Proc. 2014 IEEE 20th Pacific Rim International Symposium on Dependable
  Computing}, Nov 2014, pp. 190--191.

\bibitem{2016:SPA:2974459.2974485}
\BIBentryALTinterwordspacing
P.~Derbekoa, S.~Dolevb, E.~Gudesb, and S.~Sharmab, ``Security and privacy
  aspects in mapreduce on clouds,'' \emph{Comput. Sci. Rev.}, vol.~20, no.~1,
  pp. 1--28, May 2016. [Online]. Available:
  \url{http://dx.doi.org/10.1016/j.cosrev.2016.05.001}
\BIBentrySTDinterwordspacing

\bibitem{Atzori}
\BIBentryALTinterwordspacing
L.~Atzori, A.~Iera, and G.~Morabito, ``The internet of things: A survey,''
  \emph{Comput. Netw.}, vol.~54, no.~15, pp. 2787--2805, Oct. 2010. [Online].
  Available: \url{http://dx.doi.org/10.1016/j.comnet.2010.05.010}
\BIBentrySTDinterwordspacing

\bibitem{Acharjya}
D.~Acharjya and K.~Ahmed, ``A survey on big data analytics: Challenges, open
  research issues and tools,'' \emph{International Journal of Advanced Computer
  Science and Applications}, vol.~7, no.~2, pp. 511--518, Feb 2016.

\bibitem{Bellavista}
\BIBentryALTinterwordspacing
P.~Bellavista, A.~Corradi, M.~Fanelli, and L.~Foschini, ``A survey of context
  data distribution for mobile ubiquitous systems,'' \emph{ACM Computing
  Surveys}, vol.~44, no.~4, pp. 24:1--24:45, Sep. 2012. [Online]. Available:
  \url{http://doi.acm.org/10.1145/2333112.2333119}
\BIBentrySTDinterwordspacing

\bibitem{7065282}
X.~Zhang, Z.~Yang, W.~Sun, Y.~Liu, S.~Tang, K.~Xing, and X.~Mao, ``Incentives
  for mobile crowd sensing: A survey,'' \emph{IEEE Communications Surveys
  Tutorials}, vol.~18, no.~1, pp. 54--67, Firstquarter 2016.

\bibitem{7087016}
S.~Tamboli and S.~S. Patel, ``A survey on innovative approach for improvement
  in efficiency of caching technique for big data application,'' in
  \emph{Pervasive Computing (ICPC), 2015 International Conference on}, Jan
  2015, pp. 1--6.

\bibitem{BRao}
B.~Rao and L.~Reddy, ``Survey on improved scheduling in {Hadoop MapReduce} in
  cloud environments,'' \emph{International Journal of Computer Applications},
  vol.~34, no.~9, pp. 29--33, Nov 2011.

\bibitem{SYI}
S.~Yi, Z.~Qin, and Q.~Li, ``Security and privacy issues of fog computing: a
  survey,'' in \emph{Wireless Algorithms, Systems, and Applications}.\hskip 1em
  plus 0.5em minus 0.4em\relax Springer International Publishing, 2015, pp.
  685--695.

\bibitem{Wang:2013:SCS:2459506.2459606}
\BIBentryALTinterwordspacing
W.~Wang and Z.~Lu, ``Survey cyber security in the smart grid: Survey and
  challenges,'' \emph{Comput. Netw.}, vol.~57, no.~5, pp. 1344--1371, Apr.
  2013. [Online]. Available:
  \url{http://dx.doi.org/10.1016/j.comnet.2012.12.017}
\BIBentrySTDinterwordspacing

\bibitem{beloglazov2011a}
A.~Beloglazov, R.~Buyya, Y.~C. Lee, A.~Zomaya, and Others, ``{A taxonomy and
  survey of energy-efficient data centers and cloud computing systems},''
  \emph{Advances in Computers}, vol.~82, no.~2, pp. 47--111, 2011.

\bibitem{7399696}
Z.~Asad and M.~A.~R. Chaudhry, ``A two-way street: Green big data processing
  for a greener smart grid,'' \emph{IEEE Systems Journal}, vol.~PP, no.~99, pp.
  1--11, 2016.

\bibitem{7430133}
C.~Estevez and J.~Wu, ``Recent advances in green internet of things,'' in
  \emph{Proc. 2015 7th IEEE Latin-American Conference on Communications
  (LATINCOM)}, Nov 2015, pp. 1--5.

\bibitem{7473815}
J.~Wu, S.~Guo, J.~Li, and D.~Zeng, ``Big data meet green challenges: Big data
  toward green applications,'' \emph{IEEE Systems Journal}, vol.~10, no.~3, pp.
  888--900, Sept 2016.

\bibitem{6096958}
J.~Shi, J.~Wan, H.~Yan, and H.~Suo, ``A survey of cyber-physical systems,'' in
  \emph{Proc. 2011 International Conference on Wireless Communications and
  Signal Processing (WCSP)}, Nov 2011, pp. 1--6.

\bibitem{Chaari2016260}
\BIBentryALTinterwordspacing
R.~Chaari, F.~Ellouze, A.~Koubaa, B.~Qureshi, N.~Pereira, H.~Youssef, and
  E.~Tovar, ``Cyber-physical systems clouds: A survey,'' \emph{Computer
  Networks}, vol. 108, pp. 260 -- 278, 2016. [Online]. Available:
  \url{http://www.sciencedirect.com/science/article/pii/S1389128616302699}
\BIBentrySTDinterwordspacing

\bibitem{gunes2014survey}
V.~Gunes, S.~Peter, T.~Givargis, and F.~Vahid, ``A survey on concepts,
  applications, and challenges in cyber-physical systems,'' \emph{KSII
  Transactions on Internet and Information Systems (TIIS)}, vol.~8, no.~12, pp.
  4242--4268, 2014.

\bibitem{nugroho2016development}
A.~P. Nugroho, T.~Okayasu, M.~Horimoto, D.~Arita, T.~Hoshi, H.~Kurosaki, K.-i.
  Yasuba, E.~Inoue, Y.~Hirai, M.~Mitsuoka \emph{et~al.}, ``Development of a
  field environmental monitoring node with over the air update function,''
  \emph{Agricultural Information Research}, vol.~25, no.~3, pp. 86--95, 2016.

\bibitem{Akdeniz:2000:ILS}
D.~N. Chorafas, \emph{Business, Marketing, and Management Principles for IT and
  Engineering}.\hskip 1em plus 0.5em minus 0.4em\relax Boca Raton, FL, USA:
  Auerbach Publications, June 22 2011.

\bibitem{10.1109/MOBIQ.2006.340411}
L.~Sanchez, M.~Bauer, J.~Lanza, R.~Olsen, and M.~Girod-Genet, ``A generic
  context management framework for personal networking environments,'' in
  \emph{Proc. the 2006 Third Annual International Conference on Mobile and
  Ubiquitous Systems: Networking and Services}.\hskip 1em plus 0.5em minus
  0.4em\relax Los Alamitos, CA, USA: IEEE Computer Society, 2006, pp. 1--8.

\bibitem{doi:10.1080/01431168008948242}
J.~A. ALLAN, ``A review of: Remote sensing: Principles and interpretation by
  {FLOYD F. SABINS, JR. (San Francisco: W. H. Freeman, 1978.)} [pp. 1-426.],''
  \emph{International Journal of Remote Sensing}, vol.~1, no.~3, pp. 307--308,
  1980.

\bibitem{Ma:2015:RSB:2794090.2794274}
\BIBentryALTinterwordspacing
Y.~Ma, H.~Wu, L.~Wang, B.~Huang, R.~Ranjan, A.~Zomaya, and W.~Jie, ``Remote
  sensing big data computing,'' \emph{Future Gener. Comput. Syst.}, vol.~51,
  no.~C, pp. 47--60, Oct. 2015. [Online]. Available:
  \url{http://dx.doi.org/10.1016/j.future.2014.10.029}
\BIBentrySTDinterwordspacing

\bibitem{7486259}
L.~Zhang, L.~Zhang, and B.~Du, ``Deep learning for remote sensing data: A
  technical tutorial on the state of the art,'' \emph{IEEE Geoscience and
  Remote Sensing Magazine}, vol.~4, no.~2, pp. 22--40, June 2016.

\bibitem{7109130}
M.~M.~U. Rathore, A.~Paul, A.~Ahmad, B.~W. Chen, B.~Huang, and W.~Ji,
  ``Real-time big data analytical architecture for remote sensing
  application,'' \emph{IEEE Journal of Selected Topics in Applied Earth
  Observations and Remote Sensing}, vol.~8, no.~10, pp. 4610--4621, Oct 2015.

\bibitem{6897952}
L.~Wang, H.~Zhong, R.~Ranjan, A.~Zomaya, and P.~Liu, ``Estimating the
  statistical characteristics of remote sensing big data in the wavelet
  transform domain,'' \emph{IEEE Transactions on Emerging Topics in Computing},
  vol.~2, no.~3, pp. 324--337, Sept 2014.

\bibitem{7128315}
H.~Xie, X.~Tong, W.~Meng, D.~Liang, Z.~Wang, and W.~Shi, ``A multilevel
  stratified spatial sampling approach for the quality assessment of
  remote-sensing-derived products,'' \emph{IEEE Journal of Selected Topics in
  Applied Earth Observations and Remote Sensing}, vol.~8, no.~10, pp.
  4699--4713, Oct 2015.

\bibitem{7474340}
M.~A. Alsheikh, D.~Niyato, S.~Lin, H.~p.~Tan, and Z.~Han, ``Mobile big data
  analytics using deep learning and apache spark,'' \emph{IEEE Network},
  vol.~30, no.~3, pp. 22--29, May 2016.

\bibitem{7562344}
X.~Zhang, Z.~Yi, Z.~Yan, G.~Min, W.~Wang, A.~Elmokashfi, S.~Maharjan, and
  Y.~Zhang, ``Social computing for mobile big data,'' \emph{Computer}, vol.~49,
  no.~9, pp. 86--90, Sept 2016.

\bibitem{7302079}
M.~Tang, H.~Zhu, and X.~Mao, ``A lightweight social computing approach to
  emergency management policy selection,'' \emph{IEEE Transactions on Systems,
  Man, and Cybernetics: Systems}, vol.~46, no.~8, pp. 1075--1087, Aug 2016.

\bibitem{parameswaran2007social}
\BIBentryALTinterwordspacing
M.~Parameswaran and A.~B. Whinston, ``Social computing: An overview,''
  \emph{Communications of the Association for Information Systems}, vol.~19,
  2007. [Online]. Available: \url{http://aisel.aisnet.org/cais/vol19/iss1/37/}
\BIBentrySTDinterwordspacing

\bibitem{J_FGCS_HNing_0_2016}
H.~Ning, H.~Liu, J.~Ma, L.~T. Yang, and R.~Huang, ``Cybermatics:
  Cyber–physical–social–thinking hyperspace based science and
  technology,'' \emph{Future generation computer systems}, vol.~56, pp.
  504--522, March 2016.

\bibitem{7097702}
S.~Chang, H.~Zhu, W.~Zhang, L.~Lu, and Y.~Zhu, ``Pure: Blind regression
  modeling for low quality data with participatory sensing,'' \emph{IEEE
  Transactions on Parallel and Distributed Systems}, vol.~27, no.~4, pp.
  1199--1211, April 2016.

\bibitem{7565025}
R.~B. Messaoud and Y.~Ghamri-Doudane, ``Fair qoi and energy-aware task
  allocation in participatory sensing,'' in \emph{2016 IEEE Wireless
  Communications and Networking Conference}, April 2016, pp. 1--6.

\bibitem{7563897}
J.~Wang, Y.~Wang, D.~Zhang, L.~Wang, H.~Xiong, S.~Helal, Y.~He, and F.~Wang,
  ``Fine-grained multi-task allocation for participatory sensing with a shared
  budget,'' \emph{IEEE Internet of Things Journal}, vol.~PP, no.~99, pp. 1--1,
  2016.

\bibitem{7044580}
L.~Liu, W.~Wei, D.~Zhao, and H.~Ma, ``Urban resolution: New metric for
  measuring the quality of urban sensing,'' \emph{IEEE Transactions on Mobile
  Computing}, vol.~14, no.~12, pp. 2560--2575, Dec 2015.

\bibitem{7337442}
X.~Sun, S.~Hu, L.~Su, T.~F. Abdelzaher, P.~Hui, W.~Zheng, H.~Liu, and J.~A.
  Stankovic, ``Participatory sensing meets opportunistic sharing: Automatic
  phone-to-phone communication in vehicles,'' \emph{IEEE Transactions on Mobile
  Computing}, vol.~15, no.~10, pp. 2550--2563, Oct 2016.

\bibitem{7355365}
C.~Xiang, P.~Yang, C.~Tian, L.~Zhang, H.~Lin, F.~Xiao, M.~Zhang, and Y.~Liu,
  ``Carm: Crowd-sensing accurate outdoor rss maps with error-prone smartphone
  measurements,'' \emph{IEEE Transactions on Mobile Computing}, vol.~15,
  no.~11, pp. 2669--2681, Nov 2016.

\bibitem{7110391}
L.~Wang, D.~Zhang, Z.~Yan, H.~Xiong, and B.~Xie, ``effsense: A novel mobile
  crowd-sensing framework for energy-efficient and cost-effective data
  uploading,'' \emph{IEEE Transactions on Systems, Man, and Cybernetics:
  Systems}, vol.~45, no.~12, pp. 1549--1563, Dec 2015.

\bibitem{7565185}
E.~Zeydan, E.~Bastug, M.~Bennis, M.~A. Kader, I.~A. Karatepe, A.~S. Er, and
  M.~Debbah, ``Big data caching for networking: moving from cloud to edge,''
  \emph{IEEE Communications Magazine}, vol.~54, no.~9, pp. 36--42, September
  2016.

\bibitem{7150324}
M.~Ji, G.~Caire, and A.~F. Molisch, ``Wireless device-to-device caching
  networks: Basic principles and system performance,'' \emph{IEEE Journal on
  Selected Areas in Communications}, vol.~34, no.~1, pp. 176--189, Jan 2016.

\bibitem{7524427}
K.~Poularakis, G.~Iosifidis, A.~Argyriou, I.~Koutsopoulos, and L.~Tassiulas,
  ``Caching and operator cooperation policies for layered video content
  delivery,'' in \emph{IEEE INFOCOM 2016 - The 35th Annual IEEE International
  Conference on Computer Communications}, April 2016, pp. 1--9.

\bibitem{7336366}
H.~Zhao, K.~Gai, J.~Li, and X.~He, ``Novel differential schema for high
  performance big data telehealth systems using pre-cache,'' in \emph{High
  Performance Computing and Communications (HPCC), 2015 IEEE 7th International
  Symposium on Cyberspace Safety and Security (CSS), 2015 IEEE 12th
  International Conferen on Embedded Software and Systems (ICESS), 2015 IEEE
  17th International Conference on}, Aug 2015, pp. 1412--1417.

\bibitem{7284356}
L.~Lundberg, H.~Grahn, D.~Ilie, and C.~Melander, ``Cache support in a high
  performance fault-tolerant distributed storage system for cloud and big
  data,'' in \emph{Parallel and Distributed Processing Symposium Workshop
  (IPDPSW), 2015 IEEE International}, May 2015, pp. 537--546.

\bibitem{7492645}
S.~G. Kanbargi and S.~K. S, ``Cache utilization for enhancing analyzation of
  big-data increasing the performance of hadoop,'' in \emph{2015 International
  Conference on Trends in Automation, Communications and Computing Technology
  (I-TACT-15)}, vol.~01, Dec 2015, pp. 1--7.

\bibitem{6733207}
Y.~Zhao, J.~Wu, and C.~Liu, ``Dache: A data aware caching for big-data
  applications using the mapreduce framework,'' \emph{Tsinghua Science and
  Technology}, vol.~19, no.~1, pp. 39--50, Feb 2014.

\bibitem{suh2014applied}
S.~C. Suh, U.~J. Tanik, J.~N. Carbone, and A.~Eroglu, \emph{Applied
  cyber-physical systems}.\hskip 1em plus 0.5em minus 0.4em\relax Springer,
  2014.

\bibitem{Wahler:2015:RMC:2737166.2737176}
\BIBentryALTinterwordspacing
M.~Wahler, M.~Oriol, and A.~Monot, ``Real-time multi-core components for
  cyber-physical systems,'' in \emph{Proc. the 18th International ACM SIGSOFT
  Symposium on Component-Based Software Engineering}, ser. CBSE '15.\hskip 1em
  plus 0.5em minus 0.4em\relax New York, NY, USA: ACM, 2015, pp. 37--42.
  [Online]. Available: \url{http://doi.acm.org/10.1145/2737166.2737176}
\BIBentrySTDinterwordspacing

\bibitem{5763145}
B.~Akesson and K.~Goossens, ``Architectures and modeling of predictable memory
  controllers for improved system integration,'' in \emph{Proc. 2011 Design,
  Automation Test in Europe}, March 2011, pp. 1--6.

\bibitem{herkersdorf2013multicore}
A.~Herkersdorf and M.~Paulitsch, ``Multicore enablement for embedded and cyber
  physical systems (dagstuhl seminar 13052),'' \emph{Dagstuhl Reports}, vol.~3,
  no.~1, 2013.

\bibitem{7553025}
J.~Huang, S.~Wang, X.~Cheng, and J.~Bi, ``Big data routing in d2d
  communications with cognitive radio capability,'' \emph{IEEE Wireless
  Communications}, vol.~23, no.~4, pp. 45--51, August 2016.

\bibitem{Mone:2014:IL:2692965.2676393}
\BIBentryALTinterwordspacing
G.~Mone, ``Intelligent living,'' \emph{Commun. ACM}, vol.~57, no.~12, pp.
  15--16, Nov. 2014. [Online]. Available:
  \url{http://doi.acm.org/10.1145/2676393}
\BIBentrySTDinterwordspacing

\bibitem{tibken2015samsung}
S.~Tibken, ``Samsung, smartthings and the open door to the smart home,''
  \emph{cnet CES}, 2015.

\bibitem{3GPP}
{3rd Generation Partnership Project}, ``{3GPP specification: 37.868; {RAN}
  improvements for machine-type communications},'' Tech. Rep., August 2011.

\bibitem{6845044}
K.~Zheng, S.~Ou, J.~Alonso-Zarate, M.~Dohler, F.~Liu, and H.~Zhu, ``Challenges
  of massive access in highly dense {LTE}-advanced networks with
  machine-to-machine communications,'' \emph{IEEE Transactions on Wireless
  Communications}, vol.~21, no.~3, pp. 12--18, June 2014.

\bibitem{6678832}
A.~Laya, L.~Alonso, and J.~Alonso-Zarate, ``Is the random access channel of
  {LTE} and {LTE-A} suitable for {M2M} communications? a survey of
  alternatives,'' \emph{IEEE Communications Surveys Tutorials}, vol.~16, no.~1,
  pp. 4--16, First 2014.

\bibitem{7343681}
M.~Condoluci, L.~Militano, A.~Orsino, J.~Alonso-Zarate, and G.~Araniti,
  ``{LTE}-direct vs. {W}i{F}i-direct for machine-type communications over
  {LTE-A} systems,'' in \emph{Proc. 2015 IEEE 26th Annual International
  Symposium on Personal, Indoor, and Mobile Radio Communications (PIMRC)}, Aug
  2015, pp. 2298--2302.

\bibitem{7248779}
G.~Rigazzi, N.~K. Pratas, P.~Popovski, and R.~Fantacci, ``Aggregation and
  trunking of {M2M} traffic via {D2D} connections,'' in \emph{Proc. 2015 IEEE
  International Conference on Communications (ICC)}, June 2015, pp. 2973--2978.

\bibitem{6906487}
G.~Rigazzi, F.~Chiti, R.~Fantacci, and C.~Carlini, ``Multi-hop {D2D} networking
  and resource management scheme for {M2M} communications over {LTE-A}
  systems,'' in \emph{Proc. 2014 International Wireless Communications and
  Mobile Computing Conference (IWCMC)}, Aug 2014, pp. 973--978.

\bibitem{8241868}
S.~Wu, R.~Atat, N.~Mastronarde, and L.~Liu, ``Improving the coverage and
  spectral efficiency of millimeter-wave cellular networks using
  device-to-device relays,'' \emph{IEEE Transactions on Communications},
  vol.~66, no.~5, pp. 2251--2265, May 2018.

\bibitem{7051287}
N.~K. Pratas and P.~Popovski, ``Zero-outage cellular downlink with fixed-rate
  {D2D} underlay,'' \emph{IEEE Transactions on Wireless Communications},
  vol.~14, no.~7, pp. 3533--3543, July 2015.

\bibitem{J_ISJ_CZhu_09_2016}
C.~Zhu, H.~Wang, X.~Liu, L.~Shu, L.~T. Yang, and V.~C.~M. Leung, ``A novel
  sensory data processing framework to integrate sensor networks with mobile
  cloud,'' \emph{IEEE Systems Journal}, vol.~10, no.~3, pp. 1125--1136, Sept.
  2016.

\bibitem{7562084}
B.~Luo, W.~Liu, and A.~Al-Anbuky, ``Sustainme if you can: Sustainable
  transmission networking design for big data,'' in \emph{2016 IEEE Conference
  on Computer Communications Workshops (INFOCOM WKSHPS)}, April 2016, pp.
  265--270.

\bibitem{7417577}
L.~W. Cheng and S.~Y. Wang, ``Application-aware {SDN} routing for big data
  networking,'' in \emph{2015 IEEE Global Communications Conference
  (GLOBECOM)}, Dec 2015, pp. 1--6.

\bibitem{7389832}
L.~Cui, F.~R. Yu, and Q.~Yan, ``When big data meets software-defined
  networking: Sdn for big data and big data for sdn,'' \emph{IEEE Network},
  vol.~30, no.~1, pp. 58--65, January 2016.

\bibitem{6659463}
M.~Bakillah, S.~H.~L. Liang, A.~Mobasheri, and A.~Zipf, ``Towards an efficient
  routing web processing service through capturing real-time road conditions
  from big data,'' in \emph{Proc. Computer Science and Electronic Engineering
  Conference (CEEC), 2013 5th}, Sept 2013, pp. 152--155.

\bibitem{7491630}
M.~H. Zaki and T.~Sayed, ``Automated cyclist data collection under high density
  conditions,'' \emph{IET Intelligent Transport Systems}, vol.~10, no.~5, pp.
  361--369, 2016.

\bibitem{7569062}
C.~T. Cheng, N.~Ganganath, and K.~Y. Fok, ``Concurrent data collection trees
  for iot applications,'' \emph{IEEE Transactions on Industrial Informatics},
  vol.~13, no.~2, pp. 793--799, April 2017.

\bibitem{6330525}
S.~Chen, X.~Zhu, S.~Zhang, and J.~Wang, ``A framework for massive data
  transmission in a remote real-time health monitoring system,'' in \emph{Proc.
  18th International Conference on Automation and Computing (ICAC)}, Sept 2012,
  pp. 1--5.

\bibitem{chao2016distribution}
H.~Chao, Y.~Chen, J.~Wu, and H.~Zhang, ``Distribution reshaping for massive
  access control in cellular networks,'' in \emph{2016 84-th IEEE Vehicular
  Technology Conference (VTC-Fall)}, September 2016, pp. 1--5.

\bibitem{J_IA_SJab_04_2018}
S.~Jabbar, K.~R. Malik, M.~Ahmad, O.~Aldabbas, M.~Asif, S.~Khalid, K.~Han, and
  S.~H. Ahmed, ``A methodology of real-time data fusion for localized big data
  analytics,'' \emph{IEEE Access}, vol.~6, pp. 24\,510--24\,520, April 2018.

\bibitem{J_IA_AAkb_02_2018}
A.~Akbar, G.~Kousiouris, H.~Pervaiz, J.~Sancho, P.~Ta-Shma, F.~Carrez, and
  K.~Moessner, ``Real-time probabilistic data fusion for large-scale iot
  applications,'' \emph{IEEE Access}, vol.~6, pp. 10\,015--10\,027, February
  2018.

\bibitem{6415919}
T.~Kraska, ``Finding the needle in the big data systems haystack,'' \emph{IEEE
  Internet Computing}, vol.~17, no.~1, pp. 84--86, Jan 2013.

\bibitem{Dewitt}
\BIBentryALTinterwordspacing
D.~J. DeWitt and M.~Stonebraker. Mapreduce: A major step backwards. Accessed on
  July 10, 2017. [Online]. Available:
  \url{http://databasecolumn.vertica.com/database-innovation/mapreduce-a-major-step-backwards/}
\BIBentrySTDinterwordspacing

\bibitem{7293302}
D.~Wang and J.~Liu, ``Optimizing big data processing performance in the public
  cloud: opportunities and approaches,'' \emph{IEEE Network}, vol.~29, no.~5,
  pp. 31--35, September 2015.

\bibitem{Armbrust:2010:VCC:1721654.1721672}
\BIBentryALTinterwordspacing
M.~Armbrust, A.~Fox, R.~Griffith, A.~D. Joseph, R.~Katz, A.~Konwinski, G.~Lee,
  D.~Patterson, A.~Rabkin, I.~Stoica, and M.~Zaharia, ``A view of cloud
  computing,'' \emph{Commun. ACM}, vol.~53, no.~4, pp. 50--58, Apr. 2010.
  [Online]. Available: \url{http://doi.acm.org/10.1145/1721654.1721672}
\BIBentrySTDinterwordspacing

\bibitem{7037728}
S.~Tang, B.~S. Lee, and B.~He, ``Towards economic fairness for big data
  processing in pay-as-you-go cloud computing,'' in \emph{Proc. 2014 IEEE 6th
  International Conference on Cloud Computing Technology and Science
  (CloudCom)}, Dec 2014, pp. 638--643.

\bibitem{Blanas:2010:CJA:1807167.1807273}
\BIBentryALTinterwordspacing
S.~Blanas, J.~M. Patel, V.~Ercegovac, J.~Rao, E.~J. Shekita, and Y.~Tian, ``A
  comparison of join algorithms for log processing in mapreduce,'' in
  \emph{Proc. the 2010 ACM SIGMOD International Conference on Management of
  Data}, ser. SIGMOD '10.\hskip 1em plus 0.5em minus 0.4em\relax New York, NY,
  USA: ACM, 2010, pp. 975--986. [Online]. Available:
  \url{http://doi.acm.org/10.1145/1807167.1807273}
\BIBentrySTDinterwordspacing

\bibitem{7524388}
A.~Destounis, G.~S. Paschos, and I.~Koutsopoulos, ``Streaming big data meets
  backpressure in distributed network computation,'' in \emph{Proc. IEEE
  INFOCOM 2016 - The 35th IEEE International Conference on Computer
  Communications}, April 2016, pp. 1--9.

\bibitem{7412709}
C.~Yang and J.~Chen, ``A scalable data chunk similarity based compression
  approach for efficient big sensing data processing on cloud,'' \emph{IEEE
  Transactions on Knowledge and Data Engineering}, vol.~PP, no.~99, pp. 1--1,
  2016.

\bibitem{7066939}
J.~M. Lillo-Castellano, I.~Mora-Jim茅nez, R.~Santiago-Mozos,
  F.~Chavarria-Asso, A.~Cano-Gonzalez, A.~Garcia-Alberola, and J.~L.
  Rojo-Alvarez, ``Symmetrical compression distance for arrhythmia
  discrimination in cloud-based big-data services,'' \emph{IEEE Journal of
  Biomedical and Health Informatics}, vol.~19, no.~4, pp. 1253--1263, July
  2015.

\bibitem{7576619}
L.~A. Tawalbeh, R.~Mehmood, E.~Benkhlifa, and H.~Song, ``Mobile cloud computing
  model and big data analysis for healthcare applications,'' \emph{IEEE
  Access}, vol.~4, pp. 6171--6180, 2016.

\bibitem{7295892}
M.~Whaiduzzaman, A.~Gani, and A.~Naveed, ``Pefc: Performance enhancement
  framework for cloudlet in mobile cloud computing,'' in \emph{Proc. 2014 IEEE
  International Symposium on Robotics and Manufacturing Automation (ROMA)}, Dec
  2014, pp. 224--229.

\bibitem{7578484}
D.~Zhang, Y.~Shou, and J.~Xu, ``The modeling of big traffic data processing
  based on cloud computing,'' in \emph{Proc. 2016 12th World Congress on
  Intelligent Control and Automation (WCICA)}, June 2016, pp. 2394--2399.

\bibitem{7364117}
H.~Yeo and C.~H. Crawford, ``Big data: Cloud computing in genomics
  applications,'' in \emph{Big Data (Big Data), 2015 IEEE International
  Conference on}, Oct 2015, pp. 2904--2906.

\bibitem{7116525}
I.~Zinno, L.~Mossucca, S.~Elefante, C.~D. Luca, V.~Casola, O.~Terzo, F.~Casu,
  and R.~Lanari, ``Cloud computing for earth surface deformation analysis via
  spaceborne radar imaging: A case study,'' \emph{IEEE Transactions on Cloud
  Computing}, vol.~4, no.~1, pp. 104--118, Jan 2016.

\bibitem{7557446}
C.~Q. Wu and H.~Cao, ``Optimizing the performance of big data workflows in
  multi-cloud environments under budget constraint,'' in \emph{Proc. 2016 IEEE
  International Conference on Services Computing (SCC)}, June 2016, pp.
  138--145.

\bibitem{7049894}
L.~M. Pham, A.~Tchana, D.~Donsez, V.~Zurczak, P.~Y. Gibello, and N.~de~Palma,
  ``An adaptable framework to deploy complex applications onto multi-cloud
  platforms,'' in \emph{Proc. 2015 IEEE International Conference on Computing
  Communication Technologies - Research, Innovation, and Vision for the Future
  (RIVF)}, Jan 2015, pp. 169--174.

\bibitem{7383293}
H.~Li, M.~Dong, K.~Ota, and M.~Guo, ``Pricing and repurchasing for big data
  processing in multi-clouds,'' \emph{IEEE Transactions on Emerging Topics in
  Computing}, vol.~4, no.~2, pp. 266--277, April 2016.

\bibitem{7236926}
X.~Li, H.~Ma, W.~Yao, and X.~Gui, ``Data-driven and feedback-enhanced trust
  computing pattern for large-scale multi-cloud collaborative services,''
  \emph{IEEE Transactions on Services Computing}, 2015.

\bibitem{7353181}
S.~Wang, A.~Zhou, C.~H. Hsu, X.~Xiao, and F.~Yang, ``Provision of
  data-intensive services through energy- and qos-aware virtual machine
  placement in national cloud data centers,'' \emph{IEEE Transactions on
  Emerging Topics in Computing}, vol.~4, no.~2, pp. 290--300, April 2016.

\bibitem{7590316}
J.~Zhang, J.~Chen, J.~Luo, and A.~Song, ``Efficient location-aware data
  placement for data-intensive applications in geo-distributed scientific data
  centers,'' \emph{Tsinghua Science and Technology}, vol.~21, no.~5, pp.
  471--481, Oct 2016.

\bibitem{7225163}
Q.~Xia, Z.~Xu, W.~Liang, and A.~Y. Zomaya, ``Collaboration- and fairness-aware
  big data management in distributed clouds,'' \emph{IEEE Transactions on
  Parallel and Distributed Systems}, vol.~27, no.~7, pp. 1941--1953, July 2016.

\bibitem{7510724}
Y.~Zhao, Z.~Huang, W.~Liu, J.~Peng, and Q.~Zhang, ``A combinatorial double
  auction based resource allocation mechanism with multiple rounds for
  geo-distributed data centers,'' in \emph{Proc. 2016 IEEE International
  Conference on Communications (ICC)}, May 2016, pp. 1--6.

\bibitem{7283582}
D.~Kumar, J.~C. Bezdek, M.~Palaniswami, S.~Rajasegarar, C.~Leckie, and T.~C.
  Havens, ``A hybrid approach to clustering in big data,'' \emph{IEEE
  Transactions on Cybernetics}, vol.~46, no.~10, pp. 2372--2385, Oct 2016.

\bibitem{7546221}
C.~K.~K. Reddy, K.~E.~B. Chandrudu, P.~R. Anisha, and G.~V.~S. Raju, ``High
  performance computing cluster system and its future aspects in processing big
  data,'' in \emph{Proc. 2015 International Conference on Computational
  Intelligence and Communication Networks (CICN)}, Dec 2015, pp. 881--885.

\bibitem{J_TII_YZhao_n1_2018}
Y.~Zhao, L.~T. Yang, and R.~Zhang, ``A tensor-based multiple clustering
  approach with its applications in automation systems,'' \emph{IEEE
  Transactions on Industrial Informatics}, vol.~14, no.~1, pp. 283--291, 2018.

\bibitem{7456910}
R.~Shettar and B.~V. Purohit, ``A mapreduce framework to implement enhanced
  k-means algorithm,'' in \emph{Proc. 2015 International Conference on Applied
  and Theoretical Computing and Communication Technology (iCATccT)}, Oct 2015,
  pp. 361--363.

\bibitem{7363909}
J.~Karimov and M.~Ozbayoglu, ``High quality clustering of big data and solving
  empty-clustering problem with an evolutionary hybrid algorithm,'' in
  \emph{Proc. 2015 IEEE International Conference on Big Data (Big Data)}, Oct
  2015, pp. 1473--1478.

\bibitem{7207256}
J.~Karimov, M.~Ozbayoglu, and E.~Dogdu, ``k-means performance improvements with
  centroid calculation heuristics both for serial and parallel environments,''
  in \emph{Proc. 2015 IEEE International Congress on Big Data}, June 2015, pp.
  444--451.

\bibitem{Visuali}
\BIBentryALTinterwordspacing
{Data Visualization Tips}. Accessed on December 17, 2016. [Online]. Available:
  \url{http://www.sthda.com/english/wiki/hierarchical-clustering-essentials-unsupervised-machine-learning}
\BIBentrySTDinterwordspacing

\bibitem{centro}
\BIBentryALTinterwordspacing
I.~K. Kabul. (2016, May 26) Understanding data mining clustering methods.
  Accessed on December 17, 2016. [Online]. Available:
  \url{http://blogs.sas.com/content/subconsciousmusings/2016/05/26/data-mining-clustering/}
\BIBentrySTDinterwordspacing

\bibitem{7207373}
Y.~Zhao, C.~H. Chi, C.~Ding, R.~Wong, W.~Zhao, and C.~Wang, ``Hierarchical
  clustering using homogeneity as similarity measure for big data analytics,''
  in \emph{Proc. 2015 IEEE International Conference on Services Computing
  (SCC)}, June 2015, pp. 348--354.

\bibitem{7272099}
T.~S. Xu, H.~D. Chiang, G.~Y. Liu, and C.~W. Tan, ``Hierarchical k-means method
  for clustering large-scale advanced metering infrastructure data,''
  \emph{IEEE Transactions on Power Delivery}, vol.~PP, no.~99, pp. 1--1, 2015.

\bibitem{7555984}
A.~AlShami, W.~Guo, and G.~Pogrebna, ``Fuzzy partition technique for clustering
  big urban dataset,'' in \emph{Proc. 2016 SAI Computing Conference (SAI)},
  July 2016, pp. 212--216.

\bibitem{7473093}
J.~Fu, Y.~Liu, Z.~Zhang, and F.~Xiong, ``Big data clustering based on summary
  statistics,'' in \emph{Proc. 2015 First International Conference on
  Computational Intelligence Theory, Systems and Applications (CCITSA)}, Dec
  2015, pp. 87--91.

\bibitem{7359276}
S.~Dutta, S.~Ghatak, M.~Roy, S.~Ghosh, and A.~K. Das, ``A graph based
  clustering technique for tweet summarization,'' in \emph{Proc. 2015 4th
  International Conference on Reliability, Infocom Technologies and
  Optimization (ICRITO) (Trends and Future Directions)}, Sept 2015, pp. 1--6.

\bibitem{5319232}
J.~Cao and H.~Li, ``Energy-efficient structuralized clustering for sensor-based
  cyber physical systems,'' in \emph{Proc. 2009 Symposia and Workshops on
  Ubiquitous, Autonomic and Trusted Computing}, July 2009, pp. 234--239.

\bibitem{6863675}
Z.~Huang, C.~Wang, A.~Nayak, and I.~Stojmenovic, ``Small cluster in cyber
  physical systems: Network topology, interdependence and cascading failures,''
  \emph{IEEE Transactions on Parallel and Distributed Systems}, vol.~26, no.~8,
  pp. 2340--2351, Aug 2015.

\bibitem{Bali2016476}
\BIBentryALTinterwordspacing
R.~S. Bali and N.~Kumar, ``Secure clustering for efficient data dissemination
  in vehicular cyber-physical systems,'' \emph{Future Generation Computer
  Systems}, vol.~56, pp. 476 -- 492, 2016. [Online]. Available:
  \url{http://www.sciencedirect.com/science/article/pii/S0167739X15002836}
\BIBentrySTDinterwordspacing

\bibitem{huo2017coalition}
Y.~Huo, W.~Dong, J.~Qian, and T.~Jing, ``Coalition game-based secure and
  effective clustering communication in vehicular cyber-physical system
  (vcps),'' \emph{Sensors}, vol.~17, no.~3, p. 475, 2017.

\bibitem{SPEZZANO20151016}
\BIBentryALTinterwordspacing
G.~Spezzano and A.~Vinci, ``Pattern detection in cyber-physical systems,''
  \emph{Procedia Computer Science}, vol.~52, pp. 1016 -- 1021, 2015. [Online].
  Available:
  \url{http://www.sciencedirect.com/science/article/pii/S1877050915008960}
\BIBentrySTDinterwordspacing

\bibitem{6967763}
Z.~Tan, A.~Jamdagni, X.~He, P.~Nanda, R.~P. Liu, and J.~Hu, ``Detection of
  denial-of-service attacks based on computer vision techniques,'' \emph{IEEE
  Transactions on Computers}, vol.~64, no.~9, pp. 2519--2533, Sept 2015.

\bibitem{7558012}
J.~S. Lee and I.~Y. Ko, ``Service recommendation for user groups in internet of
  things environments using member organization-based group similarity
  measures,'' in \emph{Proc. 2016 IEEE International Conference on Web Services
  (ICWS)}, June 2016, pp. 276--283.

\bibitem{6960906}
S.~Chen, X.~Yang, and Y.~Tian, ``Discriminative hierarchical k-means tree for
  large-scale image classification,'' \emph{IEEE Transactions on Neural
  Networks and Learning Systems}, vol.~26, no.~9, pp. 2200--2205, Sept 2015.

\bibitem{7000907}
H.~R. Arkian, R.~E. Atani, and S.~Kamali, ``Fcvca: A fuzzy clustering-based
  vehicular cloud architecture,'' in \emph{Proc. 2014 7th International
  Workshop on Communication Technologies for Vehicles (Nets4Cars-Fall)}, Oct
  2014, pp. 24--28.

\bibitem{7292782}
R.~Poorvadevi and S.~Rajalakshmi, ``A cluster based signature evaluation
  mechanism for protecting the user data in cloud environment through fuzzy
  ordering approach,'' in \emph{Proc. 2015 International Conference on
  Computing and Communications Technologies (ICCCT)}, Feb 2015, pp. 392--397.

\bibitem{7743854}
J.~C.~C. Tseng, J.~Y. Gu, P.~F. Wang, C.~Y. Chen, C.~F. Li, and V.~S. Tseng,
  ``A scalable complex event analytical system with incremental episode mining
  over data streams,'' in \emph{Proc. 2016 IEEE Congress on Evolutionary
  Computation (CEC)}, July 2016, pp. 648--655.

\bibitem{7511386}
L.~Yang, K.~Wang, C.~Xu, C.~Zhu, and Y.~Sun, ``An incremental learning
  classification algorithm based on forgetting factor for ehealth networks,''
  in \emph{Proc. 2016 IEEE International Conference on Communications (ICC)},
  May 2016, pp. 1--6.

\bibitem{7578325}
D.~Fu, T.~Zhou, ZeyuZheng, Y.~Fu, and Y.~Tong, ``A modified-distance-based
  minimum spanning tree method for analyzing hierarchical structure of power
  generation system,'' in \emph{2016 12th World Congress on Intelligent Control
  and Automation (WCICA)}, June 2016, pp. 422--425.

\bibitem{6740862}
S.~Li, G.~Oikonomou, T.~Tryfonas, T.~M. Chen, and L.~D. Xu, ``A distributed
  consensus algorithm for decision making in service-oriented internet of
  things,'' \emph{IEEE Transactions on Industrial Informatics}, vol.~10, no.~2,
  pp. 1461--1468, May 2014.

\bibitem{7551774}
E.~Al-sharoa and S.~Aviyente, ``Evolutionary spectral graph clustering through
  subspace distance measure,'' in \emph{Proc. 2016 IEEE Statistical Signal
  Processing Workshop (SSP)}, June 2016, pp. 1--5.

\bibitem{7279070}
Y.~S. Kang, I.~H. Park, J.~Rhee, and Y.~H. Lee, ``Mongodb-based repository
  design for iot-generated rfid/sensor big data,'' \emph{IEEE Sensors Journal},
  vol.~16, no.~2, pp. 485--497, Jan 2016.

\bibitem{Cattell:2011:SSN:1978915.1978919}
\BIBentryALTinterwordspacing
R.~Cattell, ``Scalable sql and nosql data stores,'' \emph{SIGMOD Rec.},
  vol.~39, no.~4, pp. 12--27, May 2011. [Online]. Available:
  \url{http://doi.acm.org/10.1145/1978915.1978919}
\BIBentrySTDinterwordspacing

\bibitem{7452296}
J.~M. Patel, ``Operational nosql systems: What's new and what's next?''
  \emph{Computer}, vol.~49, no.~4, pp. 23--30, Apr 2016.

\bibitem{strauch2011nosql}
C.~Strauch, U.-L.~S. Sites, and W.~Kriha, ``{NoSQL databases},'' \emph{URL:
  http://www. christof-strauch. de/nosqldbs. pdf (写邪褌邪
  芯斜褉邪褖械薪懈褟 07.11. 2012)}, 2011.

\bibitem{7584920}
A.~Mohan, M.~Ebrahimi, S.~Lu, and A.~Kotov, ``A nosql data model for scalable
  big data workflow execution,'' in \emph{Proc. 2016 IEEE International
  Congress on Big Data (BigData Congress)}, June 2016, pp. 52--59.

\bibitem{Bonomi}
\BIBentryALTinterwordspacing
F.~Bonomi, R.~Milito, J.~Zhu, and S.~Addepalli, ``Fog computing and its role in
  the internet of things,'' in \emph{Proceedings of the First Edition of the
  MCC Workshop on Mobile Cloud Computing}, ser. MCC '12.\hskip 1em plus 0.5em
  minus 0.4em\relax New York, NY, USA: ACM, 2012, pp. 13--16. [Online].
  Available: \url{http://doi.acm.org/10.1145/2342509.2342513}
\BIBentrySTDinterwordspacing

\bibitem{FedCSIS2014503}
\BIBentryALTinterwordspacing
I.~Stojmenovic and S.~Wen, ``The fog computing paradigm: Scenarios and security
  issues,'' in \emph{Proc. the 2014 Federated Conference on Computer Science
  and Information Systems}, ser. Annals of Computer Science and Information
  Systems, M.~P. M.~Ganzha, L.~Maciaszek, Ed., vol.~2.\hskip 1em plus 0.5em
  minus 0.4em\relax IEEE, 2014, pp. 1--8. [Online]. Available:
  \url{http://dx.doi.org/10.15439/2014F503}
\BIBentrySTDinterwordspacing

\bibitem{Roy}
.~S. Roy, R.~Bose, and D.~Sarddar, ``A fog-based dss model for driving rule
  violation monitoring framework on the internet of things,'' \emph{Journal of
  Advanced Science and Technology}, vol.~82, p. 23鈥?2, 2015.

\bibitem{7545004}
N.~Chen, Y.~Chen, Y.~You, H.~Ling, P.~Liang, and R.~Zimmermann, ``Dynamic urban
  surveillance video stream processing using fog computing,'' in \emph{Proc.
  2016 IEEE Second International Conference on Multimedia Big Data (BigMM)},
  April 2016, pp. 105--112.

\bibitem{7255196}
Y.~Cao, S.~Chen, P.~Hou, and D.~Brown, ``Fast: A fog computing assisted
  distributed analytics system to monitor fall for stroke mitigation,'' in
  \emph{Proc. 2015 IEEE International Conference on Networking, Architecture
  and Storage (NAS)}, Aug 2015, pp. 2--11.

\bibitem{7421170}
R.~Craciunescu, A.~Mihovska, M.~Mihaylov, S.~Kyriazakos, R.~Prasad, and
  S.~Halunga, ``Implementation of fog computing for reliable e-health
  applications,'' in \emph{Proc. 2015 49th Asilomar Conference on Signals,
  Systems and Computers}, Nov 2015, pp. 459--463.

\bibitem{BRZOZAWOCH20152387}
\BIBentryALTinterwordspacing
R.~Brzoza-Woch, M.~Konieczny, B.~Kwolek, P.~Nawrocki, T.~Szyd艂o, and
  K.~Zieli艅ski, ``Holistic approach to urgent computing for flood decision
  support,'' \emph{Procedia Computer Science}, vol.~51, pp. 2387 -- 2396, 2015.
  [Online]. Available:
  \url{http://www.sciencedirect.com/science/article/pii/S1877050915012223}
\BIBentrySTDinterwordspacing

\bibitem{6910482}
J.~K. Zao, T.~T. Gan, C.~K. You, S.~J.~R. M茅ndez, C.~E. Chung, Y.~T. Wang,
  T.~Mullen, and T.~P. Jung, ``Augmented brain computer interaction based on
  fog computing and linked data,'' in \emph{Proc. 2014 International Conference
  on Intelligent Environments}, June 2014, pp. 374--377.

\bibitem{7363093}
T.~N. Gia, M.~Jiang, A.~M. Rahmani, T.~Westerlund, P.~Liljeberg, and
  H.~Tenhunen, ``Fog computing in healthcare internet of things: A case study
  on ecg feature extraction,'' in \emph{2015 IEEE International Conference on
  Computer and Information Technology; Ubiquitous Computing and Communications;
  Dependable, Autonomic and Secure Computing; Pervasive Intelligence and
  Computing}, Oct 2015, pp. 356--363.

\bibitem{Hong:2013:MFP:2491266.2491270}
\BIBentryALTinterwordspacing
K.~Hong, D.~Lillethun, U.~Ramachandran, B.~Ottenw\"{a}lder, and B.~Koldehofe,
  ``Mobile fog: A programming model for large-scale applications on the
  internet of things,'' in \emph{Proceedings of the Second ACM SIGCOMM Workshop
  on Mobile Cloud Computing}, ser. MCC '13.\hskip 1em plus 0.5em minus
  0.4em\relax New York, NY, USA: ACM, 2013, pp. 15--20. [Online]. Available:
  \url{http://doi.acm.org/10.1145/2491266.2491270}
\BIBentrySTDinterwordspacing

\bibitem{Ottenwalder:2013:MOM:2488222.2488265}
\BIBentryALTinterwordspacing
B.~Ottenw\"{a}lder, B.~Koldehofe, K.~Rothermel, and U.~Ramachandran, ``Migcep:
  Operator migration for mobility driven distributed complex event
  processing,'' in \emph{Proceedings of the 7th ACM International Conference on
  Distributed Event-based Systems}, ser. DEBS '13.\hskip 1em plus 0.5em minus
  0.4em\relax New York, NY, USA: ACM, 2013, pp. 183--194. [Online]. Available:
  \url{http://doi.acm.org/10.1145/2488222.2488265}
\BIBentrySTDinterwordspacing

\bibitem{7372286}
S.~Yi, Z.~Hao, Z.~Qin, and Q.~Li, ``Fog computing: Platform and applications,''
  in \emph{2015 Third IEEE Workshop on Hot Topics in Web Systems and
  Technologies (HotWeb)}, Nov 2015, pp. 73--78.

\bibitem{5741143}
R.~Lu, X.~Li, X.~Liang, X.~Shen, and X.~Lin, ``Grs: The green, reliability, and
  security of emerging machine to machine communications,'' \emph{IEEE
  Communications Magazine}, vol.~49, no.~4, pp. 28--35, April 2011.

\bibitem{5471501}
L.~Zhang, W.~Jia, S.~Wen, and D.~Yao, ``A man-in-the-middle attack on 3g-wlan
  interworking,'' in \emph{2010 International Conference on Communications and
  Mobile Computing}, vol.~1, April 2010, pp. 121--125.

\bibitem{J_CAIS_HJWat_2014}
H.~J. Watson, ``Tutorial: Big data analytics: Concepts, technologies, and
  applications,'' \emph{Communication of the Association for Information
  Systems}, vol. 34, Article 65, pp. 124–--168, 2014.

\bibitem{5621980}
F.~Qu, F.~Y. Wang, and L.~Yang, ``Intelligent transportation spaces: vehicles,
  traffic, communications, and beyond,'' \emph{IEEE Communications Magazine},
  vol.~48, no.~11, pp. 136--142, November 2010.

\bibitem{6755512}
L.~Xu, J.~Li, Y.~Shu, and J.~Peng, ``Sar image denoising via clustering-based
  principal component analysis,'' \emph{IEEE Transactions on Geoscience and
  Remote Sensing}, vol.~52, no.~11, pp. 6858--6869, Nov 2014.

\bibitem{5170925}
T.~C. Chen, S.~Sanga, T.~Y. Chou, V.~Cristini, and M.~E. Edgerton, ``Neural
  network with k-means clustering via pca for gene expression profile
  analysis,'' in \emph{Proc. 2009 WRI World Congress on Computer Science and
  Information Engineering}, vol.~3, March 2009, pp. 670--673.

\bibitem{Fayyad:1996:DMK:257938.257942}
\BIBentryALTinterwordspacing
U.~M. Fayyad, G.~Piatetsky-Shapiro, and P.~Smyth, ``Advances in knowledge
  discovery and data mining,'' U.~M. Fayyad, G.~Piatetsky-Shapiro, P.~Smyth,
  and R.~Uthurusamy, Eds.\hskip 1em plus 0.5em minus 0.4em\relax Menlo Park,
  CA, USA: American Association for Artificial Intelligence, 1996, ch. From
  Data Mining to Knowledge Discovery: An Overview, pp. 1--34. [Online].
  Available: \url{http://dl.acm.org/citation.cfm?id=257938.257942}
\BIBentrySTDinterwordspacing

\bibitem{7515055}
M.~Behl and R.~Mangharam, ``Interactive analytics for smart cities
  infrastructures,'' in \emph{Proc. 2016 1st International Workshop on Science
  of Smart City Operations and Platforms Engineering (SCOPE) in partnership
  with Global City Teams Challenge (GCTC) (SCOPE - GCTC)}, April 2016, pp.
  1--6.

\bibitem{60}
V.~Kardeby, ``Automatic sensor clustering: connectivity for the internet of
  things,'' in \emph{Licentiate thesis, Mid Sweden University, Department of
  Information Technology and Media}, 2011.

\bibitem{5560652}
P.~Rashidi, D.~J. Cook, L.~B. Holder, and M.~Schmitter-Edgecombe, ``Discovering
  activities to recognize and track in a smart environment,'' \emph{IEEE
  Transactions on Knowledge and Data Engineering}, vol.~23, no.~4, pp.
  527--539, April 2011.

\bibitem{J_TCOMPU_DZhang_n2_2015}
D.~Zhang, D.~Zhang, H.~Xiong, L.~T. Yang, and V.~Gauthier, ``Nextcell:
  predicting location using social interplay from cell phone traces,''
  \emph{IEEE Transactions on Computers}, vol.~64, no.~2, pp. 452--463, 2015.

\bibitem{7592729}
A.~Sotsenko, M.~Jansen, M.~Milrad, and J.~Rana, ``Using a rich context model
  for real-time big data analytics in twitter,'' in \emph{Proc. 2016 IEEE 4th
  International Conference on Future Internet of Things and Cloud Workshops
  (FiCloudW)}, Aug 2016, pp. 228--233.

\bibitem{7568906}
I.~Toure and A.~Gangopadhyay, ``Real time big data analytics for predicting
  terrorist incidents,'' in \emph{Proc. 2016 IEEE Symposium on Technologies for
  Homeland Security (HST)}, May 2016, pp. 1--6.

\bibitem{7529531}
V.-D. Ta, C.-M. Liu, and G.~W. Nkabinde, ``Big data stream computing in
  healthcare real-time analytics,'' in \emph{Proc. 2016 IEEE International
  Conference on Cloud Computing and Big Data Analysis (ICCCBDA)}, July 2016,
  pp. 37--42.

\bibitem{7292404}
A.~Daniel, A.~Paul, and A.~Ahmad, ``Near real-time big data analysis on
  vehicular networks,'' in \emph{Proc. 2015 International Conference on
  Soft-Computing and Networks Security (ICSNS)}, Feb 2015, pp. 1--7.

\bibitem{C_ICCC_KWang_07_2016}
K.~Wang, J.~Mi, C.~Xu, L.~Shu, and D.-J. Deng, ``Real-time big data analytics
  for multimedia transmission and storage,'' in \emph{Proc. 2016 IEEE/CIC
  International Conference on Communications in China (ICCC)}, July 2016, pp.
  1--6.

\bibitem{C_GLOBECOM_BZhou_12_2017}
B.~Zhou, J.~Li, S.~Guo, J.~Wu, Y.~Hu, and L.~Zhu, ``Online internet traffic
  measurement and monitoring using spark streaming,'' in \emph{Proc. IEEE
  Global Communications Conference (GLOBECOM)}, Dec. 2017, pp. 1--6.

\bibitem{C_ICBD_HCao_12_2017}
H.~Cao, M.~Wachowicz, and S.~Cha, ``Developing an edge computing platform for
  real-time descriptive analytics,'' in \emph{Proc. 2017 IEEE International
  Conference on Big Data (Big Data)}, Dec. 2017, pp. 4546--4554.

\bibitem{C_ICBD_STri_12_2017}
S.~Trinks and C.~Felden, ``Real time analytics — state of the art: Potentials
  and limitations in the smart factory,'' in \emph{Proc. 2017 IEEE
  International Conference on Big Data (Big Data)}, Dec. 2017, pp. 4843--4845.

\bibitem{C_ICBDA_ARAli_03_2018}
A.~R. Ali, ``Real-time big data warehousing and analysis framework,'' in
  \emph{Proc. 2018 IEEE 3rd International Conference on Big Data Analysis
  (ICBDA)}, March 2018, pp. 43--49.

\bibitem{7360233}
S.~Liu, J.~Yin, X.~Wang, W.~Cui, K.~Cao, and J.~Pei, ``Online visual analytics
  of text streams,'' \emph{IEEE Transactions on Visualization and Computer
  Graphics}, vol.~22, no.~11, pp. 2451--2466, Nov 2016.

\bibitem{7338157}
P.~Chopade, J.~Zhan, K.~Roy, and K.~Flurchick, ``Real-time large-scale big data
  networks analytics and visualization architecture,'' in \emph{Proc. Emerging
  Technologies for a Smarter World (CEWIT), 2015 12th International Conference
  Expo on}, Oct 2015, pp. 1--6.

\bibitem{7520541}
L.~Mo, F.~Li, Y.~Zhu, and A.~Huang, ``Human physical activity recognition based
  on computer vision with deep learning model,'' in \emph{Proc. 2016 IEEE
  International Instrumentation and Measurement Technology Conference
  Proceedings}, May 2016, pp. 1--6.

\bibitem{7399283}
S.~Singh and Y.~Liu, ``A cloud service architecture for analyzing big
  monitoring data,'' \emph{Tsinghua Science and Technology}, vol.~21, no.~1,
  pp. 55--70, Feb 2016.

\bibitem{DBLP}
\BIBentryALTinterwordspacing
{Microsoft Download Center}, ``Project daytona: Iterative mapreduce on windows
  azure,'' 2016. [Online]. Available:
  \url{https://www.microsoft.com/en-us/download/details.aspx?id=52431}
\BIBentrySTDinterwordspacing

\bibitem{7582986}
Y.~Yetis, R.~G. Sara, B.~A. Erol, H.~Kaplan, A.~Akuzum, and M.~Jamshidi,
  ``Application of big data analytics via cloud computing,'' in \emph{Proc.
  2016 World Automation Congress (WAC)}, July 2016, pp. 1--5.

\bibitem{7431422}
F.~J. Clemente-Castell贸, B.~Nicolae, K.~Katrinis, M.~M. Rafique, R.~Mayo,
  J.~C. Fern谩ndez, and D.~Loreti, ``Enabling big data analytics in the hybrid
  cloud using iterative mapreduce,'' in \emph{Proc. 2015 IEEE/ACM 8th
  International Conference on Utility and Cloud Computing (UCC)}, Dec 2015, pp.
  290--299.

\bibitem{6877244}
M.~V. Neves, C.~A. F.~D. Rose, K.~Katrinis, and H.~Franke, ``Pythia: Faster big
  data in motion through predictive software-defined network optimization at
  runtime,'' in \emph{Proc. Parallel and Distributed Processing Symposium, 2014
  IEEE 28th International}, May 2014, pp. 82--90.

\bibitem{Islam:2012:EPM:2304777.2304880}
\BIBentryALTinterwordspacing
S.~Islam, J.~Keung, K.~Lee, and A.~Liu, ``Empirical prediction models for
  adaptive resource provisioning in the cloud,'' \emph{Future Gener. Comput.
  Syst.}, vol.~28, no.~1, pp. 155--162, Jan. 2012. [Online]. Available:
  \url{http://dx.doi.org/10.1016/j.future.2011.05.027}
\BIBentrySTDinterwordspacing

\bibitem{7384281}
R.~Buyya, K.~Ramamohanarao, C.~Leckie, R.~N. Calheiros, A.~V. Dastjerdi, and
  S.~Versteeg, ``Big data analytics-enhanced cloud computing: Challenges,
  architectural elements, and future directions,'' in \emph{Proc. 2015 IEEE
  21st International Conference on Parallel and Distributed Systems (ICPADS)},
  Dec 2015, pp. 75--84.

\bibitem{C_IGARSS_YZhong_07_2013}
Y.~Zhong, J.~Fang, and X.~Zhao, ``{VegaIndexer}: A distributed composite index
  scheme for big spatio-temporal sensor data on cloud,'' in \emph{Proc. 2013
  IEEE International Geoscience and Remote Sensing Symposium (IGARSS)}, July
  2013, pp. 1713--1716.

\bibitem{C_ACBD_DZheng_11_2014}
D.~Zheng, K.~Ben, and H.~Yuan, ``Research of big data space-time analytics for
  clouding based contexts-aware {IOV} applications,'' in \emph{Proc. 2014
  Second International Conference on Advanced Cloud and Big Data}, Nov. 2014,
  pp. 150--156.

\bibitem{C_DASC_MSin_11_2017}
M.~Sinda and Q.~Liao, ``Spatial-temporal anomaly detection using security
  visual analytics via entropy graph and eigen matrix,'' in \emph{Proc. 2017
  IEEE 15th Intl Conf on Dependable, Autonomic and Secure Computing, 15th Intl
  Conf on Pervasive Intelligence and Computing, 3rd Intl Conf on Big Data
  Intelligence and Computing and Cyber Science and Technology
  Congress(DASC/PiCom/DataCom/CyberSciTech)}, Nov. 2017, pp. 511--518.

\bibitem{J_IEICE_TCOM_GDing_08_2014}
G.~Ding, Z.~Tan, J.~Wu, and J.~Zhang, ``Efficient indoor fingerprinting
  localization technique using regional propagation model,'' \emph{IEICE
  Transactions on Communications}, vol. E97-B, no.~8, pp. 1728--1741, Aug.
  2014.

\bibitem{J_IEICETCOM_GDing_03_2015}
G.~Ding, Z.~Tan, J.~Wu, J.~Zeng, and L.~Zhang, ``Indoor fingerprinting
  localization and tracking system using particle swarm optimization and kalman
  filter,'' \emph{IEICE Transactions on Communications}, vol. E98-B, no.~3, pp.
  502--514, Mar. 2015.

\bibitem{6779407}
K.~Singh and R.~Kaur, ``Hadoop: Addressing challenges of big data,'' in
  \emph{Proc. 2014 IEEE International Advance Computing Conference (IACC)}, Feb
  2014, pp. 686--689.

\bibitem{Hashem:2015:RBD:2946158.2946407}
\BIBentryALTinterwordspacing
I.~A.~T. Hashem, I.~Yaqoob, N.~B. Anuar, S.~Mokhtar, A.~Gani, and
  S.~Ullah~Khan, ``The rise of {Big Data} on cloud computing,'' \emph{Inf.
  Syst.}, vol.~47, no.~C, pp. 98--115, Jan. 2015. [Online]. Available:
  \url{http://dx.doi.org/10.1016/j.is.2014.07.006}
\BIBentrySTDinterwordspacing

\bibitem{IETCPS_RAtat_04_2017}
R.~Atat, L.~Liu, H.~Chen, J.~Wu, H.~Li, and Y.~Yi, ``Enabling cyber-physical
  communication in {5G} cellular networks: challenges, spatial spectrum
  sensing, and cyber-security,'' \emph{IET Cyber-Physical Systems: Theory
  Applications}, vol.~2, no.~1, pp. 49--54, Apr. 2017.

\bibitem{7502279}
K.~Gai, M.~Qiu, and H.~Zhao, ``Security-aware efficient mass distributed
  storage approach for cloud systems in big data,'' in \emph{Proc. 2016 IEEE
  2nd International Conference on Big Data Security on Cloud (BigDataSecurity),
  IEEE International Conference on High Performance and Smart Computing (HPSC),
  and IEEE International Conference on Intelligent Data and Security (IDS)},
  April 2016, pp. 140--145.

\bibitem{7502311}
S.~Kang, B.~Veeravalli, and K.~M.~M. Aung, ``A security-aware data placement
  mechanism for big data cloud storage systems,'' in \emph{Proc. 2016 IEEE 2nd
  International Conference on Big Data Security on Cloud (BigDataSecurity),
  IEEE International Conference on High Performance and Smart Computing (HPSC),
  and IEEE International Conference on Intelligent Data and Security (IDS)},
  April 2016, pp. 327--332.

\bibitem{7724531}
K.~Sekar and M.~Padmavathamma, ``Comparative study of encryption algorithm over
  big data in cloud systems,'' in \emph{Proc. 2016 3rd International Conference
  on Computing for Sustainable Global Development (INDIACom)}, March 2016, pp.
  1571--1574.

\bibitem{J_IoTJ_CJDo_04_2017}
C.~J. D’Orazio, K.-K.~R. Choo, and L.~T. Yang, ``Data exfiltration from
  internet of things devices: {iOS} devices as case studies,'' \emph{IEEE
  Internet of Things Journal}, vol.~4, no.~2, pp. 524--535, April 2017.

\bibitem{7636866}
J.~Ni, X.~Lin, K.~Zhang, Y.~Yu, and X.~S. Shen, ``Secure outsourced data
  transfer with integrity verification in cloud storage,'' in \emph{2016
  IEEE/CIC International Conference on Communications in China (ICCC)}, July
  2016, pp. 1--6.

\bibitem{opendedup}
\BIBentryALTinterwordspacing
opendedup. (2016) Accessed on July 1, 2017. [Online]. Available:
  \url{http://opendedup.org/}
\BIBentrySTDinterwordspacing

\bibitem{Meyer:2012:SPD:2078861.2078864}
D.~T. Meyer and W.~J. Bolosky, ``A study of practical deduplication,'' in
  \emph{the 9th USENIX conference on File and stroage technologies}, Feb. 2011,
  pp. 14:1--14:20.

\bibitem{7511769}
Z.~Yan, W.~Ding, X.~Yu, H.~Zhu, and R.~H. Deng, ``Deduplication on encrypted
  big data in cloud,'' \emph{IEEE Transactions on Big Data}, vol.~2, no.~2, pp.
  138--150, June 2016.

\bibitem{7543859}
Y.~Gahi, M.~Guennoun, and H.~T. Mouftah, ``Big data analytics: Security and
  privacy challenges,'' in \emph{2016 IEEE Symposium on Computers and
  Communication (ISCC)}, June 2016, pp. 952--957.

\bibitem{7306738}
S.~Rao, S.~N. Suma, and M.~Sunitha, ``Security solutions for big data analytics
  in healthcare,'' in \emph{Advances in Computing and Communication Engineering
  (ICACCE), 2015 Second International Conference on}, May 2015, pp. 510--514.

\bibitem{6725337}
T.~Mahmood and U.~Afzal, ``Security analytics: Big data analytics for
  cybersecurity: A review of trends, techniques and tools,'' in
  \emph{Information Assurance (NCIA), 2013 2nd National Conference on}, Dec
  2013, pp. 129--134.

\bibitem{7429688}
Y.~He, F.~R. Yu, N.~Zhao, H.~Yin, H.~Yao, and R.~C. Qiu, ``Big data analytics
  in mobile cellular networks,'' \emph{IEEE Access}, vol.~4, pp. 1985--1996,
  2016.

\bibitem{J_TII_JFeng_EA_2018}
Y.~Zhao, L.~T. Yang, and J.~Sun, ``A secure high-order {CFS} algorithm on
  clouds for industrial internet of things,'' \emph{IEEE Transactions on
  Industrial Informatics}, vol.~14, no.~8, pp. 3766--3774, 2018.

\bibitem{he}
D.~He, S.~Chan, Y.~Zhang, C.~Wu, and B.~Wang, ``How effective are the
  prevailing attack-defense models for cybersecurity anyway?'' \emph{IEEE
  Intelligent Systems}, vol.~29, no.~5, pp. 14--21, Sept 2014.

\bibitem{ristic}
B.~Ristic, ``Detecting anomalies from a multitarget tracking output,''
  \emph{IEEE Transactions on Aerospace and Electronic Systems}, vol.~50, no.~1,
  pp. 798--803, January 2014.

\bibitem{rocha}
E.~Rocha, P.~Salvador, and A.~Nogueira, ``Detection of illicit network
  activities based on multivariate gaussian fitting of multi-scale traffic
  characteristics,'' in \emph{2011 IEEE International Conference on
  Communications (ICC)}, June 2011, pp. 1--6.

\bibitem{jia}
L.~Jia, M.~Li, P.~Zhang, Y.~Wu, and H.~Zhu, ``Sar image change detection based
  on multiple kernel k-means clustering with local-neighborhood information,''
  \emph{IEEE Geoscience and Remote Sensing Letters}, vol.~13, no.~6, pp.
  856--860, June 2016.

\bibitem{murphree}
J.~Murphree, ``Machine learning anomaly detection in large systems,'' in
  \emph{2016 IEEE AUTOTESTCON}, Sept 2016, pp. 1--9.

\bibitem{yuan}
Y.~Yuan and K.~Jia, ``A distributed anomaly detection method of operation
  energy consumption using smart meter data,'' in \emph{2015 International
  Conference on Intelligent Information Hiding and Multimedia Signal Processing
  (IIH-MSP)}, Sept 2015, pp. 310--313.

\bibitem{6650119}
T.~Plantard, W.~Susilo, and Z.~Zhang, ``Fully homomorphic encryption using
  hidden ideal lattice,'' \emph{IEEE Transactions on Information Forensics and
  Security}, vol.~8, no.~12, pp. 2127--2137, Dec 2013.

\bibitem{7478543}
H.~Kumarage, I.~Khalil, A.~Alabdulatif, Z.~Tari, and X.~Yi, ``Secure data
  analytics for cloud-integrated internet of things applications,'' \emph{IEEE
  Cloud Computing}, vol.~3, no.~2, pp. 46--56, Mar 2016.

\bibitem{8005854}
Y.~Li, Z.~L. Jiang, X.~Wang, and S.~M. Yiu, ``Privacy-preserving id3 data
  mining over encrypted data in outsourced environments with multiple keys,''
  in \emph{Proc. 2017 IEEE International Conference on Computational Science
  and Engineering (CSE) and IEEE International Conference on Embedded and
  Ubiquitous Computing (EUC)}, vol.~1, July 2017, pp. 548--555.

\bibitem{8010376}
S.~Qiu, B.~Wang, M.~Li, J.~Liu, and Y.~Shi, ``Toward practical
  privacy-preserving frequent itemset mining on encrypted cloud data,''
  \emph{IEEE Transactions on Cloud Computing}, 2017.

\bibitem{6863131}
R.~Lu, H.~Zhu, X.~Liu, J.~K. Liu, and J.~Shao, ``Toward efficient and
  privacy-preserving computing in big data era,'' \emph{IEEE Network}, vol.~28,
  no.~4, pp. 46--50, July 2014.

\bibitem{6565224}
A.~Chakravorty, T.~Wlodarczyk, and C.~Rong, ``Privacy preserving data analytics
  for smart homes,'' in \emph{Proc. 2013 IEEE Security and Privacy Workshops},
  May 2013, pp. 23--27.

\bibitem{1623889}
T.~M. Truta and B.~Vinay, ``Privacy protection: p-sensitive k-anonymity
  property,'' in \emph{Proc. 22nd International Conference on Data Engineering
  Workshops (ICDEW'06)}, 2006, pp. 94--94.

\bibitem{7431421}
Z.~Gheid and Y.~Challal, ``An efficient and privacy-preserving similarity
  evaluation for big data analytics,'' in \emph{Proc. 2015 IEEE/ACM 8th
  International Conference on Utility and Cloud Computing (UCC)}, Dec 2015, pp.
  281--289.

\bibitem{8013260}
Y.~T. Lee, W.~H. Hsiao, Y.~S. Lin, and S.~C.~T. Chou, ``Privacy-preserving data
  analytics in cloud-based smart home with community hierarchy,'' \emph{IEEE
  Transactions on Consumer Electronics}, vol.~63, no.~2, pp. 200--207, May
  2017.

\bibitem{7414062}
S.~Chen, M.~Ma, and Z.~Luo, ``An authentication framework for multi-domain
  machine-to-machine communication in cyber-physical systems,'' in \emph{Proc.
  2015 IEEE Globecom Workshops (GC Wkshps)}, Dec 2015, pp. 1--6.

\bibitem{7737997}
N.~Tamani and Y.~Ghamri-Doudane, ``Towards a user privacy preservation system
  for iot environments: a habit-based approach,'' in \emph{Proc. 2016 IEEE
  International Conference on Fuzzy Systems (FUZZ-IEEE)}, July 2016, pp.
  2425--2432.

\bibitem{7347956}
V.~Sivaraman, H.~H. Gharakheili, A.~Vishwanath, R.~Boreli, and O.~Mehani,
  ``Network-level security and privacy control for smart-home iot devices,'' in
  \emph{Proc. 2015 IEEE 11th International Conference on Wireless and Mobile
  Computing, Networking and Communications (WiMob)}, Oct 2015, pp. 163--167.

\bibitem{7479069}
C.~Liu, S.~Ghosal, Z.~Jiang, and S.~Sarkar, ``An unsupervised spatiotemporal
  graphical modeling approach to anomaly detection in distributed cps,'' in
  \emph{Proc. 2016 ACM/IEEE 7th International Conference on Cyber-Physical
  Systems (ICCPS)}, April 2016, pp. 1--10.

\bibitem{6882174}
P.~Y. Chen, S.~Yang, and J.~A. McCann, ``Distributed real-time anomaly
  detection in networked industrial sensing systems,'' \emph{IEEE Transactions
  on Industrial Electronics}, vol.~62, no.~6, pp. 3832--3842, June 2015.

\bibitem{7232339}
Y.~Kwon, H.~K. Kim, Y.~H. Lim, and J.~I. Lim, ``A behavior-based intrusion
  detection technique for smart grid infrastructure,'' in \emph{Proc. 2015 IEEE
  Eindhoven PowerTech}, June 2015, pp. 1--6.

\bibitem{7762123}
H.~H. Pajouh, R.~Javidan, R.~Khayami, D.~Ali, and K.~K.~R. Choo, ``A two-layer
  dimension reduction and two-tier classification model for anomaly-based
  intrusion detection in iot backbone networks,'' \emph{IEEE Transactions on
  Emerging Topics in Computing}, vol.~PP, no.~99, pp. 1--1, 2016.

\bibitem{7510811}
H.~Sedjelmaci, S.~M. Senouci, and M.~Al-Bahri, ``A lightweight anomaly
  detection technique for low-resource iot devices: A game-theoretic
  methodology,'' in \emph{Proc. 2016 IEEE International Conference on
  Communications (ICC)}, May 2016, pp. 1--6.

\bibitem{7379452}
M.~S. Zitouni, J.~Dias, M.~Al-Mualla, and H.~Bhaskar, ``Hierarchical crowd
  detection and representation for big data analytics in visual surveillance,''
  in \emph{Proc. 2015 IEEE International Conference on Systems, Man, and
  Cybernetics (SMC)}, Oct 2015, pp. 1827--1832.

\bibitem{7247576}
P.~Sarigiannidis, E.~Karapistoli, and A.~A. Economides, ``Visiot: A threat
  visualisation tool for iot systems security,'' in \emph{Proc. 2015 IEEE
  International Conference on Communication Workshop (ICCW)}, June 2015, pp.
  2633--2638.

\bibitem{6800057}
D.~Takaishi, H.~Nishiyama, N.~Kato, and R.~Miura, ``Toward energy efficient big
  data gathering in densely distributed sensor networks,'' \emph{IEEE
  Transactions on Emerging Topics in Computing}, vol.~2, no.~3, pp. 388--397,
  Sept 2014.

\bibitem{7389318}
Z.~Asad, M.~A.~R. Chaudhry, and D.~Malone, ``Greener data exchange in the
  cloud: A coding-based optimization for big data processing,'' \emph{IEEE
  Journal on Selected Areas in Communications}, vol.~34, no.~5, pp. 1360--1377,
  May 2016.

\bibitem{179401}
\BIBentryALTinterwordspacing
A.~Das, C.~Lumezanu, Y.~Zhang, V.~Singh, G.~Jiang, and C.~Yu, ``Transparent and
  flexible network management for big data processing in the cloud,'' in
  \emph{Proc. the 5th USENIX Workshop on Hot Topics in Cloud Computing}.\hskip
  1em plus 0.5em minus 0.4em\relax Berkeley, CA: USENIX, 2013. [Online].
  Available:
  \url{https://www.usenix.org/conference/hotcloud13/workshop-program/presentations/Das}
\BIBentrySTDinterwordspacing

\bibitem{Perino}
\BIBentryALTinterwordspacing
D.~Perino, M.~Varvello, and K.~P.~N. Puttaswamy, ``Icn-re: Redundancy
  elimination for information-centric networking,'' in \emph{Proc. the Second
  Edition of the ICN Workshop on Information-centric Networking}, ser. ICN
  '12.\hskip 1em plus 0.5em minus 0.4em\relax New York, NY, USA: ACM, 2012, pp.
  91--96. [Online]. Available: \url{http://doi.acm.org/10.1145/2342488.2342508}
\BIBentrySTDinterwordspacing

\bibitem{7152629}
C.~Yan, Y.~Song, J.~Wang, and W.~Guo, ``Eliminating the redundancy in
  mapreduce-based entity resolution,'' in \emph{Cluster, Cloud and Grid
  Computing (CCGrid), 2015 15th IEEE/ACM International Symposium on}, May 2015,
  pp. 1233--1236.

\bibitem{7593221}
K.~Lee, D.~Kim, and I.~Shin, ``Reboost: Improving throughput in wireless
  networks using redundancy elimination,'' \emph{IEEE Communications Letters},
  vol.~PP, no.~99, pp. 1--1, 2016.

\bibitem{1427709}
N.~Li, J.~C. Hou, and L.~Sha, ``Design and analysis of an mst-based topology
  control algorithm,'' \emph{IEEE Transactions on Wireless Communications},
  vol.~4, no.~3, pp. 1195--1206, May 2005.

\bibitem{5351723}
K.~Miyao, H.~Nakayama, N.~Ansari, and N.~Kato, ``Ltrt: An efficient and
  reliable topology control algorithm for ad-hoc networks,'' \emph{IEEE
  Transactions on Wireless Communications}, vol.~8, no.~12, pp. 6050--6058,
  December 2009.

\bibitem{7217796}
H.~Lin, L.~Wang, and R.~Kong, ``Energy efficient clustering protocol for
  large-scale sensor networks,'' \emph{IEEE Sensors Journal}, vol.~15, no.~12,
  pp. 7150--7160, Dec 2015.

\bibitem{7588229}
L.~Kong, D.~Zhang, Z.~He, Q.~Xiang, J.~Wan, and M.~Tao, ``Embracing big data
  with compressive sensing: a green approach in industrial wireless networks,''
  \emph{IEEE Communications Magazine}, vol.~54, no.~10, pp. 53--59, October
  2016.

\bibitem{7478010}
Z.~Zhao, J.~Feng, and B.~Peng, ``A green distributed signal reconstruction
  algorithm in wireless sensor networks,'' \emph{IEEE Access}, vol.~4, pp.
  5908--5917, 2016.

\bibitem{6475927}
Y.~Simmhan, S.~Aman, A.~Kumbhare, R.~Liu, S.~Stevens, Q.~Zhou, and V.~Prasanna,
  ``Cloud-based software platform for big data analytics in smart grids,''
  \emph{Computing in Science Engineering}, vol.~15, no.~4, pp. 38--47, July
  2013.

\bibitem{7828559}
K.~Li, L.~Shu, M.~Mukherjee, D.~Wang, and L.~Hu, ``Prolonging network lifetime
  with sleep scheduling for solar harvesting industrial wsns,'' in \emph{Proc.
  2016 IEEE 18th International Conference on High Performance Computing and
  Communications; IEEE 14th International Conference on Smart City; IEEE 2nd
  International Conference on Data Science and Systems (HPCC/SmartCity/DSS)},
  Dec 2016, pp. 1532--1533.

\bibitem{7590092}
M.~Shojafar, C.~Canali, R.~Lancellotti, and J.~Abawajy, ``Adaptive
  computing-plus-communication optimization framework for multimedia processing
  in cloud systems,'' \emph{IEEE Transactions on Cloud Computing}, vol.~PP,
  no.~99, pp. 1--1, 2016.

\bibitem{Chao2015269}
\BIBentryALTinterwordspacing
H.~Chao and J.~Wu, ``15 - optimizing power saving in cellular networks for
  machine-to-machine (m2m) communications,'' in \emph{Machine-to-machine (M2M)
  Communications}, C.~Ant贸n-Haro and M.~Dohler, Eds.\hskip 1em plus 0.5em
  minus 0.4em\relax Oxford: Woodhead Publishing, 2015, pp. 269 -- 290.
  [Online]. Available:
  \url{http://www.sciencedirect.com/science/article/pii/B9781782421023000150}
\BIBentrySTDinterwordspacing

\bibitem{6162477}
H.~Chao, Y.~Chen, and J.~Wu, ``Power saving for machine to machine
  communications in cellular networks,'' in \emph{Proc. 2011 IEEE GLOBECOM
  Workshops (GC Wkshps)}, Dec 2011, pp. 389--393.

\bibitem{Mahadevan}
P.~Mahadevan, P.~Sharma, S.~Banerjee, and P.~Ranganathan, ``A power
  benchmarking framework for network devices,'' in \emph{Proc. the 8th
  International IFIP-TC 6 Networking Conference}, ser. NETWORKING '09.\hskip
  1em plus 0.5em minus 0.4em\relax Berlin, Heidelberg: Springer-Verlag, 2009,
  pp. 795--808.

\bibitem{6195471}
X.~Wang, Y.~Yao, X.~Wang, K.~Lu, and Q.~Cao, ``Carpo: Correlation-aware power
  optimization in data center networks,'' in \emph{Proc. 2012 Proceedings IEEE
  INFOCOM}, March 2012, pp. 1125--1133.

\bibitem{Heller:2010:ESE:1855711.1855728}
\BIBentryALTinterwordspacing
B.~Heller, S.~Seetharaman, P.~Mahadevan, Y.~Yiakoumis, P.~Sharma, S.~Banerjee,
  and N.~McKeown, ``Elastictree: Saving energy in data center networks,'' in
  \emph{Proc. the 7th USENIX Conference on Networked Systems Design and
  Implementation}, ser. NSDI'10.\hskip 1em plus 0.5em minus 0.4em\relax
  Berkeley, CA, USA: USENIX Association, 2010, pp. 17--17. [Online]. Available:
  \url{http://dl.acm.org/citation.cfm?id=1855711.1855728}
\BIBentrySTDinterwordspacing

\bibitem{6648647}
Y.~Zhang and N.~Ansari, ``Hero: Hierarchical energy optimization for data
  center networks,'' \emph{IEEE Systems Journal}, vol.~9, no.~2, pp. 406--415,
  June 2015.

\bibitem{6996601}
Y.~Han, S.~s.~Seo, J.~Li, J.~Hyun, J.~H. Yoo, and J.~W.~K. Hong, ``Software
  defined networking-based traffic engineering for data center networks,'' in
  \emph{Proc. 2014 16th Asia-Pacific Network Operations and Management
  Symposium (APNOMS)}, Sept 2014, pp. 1--6.

\bibitem{6888901}
L.~Wang, F.~Zhang, K.~Zheng, A.~V. Vasilakos, S.~Ren, and Z.~Liu,
  ``Energy-efficient flow scheduling and routing with hard deadlines in data
  center networks,'' in \emph{Proc. 2014 IEEE 34th International Conference on
  Distributed Computing Systems (ICDCS)}, June 2014, pp. 248--257.

\bibitem{7547281}
H.~Zhang, K.~Chen, W.~Bai, D.~Han, C.~Tian, H.~Wang, H.~Guan, and M.~Zhang,
  ``Guaranteeing deadlines for inter-data center transfers,'' \emph{IEEE/ACM
  Transactions on Networking}, vol.~PP, no.~99, pp. 1--17, 2016.

\bibitem{7348690}
J.~Zhang, K.~Li, D.~Guo, H.~Qi, W.~Li, and Y.~Jin, ``Mdfs: Deadline-driven flow
  scheduling scheme in multi-resource environments,'' \emph{IEEE Transactions
  on Multi-Scale Computing Systems}, vol.~1, no.~4, pp. 207--219, Oct 2015.

\bibitem{6689479}
L.~Wang, F.~Zhang, J.~A. Aroca, A.~V. Vasilakos, K.~Zheng, C.~Hou, D.~Li, and
  Z.~Liu, ``Greendcn: A general framework for achieving energy efficiency in
  data center networks,'' \emph{IEEE Journal on Selected Areas in
  Communications}, vol.~32, no.~1, pp. 4--15, January 2014.

\bibitem{Renault}
\BIBentryALTinterwordspacing
E.~Renault, ``Parallel execution of for loops using checkpointing techniques,''
  in \emph{Proc. the 2005 International Conference on Parallel Processing
  Workshops}, ser. ICPPW '05.\hskip 1em plus 0.5em minus 0.4em\relax
  Washington, DC, USA: IEEE Computer Society, 2005, pp. 313--319. [Online].
  Available: \url{http://dx.doi.org/10.1109/ICPPW.2005.65}
\BIBentrySTDinterwordspacing

\bibitem{Mereuta}
\BIBentryALTinterwordspacing
L.~Mereuta and E.~Renault, ``Checkpointing aided parallel execution model and
  analysis,'' in \emph{Proc. the Third International Conference on High
  Performance Computing and Communications}, ser. HPCC'07.\hskip 1em plus 0.5em
  minus 0.4em\relax Berlin, Heidelberg: Springer-Verlag, 2007, pp. 707--717.
  [Online]. Available: \url{http://dl.acm.org/citation.cfm?id=2401945.2402023}
\BIBentrySTDinterwordspacing

\bibitem{6984236}
.~Renault and S.~Boumerdassi, ``Towards an energy-efficient tool for processing
  the big data,'' in \emph{Proc. 2014 International Conference on Future
  Internet of Things and Cloud (FiCloud)}, Aug 2014, pp. 448--452.

\bibitem{6612229}
A.~Katal, M.~Wazid, and R.~H. Goudar, ``Big data: Issues, challenges, tools and
  good practices,'' in \emph{Contemporary Computing (IC3), 2013 Sixth
  International Conference on}, Aug 2013, pp. 404--409.

\bibitem{7438894}
X.~Tong, C.~Kang, and Q.~Xia, ``Smart metering load data compression based on
  load feature identification,'' \emph{IEEE Transactions on Smart Grid},
  vol.~7, no.~5, pp. 2414--2422, Sept 2016.

\bibitem{6412159}
S.~W. Jun, K.~E. Fleming, M.~Adler, and J.~Emer, ``Zip-io: Architecture for
  application-specific compression of big data,'' in \emph{Field-Programmable
  Technology (FPT), 2012 International Conference on}, Dec 2012, pp. 343--351.

\bibitem{7723680}
L.~Tian, H.~Wang, Q.~Tang, and Y.~Zhou, ``Surveillance source compression with
  background modeling for video big data,'' in \emph{2016 IEEE International
  Conferences on Big Data and Cloud Computing (BDCloud), Social Computing and
  Networking (SocialCom), Sustainable Computing and Communications (SustainCom)
  (BDCloud-SocialCom-SustainCom)}, Oct 2016, pp. 105--110.

\bibitem{chao2011power}
H.~Chao, Y.~Chen, and J.~Wu, ``Power saving for machine to machine
  communications in cellular networks,'' in \emph{2011 IEEE GLOBECOM Workshops
  (GC Wkshps}, December 2011, pp. 389--393.

\bibitem{SSS_Hao}
H.~Chen, L.~Liu, T.~Novlan, J.~D. Matyjas, B.~L. Ng, and J.~Zhang, ``Spatial
  spectrum sensing-based device-to-device cellular networks,'' \emph{IEEE
  Transactions on Wireless Communications}, vol.~15, no.~11, pp. 7299--7313,
  Nov 2016.

\bibitem{yang2016energy}
K.~Yang, S.~Martin, C.~Xing, J.~Wu, and R.~Fan, ``Energy-efficient power
  control for device-to-device communications.'' \emph{IEEE Journal on Selected
  Areas in Communications}, vol.~34, no.~12, pp. 3208--3220, December 2016.

\bibitem{xu2018towards}
Y.~Xu, S.~Jiang, and J.~Wu, ``Towards energy efficient device-to-device content
  dissemination in cellular networks,'' \emph{IEEE Access}, vol.~6, pp.
  25\,816--25\,828, 2018.

\bibitem{6388472}
A. and H.~Y. Kong, ``Energy efficient cooperative leach protocol for wireless
  sensor networks,'' \emph{Journal of Communications and Networks}, vol.~12,
  no.~4, pp. 358--365, Aug 2010.

\bibitem{ComMag_KWang_12_2016}
K.~Wang, Y.~Wang, Y.~Sun, S.~Guo, and J.~Wu, ``Green industrial internet of
  things architecture: An energy-efficient perspective,'' \emph{IEEE
  Communications Magazine}, vol.~54, no.~12, pp. 48--54, December 2016.

\bibitem{atat2018green}
R.~Atat, L.~Liu, J.~Wu, J.~Ashdown, and Y.~Yi, ``Green massive traffic
  offloading for cyber-physical systems over heterogeneous cellular networks,''
  \emph{Mobile Networks and Applications}, pp. 1--9, 2018.

\bibitem{Greenberg:2008:CCR:1496091.1496103}
\BIBentryALTinterwordspacing
A.~Greenberg, J.~Hamilton, D.~A. Maltz, and P.~Patel, ``The cost of a cloud:
  Research problems in data center networks,'' \emph{SIGCOMM Comput. Commun.
  Rev.}, vol.~39, no.~1, pp. 68--73, Dec. 2008. [Online]. Available:
  \url{http://doi.acm.org/10.1145/1496091.1496103}
\BIBentrySTDinterwordspacing

\bibitem{J_TBD_HLiu_n2_2018}
H.~Liu, B.~Liu, L.~T. Yang, M.~Lin, Y.~Deng, K.~Bilal, and S.~U. Khan,
  ``Thermal-aware and {DVFS}-enabled big data task scheduling for data
  centers,'' \emph{IEEE Transactions on Big Data}, vol.~4, no.~2, pp. 177--190,
  2018.

\bibitem{7474371}
H.~Ayyalasomayajula, E.~Gabriel, P.~Lindner, and D.~Price, ``Air quality
  simulations using big data programming models,'' in \emph{2016 IEEE Second
  International Conference on Big Data Computing Service and Applications
  (BigDataService)}, March 2016, pp. 182--184.

\bibitem{7179453}
J.~Y. Zhu, C.~Sun, and V.~O.~K. Li, ``Granger-causality-based air quality
  estimation with spatio-temporal (s-t) heterogeneous big data,'' in \emph{2015
  IEEE Conference on Computer Communications Workshops (INFOCOM WKSHPS)}, April
  2015, pp. 612--617.

\bibitem{7562036}
J.~Y. Zhu, Y.~Zheng, X.~Yi, and V.~O.~K. Li, ``A gaussian bayesian model to
  identify spatio-temporal causalities for air pollution based on urban big
  data,'' in \emph{2016 IEEE Conference on Computer Communications Workshops
  (INFOCOM WKSHPS)}, April 2016, pp. 3--8.

\bibitem{Fingas20149}
\BIBentryALTinterwordspacing
M.~Fingas and C.~Brown, ``Review of oil spill remote sensing,'' \emph{Marine
  Pollution Bulletin}, vol.~83, no.~1, pp. 9 -- 23, 2014. [Online]. Available:
  \url{http://www.sciencedirect.com/science/article/pii/S0025326X14002021}
\BIBentrySTDinterwordspacing

\bibitem{7168518}
G.~Suciu, V.~Suciu, C.~Dobre, and C.~Chilipirea, ``Tele-monitoring system for
  water and underwater environments using cloud and big data systems,'' in
  \emph{Proc. 2015 20th International Conference on Control Systems and
  Computer Science}, May 2015, pp. 809--813.

\bibitem{Zheng:2014}
\BIBentryALTinterwordspacing
Y.~Zheng, T.~Liu, Y.~Wang, Y.~Zhu, Y.~Liu, and E.~Chang, ``Diagnosing new york
  city's noises with ubiquitous data,'' in \emph{Proc. the 2014 ACM
  International Joint Conference on Pervasive and Ubiquitous Computing}, ser.
  UbiComp '14.\hskip 1em plus 0.5em minus 0.4em\relax New York, NY, USA: ACM,
  2014, pp. 715--725. [Online]. Available:
  \url{http://doi.acm.org/10.1145/2632048.2632102}
\BIBentrySTDinterwordspacing

\bibitem{7184864}
S.~Hachem, V.~Mallet, R.~Ventura, A.~Pathak, V.~Issarny, P.~G. Raverdy, and
  R.~Bhatia, ``Monitoring noise pollution using the urban civics middleware,''
  in \emph{Proc. 2015 IEEE First International Conference on Big Data Computing
  Service and Applications (BigDataService)}, March 2015, pp. 52--61.

\bibitem{7037178}
S.~Bera, S.~Misra, and M.~S. Obaidat, ``Energy-efficient smart metering for
  green smart grid communication,'' in \emph{Proc. 2014 IEEE Global
  Communications Conference}, Dec 2014, pp. 2466--2471.

\bibitem{5175339}
X.~Li, C.~P. Bowers, and T.~Schnier, ``Classification of energy consumption in
  buildings with outlier detection,'' \emph{IEEE Transactions on Industrial
  Electronics}, vol.~57, no.~11, pp. 3639--3644, Nov 2010.

\bibitem{7113543}
C.~H. Lee and C.~H. Wu, ``Collecting and mining big data for electric vehicle
  systems using battery modeling data,'' in \emph{Proc. 2015 12th International
  Conference on Information Technology - New Generations (ITNG)}, April 2015,
  pp. 626--631.

\bibitem{Qin:2014}
\BIBentryALTinterwordspacing
Y.~B. Qin, J.~Housell, and I.~Rodero, ``Cloud-based data analytics framework
  for autonomic smart grid management,'' in \emph{Proc. the 2014 International
  Conference on Cloud and Autonomic Computing}, ser. ICCAC '14.\hskip 1em plus
  0.5em minus 0.4em\relax Washington, DC, USA: IEEE Computer Society, 2014, pp.
  97--100. [Online]. Available: \url{http://dx.doi.org/10.1109/ICCAC.2014.39}
\BIBentrySTDinterwordspacing

\bibitem{7084108}
F.~Saremi, O.~Fatemieh, H.~Ahmadi, H.~Wang, T.~Abdelzaher, R.~Ganti, H.~Liu,
  S.~Hu, S.~Li, and L.~Su, ``Experiences with greengps--fuel-efficient
  navigation using participatory sensing,'' \emph{IEEE Transactions on Mobile
  Computing}, vol.~15, no.~3, pp. 672--689, March 2016.

\bibitem{hamida2015security}
E.~B. Hamida, H.~Noura, and W.~Znaidi, ``Security of cooperative intelligent
  transport systems: Standards, threats analysis and cryptographic
  countermeasures,'' \emph{Electronics}, vol.~4, no.~3, pp. 380--423, 2015.

\bibitem{ge3s}
\BIBentryALTinterwordspacing
Environmental monitoring. GE3S. Accessed on Accessed on December 1, 2017.
  [Online]. Available:
  \url{http://www.ge3s.org/service/environmental-monitoring/}
\BIBentrySTDinterwordspacing

\bibitem{defense}
\BIBentryALTinterwordspacing
Defense industry - remote management application. Defense Industry Remote
  Management Application. Accessed on December 1, 2017. [Online]. Available:
  \url{https://www.wti.com/t-defense-remote-management-application.aspx}
\BIBentrySTDinterwordspacing

\bibitem{facilities}
\BIBentryALTinterwordspacing
Smart facilities. Facilities Booking System. Accessed on December 1, 2017.
  [Online]. Available: \url{http://www.facilitiesbooking.com/smart-facilities}
\BIBentrySTDinterwordspacing

\bibitem{nanoworks}
\BIBentryALTinterwordspacing
Naonworks. Accessed on January 10, 2018. [Online]. Available:
  \url{http://www.naonworks.com/old/inc\_html/sub2\_3.html}
\BIBentrySTDinterwordspacing

\bibitem{google}
\BIBentryALTinterwordspacing
Mp08 comunicacions industrials - francesc cazorla. Accessed on June 1, 2018.
  [Online]. Available: \url{https://sites.google.com/site/fcazorlay/mp08}
\BIBentrySTDinterwordspacing

\bibitem{6905754}
J.~Baek, Q.~H. Vu, J.~K. Liu, X.~Huang, and Y.~Xiang, ``A secure cloud
  computing based framework for big data information management of smart
  grid,'' \emph{IEEE Transactions on Cloud Computing}, vol.~3, no.~2, pp.
  233--244, April 2015.

\bibitem{6873776}
A.~Ukil and R.~Zivanovic, ``Automated analysis of power systems disturbance
  records: Smart grid big data perspective,'' in \emph{Proc. 2014 IEEE
  Innovative Smart Grid Technologies - Asia (ISGT ASIA)}, May 2014, pp.
  126--131.

\bibitem{7741444}
J.~Yang, J.~Zhao, F.~Wen, W.~Kong, and Z.~Dong, ``Mining the big data of
  residential appliances in the smart grid environment,'' in \emph{Proc. 2016
  IEEE Power and Energy Society General Meeting (PESGM)}, July 2016, pp. 1--5.

\bibitem{7069405}
L.~Liu and Z.~Han, ``Multi-block admm for big data optimization in smart
  grid,'' in \emph{Proc. 2015 International Conference on Computing, Networking
  and Communications (ICNC)}, Feb 2015, pp. 556--561.

\bibitem{6279584}
S.~Cui, Z.~Han, S.~Kar, T.~T. Kim, H.~V. Poor, and A.~Tajer, ``Coordinated
  data-injection attack and detection in the smart grid: A detailed look at
  enriching detection solutions,'' \emph{IEEE Signal Processing Magazine},
  vol.~29, no.~5, pp. 106--115, Sept 2012.

\bibitem{7342880}
A.~Yassine, A.~A.~N. Shirehjini, and S.~Shirmohammadi, ``Smart meters big data:
  Game theoretic model for fair data sharing in deregulated smart grids,''
  \emph{IEEE Access}, vol.~3, pp. 2743--2754, 2015.

\bibitem{7154500}
X.~He, Q.~Ai, R.~C. Qiu, W.~Huang, L.~Piao, and H.~Liu, ``A big data
  architecture design for smart grids based on random matrix theory,''
  \emph{IEEE Transactions on Smart Grid}, vol.~8, no.~2, pp. 674--686, March
  2017.

\bibitem{7587350}
J.~Wu, K.~Ota, M.~Dong, J.~Li, and H.~Wang, ``Big data analysis based security
  situational awareness for smart grid,'' \emph{IEEE Transactions on Big Data},
  vol.~PP, no.~99, pp. 1--1, 2017.

\bibitem{7909159}
K.~Wang, Y.~Wang, X.~Hu, Y.~Sun, D.~J. Deng, A.~Vinel, and Y.~Zhang, ``Wireless
  big data computing in smart grid,'' \emph{IEEE Wireless Communications},
  vol.~24, no.~2, pp. 58--64, April 2017.

\bibitem{J_TII_KHam_11_2017}
K.~Hamedani, L.~Liu, A.~Rachad, J.~Wu, and Y.~Yi, ``Reservoir computing meets
  smart grids: Attack detection using delayed feedback networks,'' \emph{IEEE
  Transactions on Industrial Informatics}, 2017.

\bibitem{6882222}
A.~A. Yavuz, ``An efficient real-time broadcast authentication scheme for
  command and control messages,'' \emph{IEEE Transactions on Information
  Forensics and Security}, vol.~9, no.~10, pp. 1733--1742, Oct 2014.

\bibitem{7028577}
V.~Nguyen, M.~Guirguis, and G.~Atia, ``A unifying approach for the
  identification of application-driven stealthy attacks on mobile cps,'' in
  \emph{Proc. 2014 52nd Annual Allerton Conference on Communication, Control,
  and Computing (Allerton)}, Sept 2014, pp. 1093--1101.

\bibitem{6735756}
G.~M. Lehto, G.~Edlund, T.~Smigla, and F.~Afinidad, ``Protection evaluation
  framework for tactical satcom architectures,'' in \emph{Proc. MILCOM 2013 -
  2013 IEEE Military Communications Conference}, Nov 2013, pp. 1008--1013.

\bibitem{7558230}
M.~V. Moreno, F.~Terroso-Saenz, A.~Gonzalez-Vidal, M.~Valdez-Vela, A.~F.
  Skarmeta, M.~A. Zamora, and V.~Chang, ``Applicability of big data techniques
  to smart cities deployments,'' \emph{IEEE Transactions on Industrial
  Informatics}, vol.~13, no.~2, pp. 800--809, April 2017.

\bibitem{7400610}
M.~M. Rathore, A.~Ahmad, A.~Paul, and G.~Jeon, ``Efficient graph-oriented smart
  transportation using internet of things generated big data,'' in \emph{Proc.
  2015 11th International Conference on Signal-Image Technology Internet-Based
  Systems (SITIS)}, Nov 2015, pp. 512--519.

\bibitem{7797384}
B.~T. de~Oliveira and C.~B. Margi, ``Distributed control plane architecture for
  software-defined wireless sensor networks,'' in \emph{Proc. 2016 IEEE
  International Symposium on Consumer Electronics (ISCE)}, Sept 2016, pp.
  85--86.

\bibitem{5306098}
A.~Pantelopoulos and N.~G. Bourbakis, ``A survey on wearable sensor-based
  systems for health monitoring and prognosis,'' \emph{IEEE Transactions on
  Systems, Man, and Cybernetics, Part C (Applications and Reviews)}, vol.~40,
  no.~1, pp. 1--12, Jan 2010.

\bibitem{s141018009}
E.~Kartsakli, A.~S. Lalos, A.~Antonopoulos, S.~Tennina, M.~D. Renzo, L.~Alonso,
  and C.~Verikoukis, ``A survey on {M2M} systems for mhealth: A wireless
  communications perspective,'' \emph{Sensors}, vol.~14, no.~10, pp.
  18\,009--18\,052, 2014.

\bibitem{Yuce2010116}
\BIBentryALTinterwordspacing
M.~R. Yuce, ``Implementation of wireless body area networks for healthcare
  systems,'' \emph{Sensors and Actuators A: Physical}, vol. 162, no.~1, pp. 116
  -- 129, 2010. [Online]. Available:
  \url{http://www.sciencedirect.com/science/article/pii/S0924424710002657}
\BIBentrySTDinterwordspacing

\bibitem{5681102}
H.~Yan, H.~Huo, Y.~Xu, and M.~Gidlund, ``Wireless sensor network based e-health
  system - implementation and experimental results,'' \emph{IEEE Transactions
  on Consumer Electronics}, vol.~56, no.~4, pp. 2288--2295, November 2010.

\bibitem{Martinez2011485}
\BIBentryALTinterwordspacing
J.~F. Martinez, M.~S. Familiar, I.~Corredor, A.~B. Garcia, S.~Bravo, and
  L.~Lopez, ``Composition and deployment of e-health services over wireless
  sensor networks,'' \emph{Mathematical and Computer Modelling}, vol.~53, no.
  3-4, pp. 485--503, 2011, telecommunications Software Engineering: Emerging
  Methods, Models and Tools. [Online]. Available:
  \url{http://www.sciencedirect.com/science/article/pii/S0895717710001548}
\BIBentrySTDinterwordspacing

\bibitem{6775278}
P.~Jiang, J.~Winkley, C.~Zhao, R.~Munnoch, G.~Min, and L.~T. Yang, ``An
  intelligent information forwarder for healthcare big data systems with
  distributed wearable sensors,'' \emph{IEEE Systems Journal}, vol.~10, no.~3,
  pp. 1147--1159, Sept 2016.

\bibitem{6599064}
G.~Baldini, S.~Karanasios, D.~Allen, and F.~Vergari, ``Survey of wireless
  communication technologies for public safety,'' \emph{IEEE Communications
  Surveys and Tutorials}, vol.~16, no.~2, pp. 619--641, Feb. 2014.

\bibitem{7462716}
D.~Cui, ``Risk early warning index system in the field of public safety in big
  data era,'' in \emph{Proc. 2015 Sixth International Conference on Intelligent
  Systems Design and Engineering Applications (ISDEA)}, Aug 2015, pp. 704--707.

\bibitem{7730540}
U.~M. Bhangale, K.~R. Kurte, S.~S. Durbha, R.~L. King, and N.~H. Younan, ``Big
  data processing using hpc for remote sensing disaster data,'' in \emph{Proc.
  2016 IEEE International Geoscience and Remote Sensing Symposium (IGARSS)},
  July 2016, pp. 5894--5897.

\bibitem{7471282}
L.~Zhong, K.~Takano, Y.~Ji, and S.~Yamada, ``Big data based service area
  estimation for mobile communications during natural disasters,'' in
  \emph{Proc. 2016 30th International Conference on Advanced Information
  Networking and Applications Workshops (WAINA)}, March 2016, pp. 687--692.

\bibitem{6846628}
P.~Tin, T.~T. Zin, T.~Toriu, and H.~Hama, ``An integrated framework for
  disaster event analysis in big data environments,'' in \emph{Proc. 2013 Ninth
  International Conference on Intelligent Information Hiding and Multimedia
  Signal Processing}, Oct 2013, pp. 255--258.

\bibitem{ABSOLUTE}
\BIBentryALTinterwordspacing
{FP7 Project ABSOLUTE}. Accessed on July 10, 2017. [Online]. Available:
  \url{http://cordis.europa.eu/project/rcn/106035_en.html}
\BIBentrySTDinterwordspacing

\bibitem{6619579}
R.~Ferrus, O.~Sallent, G.~Baldini, and L.~Goratti, ``{LTE}: the technology
  driver for future public safety communications,'' \emph{IEEE Communications
  Magazine}, vol.~51, no.~10, pp. 154--161, October 2013.

\bibitem{flynn5etsi}
K.~Flynn, ``{ETSI Summit on Future Mobile and Standards for 5G, 3GPP}, nov.
  2013.''

\bibitem{lucent2010government}
A.~Lucent, ``Government technology how-to guide: for {LTE} in public safety,''
  \emph{Government Technology Magazine}, 2010.

\bibitem{witkowski2013effectively}
D.~Witkowski, ``Effectively testing 700 {MHz} public safety {LTE} broadband and
  {P25} narrowband networks,'' \emph{White paper by Anritsu Company, Accessed
  on July 10, 2017. [Online]. Available http://www. anritsu.
  com/en-US/Products-Solutions/Solution/Effectively-testing-700MHz-public-safety.
  aspx [Accessed April 2013]}, 2013.

\bibitem{6817780}
Y.~Xiao, K.-C. Chen, C.~Yuen, and L.~DaSilva, ``Spectrum sharing for
  device-to-device communications in cellular networks: A game theoretic
  approach,'' in \emph{Proc. 2014 IEEE International Symposium on Dynamic
  Spectrum Access Networks (DYSPAN)}, April 2014, pp. 60--71.

\bibitem{6093225}
A.~Bagayoko, B.~Paillassa, and R.~Dhaou, ``Practical link reliability for ad
  hoc routing protocol,'' in \emph{Proc. 2011 IEEE Vehicular Technology
  Conference (VTC Fall)}, Sept 2011, pp. 1--5.

\bibitem{7345413}
M.~Wang and Z.~Yan, ``Security in {D2D} communications: A review,'' in
  \emph{Proc. 2015 IEEE Trustcom/BigDataSE/ISPA}, vol.~1, Aug 2015, pp.
  1199--1204.

\bibitem{7417841}
Y.~Gong, Y.~Fang, and Y.~Guo, ``Privacy-preserving collaborative learning for
  mobile health monitoring,'' in \emph{Proc. 2015 IEEE Global Communications
  Conference (GLOBECOM)}, Dec 2015.

\bibitem{6386942}
J.~Pospiil and M.~Novotn媒, ``Evaluating cryptanalytical strength of
  lightweight cipher present on reconfigurable hardware,'' in \emph{Proc. 2012
  15th Euromicro Conference on Digital System Design (DSD)}, Sept 2012, pp.
  560--567.

\bibitem{7101220}
Z.~Li and T.~J. Oechtering, ``Privacy-aware distributed bayesian detection,''
  \emph{IEEE Journal of Selected Topics in Signal Processing}, vol.~9, no.~7,
  pp. 1345--1357, Oct 2015.

\bibitem{6831320}
F.~Canelo, B.~M.~C. Silva, J.~J. P.~C. Rodrigues, and Z.~Zhu, ``Performance
  evaluation of an enhanced cryptography solution for m-health applications in
  cooperative environments,'' in \emph{Proc. 2013 IEEE Global Communications
  Conference (GLOBECOM)}, Dec 2013, pp. 1711--1716.

\bibitem{rayaccess}
I.~Ray and I.~Ray, ``Access control challenges for cyber-physical systems.''

\bibitem{7562128}
K.~Ly, W.~Sun, and Y.~Jin, ``Emerging challenges in cyber-physical systems: A
  balance of performance, correctness, and security,'' in \emph{Proc. 2016 IEEE
  Conference on Computer Communications Workshops (INFOCOM WKSHPS)}, April
  2016, pp. 498--502.

\bibitem{Cancila}
\BIBentryALTinterwordspacing
D.~Cancila, H.~Zaatiti, and R.~Passerone, ``Cyber-physical system and
  contract-based design: A three dimensional view,'' in \emph{Proc. the
  WESE'15: Workshop on Embedded and Cyber-Physical Systems Education}, ser.
  WESE'15.\hskip 1em plus 0.5em minus 0.4em\relax New York, NY, USA: ACM, 2015,
  pp. 4:1--4:4. [Online]. Available:
  \url{http://doi.acm.org/10.1145/2832920.2832924}
\BIBentrySTDinterwordspacing

\bibitem{4606742}
A.~G. T茅llez and M.~M. Pl谩, ``Multithreaded translation of ptolemy ii
  designs on multicore platforms,'' in \emph{Complex, Intelligent and Software
  Intensive Systems, 2008. CISIS 2008. International Conference on}, March
  2008, pp. 607--612.

\bibitem{6843708}
H.~Chen and S.~Mitra, ``Synthesis and verification of motor-transmission shift
  controller for electric vehicles,'' in \emph{Proc. 2014 ACM/IEEE
  International Conference on Cyber-Physical Systems (ICCPS)}, April 2014, pp.
  25--35.

\bibitem{7167224}
J.~Espinosa, C.~Hernandez, J.~Abella, D.~de~Andres, and J.~C. Ruiz, ``Analysis
  and rtl correlation of instruction set simulators for automotive
  microcontroller robustness verification,'' in \emph{2015 52nd ACM/EDAC/IEEE
  Design Automation Conference (DAC)}, June 2015, pp. 1--6.

\bibitem{7423131}
M.~M. Bersani and M.~Garcia-Valls, ``The cost of formal verification in
  adaptive cps. an example of a virtualized server node,'' in \emph{Proc. 2016
  IEEE 17th International Symposium on High Assurance Systems Engineering
  (HASE)}, Jan 2016, pp. 39--46.

\bibitem{Mastronarde_D2D}
N.~Mastronarde, V.~Patel, J.~Xu, L.~Liu, and M.~van~der Schaar, ``To relay or
  not to relay: Learning device-to-device relaying strategies in cellular
  networks,'' \emph{IEEE Transactions on Mobile Computing}, vol.~15, no.~6, pp.
  1569--1585, June 2016.

\bibitem{Goodman:1997:MDF:549931}
I.~R. Goodman, R.~P. Mahler, and H.~T. Nguyen, \emph{Mathematics of Data
  Fusion}.\hskip 1em plus 0.5em minus 0.4em\relax Norwell, MA, USA: Kluwer
  Academic Publishers, 1997.

\bibitem{Ali2016}
A.~Ali, J.~Qadir, R.~u. Rasool, A.~Sathiaseelan, A.~Zwitter, and J.~Crowcroft,
  ``Big data for development: applications and techniques,'' \emph{Big Data
  Analytics}, vol.~1, no.~2, 2016.

\end{thebibliography}

\begin{IEEEbiography}{Rachad~Atat} received the B.E. degree (with distinction) in computer engineering from the Lebanese American University, Beirut, Lebanon, in 2010, the M.Sc. degree in electrical engineering from the King Abdullah University of Science and Technology (KAUST), Thuwal, Saudi Arabia, in 2012, and the Ph.D. degree (with Hons.) in electrical engineering from the University of Kansas (KU), Lawrence, KS, USA, in 2017. He is currently a postdoctoral research associate at Texas A\&M University at Qatar, working on dynamic metering allocation with integrated cybersecurity measures in smart grids. His current research interests include smart grids, cybersecurity, and Internet of Things. Dr. Atat was a recipient of the 2016 IEEE Global Communications Conference Best Paper Award, the NSF Travel Grant Award in 2016, the KU Engineering Fellowship Award, and the KAUST Discovery Scholarship Award.
\end{IEEEbiography}

\begin{IEEEbiography}{Lingjia~Liu} (SM'15) received the B.S. degree in Electronic Engineering from Shanghai Jiao Tong University, Shanghai, China, and the Ph.D. degree in Electrical and Computer Engineering from Texas A\&M University, College Station, TX, USA. He spent more than three years working in the Standards \& Mobility Innovation Laboratory, Samsung Research America (SRA) where he received the Global Samsung Best Paper Award twice (in 2008 and 2010, respectively). He was leading Samsung’s efforts on multiuser MIMO, coordinated multipoint (CoMP), and heterogeneous networks (HetNets) in 3GPP LTE/LTE-Advanced standards.

Lingjia Liu is an Associate Professor in the Bradley Department of Electrical and Computer Engineering (ECE) of Virginia Tech (VT). He is also serving as the Associate Director for Affiliate Relations in Wireless@Virginia Tech. Prior to that he was an Associate Professor in the Electrical Engineering and Computer Sciences Department, University of Kansas (KU). His research interests include emerging technologies for Beyond 5G cellular networks including machine learning for wireless networks, massive MIMO, massive MTC communications, and mmWave communications. Lingjia Liu was a recipient of the Air Force Summer Faculty Fellow from 2013 to 2017, the Miller Scholar at KU in 2014, the Miller Professional Development Award for Distinguished Research at KU in 2015, the 2016 \emph{IEEE} GLOBECOM Best Paper Award, the 2018 \emph{IEEE} ISQED Best Paper Award, the 2018 \emph{IEEE} TAOS Best Paper Award, and the 2018 \emph{IEEE} TCGCC Best Conference Paper Award.
\end{IEEEbiography}

\begin{IEEEbiography}{Jinsong Wu} (SM'11) received his Ph.D. in Electrical and Computer Engineering from Queen's University, Kingston, Canada. He was the leading Editor and a co-author of the comprehensive book, entitled ``Green Communications: Theoretical Fundamentals, Algorithms, and Applications'', published by CRC Press in September 2012. His first-authored paper won 2017 IEEE System Journal Best Paper Award. He received 3 best paper awards for IEEE conference papers. He received IEEE Green Communications and Computing Technical Committee 2017 Excellent Services Award for Excellent Technical Leadership and Services in the Green Communications and Computing Community. He also received IEEE Outstanding Leadership Award in 2013 IEEE International Conference on Green Computing and Communications (Greencom) and 2016  IEEE International Conference on Big Data Intelligence and Computing (DataCom), IEEE Computer Society. He is elected Vice Chair, Technical Activities, IEEE Environmental Engineering Initiative, a pan-IEEE effort under IEEE Technical Activities Board (TAB). He was the founder (2011) and founding Chair (2011-2017) of IEEE Technical Committee on Green Communications and Computing (TCGCC). He is also the primary co-founder (2015) and founding Vice-Chair of IEEE Technical Committee on Big Data (TCBD). He was the proposer (2012), founder (2014) and Series Editor (2014 - 2017) of IEEE Series on Green Communication and Computing Networks in the IEEE Communications Magazine. He is Area Editor (2016 - present) of IEEE Transactions on Green Communications and Networking. He was the original proposer (2012) and Series Editor (2014-2016) of the IEEE Journal of Selected Areas on Communications (JSAC) Series on Green Communications and Networking. He is Editor of IEEE Communications Surveys and Tutorials, Associate Editor of IEEE Systems Journal, Associate Editor of IEEE Access.
He was the one of the earliest key proposers and a long term promotor from Green Track to a full green symposium in flagship conferences of IEEE Communications Society. He opened and established Big Data Track with general and wide topic coverage in the flagship conferences of IEEE Communications Society, starting at IEEE Globecom 2016. He has been conference chairs and main organizers in a number of IEEE conferences and workshops.
\end{IEEEbiography}

\begin{IEEEbiography}{Guangyu Li} received the B.S. degree from China University of Mining and Technology and M.S. degree from Tongji University, China, in 2008 and 2011, respectively, and the Ph.D. degree in computer science from University of Paris-Sud, Paris, France, in 2015. He is currently working as an assistant professor with the Key Laboratory of Intelligent Perception and Systems for High-Dimensional Information of Ministry of Education, Nanjing University of Science
and Technology, Nanjing, China. His current research interests include routing protocol design for vehicular networks, big data transmission, electric vehicles charging/discharging strategy design in smart grid and traffic control.
\end{IEEEbiography}

\begin{IEEEbiography}{Chunxuan Ye} received his Ph.D. from University of Maryland, College Park in 2005. Between 2005 and 2012, he worked for InterDigital Communications, Inc. on multiple advanced research projects in the wireless communications field. He worked for Intel on LTE modem chipset development between 2012 and 2015. He has been working for InterDigital Communications, Inc. as a 3GPP RAN1 standards delegate for new radios (5G) since 2015. Chunxuan Ye has more than 30 book chapters, conference and journal papers. He also has more than 60 U.S. and international patent applications with 25 U.S. patents granted. His specialties include wireless communications, wireless standards and technologies, information theory, software and firmware development.
\end{IEEEbiography}

\begin{IEEEbiography}{Yang Yi} (M'09-SM'16) is an assistant professor in the Bradley Department of Electrical Engineering and Computer engineering at the Virginia Tech. She is an IEEE senior member. She received the B.S. and M.S. degrees in electronic engineering at Shanghai Jiao Tong University, and the Ph.D. degree in electrical and computer engineering at Texas A\&M University. Her research interests include very large scale integrated (VLSI) circuits and systems, computer-aided design (CAD), neuromorphic architecture for brain-inspired computing systems, and low-power circuits design with advanced nano-technologies for high-speed wireless systems..
\end{IEEEbiography}








\end{document}